\newif\ifarxiv 
\newif\ifsimods
\newif\ifaistats

\simodsfalse
\aistatsfalse
\arxivtrue

\ifarxiv
    \documentclass[11pt]{article}
    \usepackage{deepthink}

\usepackage{amsthm}

\usepackage[utf8]{inputenc}

\usepackage[utf8]{inputenc} 
\usepackage[T1]{fontenc}    
\usepackage{url}
\usepackage{hyperref}
\usepackage{amsmath,amssymb,amsbsy,amsfonts,amscd,bm}

\usepackage{xcolor}
\usepackage{xspace}
\usepackage{color}
\usepackage{graphicx}
\graphicspath{{./figs/}}
\usepackage{algorithm}
\usepackage{algorithmic}
\usepackage{comment}
\usepackage{multirow}
\usepackage{enumitem}
\usepackage{fancyhdr}
\usepackage{authblk}
\usepackage{tikz}
\usepackage{tikz-3dplot}
\usetikzlibrary{quotes, angles}
\usepackage{afterpage}
\usepackage{cleveref}
\usepackage{subcaption}
\usepackage{wrapfig}
\usepackage[toc, page]{appendix}
\usepackage[authoryear]{natbib} 
\usepackage{bbm}
\usepackage{tcolorbox}

\newtheorem{assum}{Assumption}

\theoremstyle{remark}
\newtheorem{problem}{Problem}

\theoremstyle{plain}
\newtheorem{theorem}{Theorem}[section]

\newtheorem{lemma}[theorem]{Lemma}
\newtheorem{corollary}[theorem]{Corollary}
\theoremstyle{definition}
\newtheorem{definition}[theorem]{Definition}

\def \reals   { \mathbb{R} }

\newcommand{\e}{\begin{equation}}
\newcommand{\ee}{\end{equation}}
\newcommand{\en}{\begin{equation*}}
\newcommand{\een}{\end{equation*}}
\newcommand{\eqn}{\begin{eqnarray}}
\newcommand{\eeqn}{\end{eqnarray}}
\newcommand{\bmat}{\begin{bmatrix}}
\newcommand{\emat}{\end{bmatrix}}

\DeclareMathAlphabet\mathbfcal{OMS}{cmsy}{b}{n}

\newcommand{\mtx}[1]{\boldsymbol{#1}}

\newcommand{\rank}{\operatorname{rank}}

\newcommand{\calA}{\mathcal{A}}

\newcommand{\calI}{\mathcal{I}}
\newcommand{\calJ}{\mathcal{J}}

\newcommand{\calN}{\mathcal{N}}

\newcommand{\calR}{\mathcal{R}}
\newcommand{\calS}{\mathcal{S}}

\newcommand{\calX}{\mathcal{X}}

\newcommand{\mC}{\mtx{C}}

\newcommand{\mS}{\mtx{S}}

\newcommand{\mSigma}{\mtx{\Sigma}}

\graphicspath{{./figs/}}

\newlength{\imgwidth}
\newboolean{twoColVersion}
\setboolean{twoColVersion}{false}
\newcommand{\twoCol}[2]{\ifthenelse{\boolean{twoColVersion}} {#1} {#2} }

\long\def\comment#1{}

\newcommand{\xmath}[1] {\ensuremath{#1}\xspace}
\newcommand{\blmath}[1] {\xmath{\bm{#1}}}
\newcommand{\A}{\blmath{A}}
\newcommand{\B}{\blmath{B}}
\newcommand{\D}{\blmath{D}}

\newcommand{\I}{\blmath{I}}
\newcommand{\Sb}{\blmath{S}}

\newcommand{\Q}{\blmath{Q}}
\newcommand{\U}{\blmath{U}}
\newcommand{\V}{\blmath{V}}
\newcommand{\W}{\blmath{W}}
\newcommand{\X}{\blmath{X}}
\newcommand{\Y}{\blmath{Y}}

\newcommand{\ab}{\blmath{a}}
\newcommand{\blb}{\blmath{b}}

\newcommand{\w}{\blmath{w}}
\newcommand{\x}{\blmath{x}}

\newcommand{\y}{\blmath{y}}
\newcommand{\z}{\blmath{z}}

\newcommand{\0}{\blmath{0}} 

\newcommand{\ie}{\text{i.e.}}
\newcommand{\eg}{\text{e.g.}}

\renewcommand{\u} {\blmath{u}}
\renewcommand{\v} {\blmath{v}}

\long\def\red#1{\bgroup\color{red}#1\egroup}

\definecolor{mich-blue}{HTML}{0027CC}
\definecolor{mich-blue-high}{HTML}{0027CC}
\definecolor{red-high}{HTML}{CA2020}
\definecolor{green-high}{HTML}{20A520}
\definecolor{mich-maize}{HTML}{FFCB05}
\definecolor{law-stone}{HTML}{655A52}
\definecolor{burton-beige}{HTML}{9B9A9D}
\definecolor{arch-ivy}{HTML}{7E732F}

 \colorlet{color1}{gray!15}

\newcommand{\eE}{\xmath{\mathbb{E}}}

\newcommand{\exprm}{\mathrm{exp}}
\newcommand{\trrm}{\mathrm{Tr}}

\newcommand{\varrm}{\mathrm{Var}}
\newcommand{\covrm}{\mathrm{Cov}}
\newcommand{\maxrm}{\mathrm{max}}
\newcommand{\minrm}{\mathrm{min}}

    \newcommand{\corrauth}{\textsuperscript{\ddag}}

    \title{Linearly Separable Features in Shallow Nonlinear Networks: Width Scales Polynomially with Intrinsic Data Dimension}
    
    \authorblock{
      \href{https://alecxu00.github.io}{\textbf{Alec S. Xu}}\corrauth,
      \href{https://canyaras.com}{\textbf{Can Yaras}},
      \href{https://peng8wang.github.io}{\textbf{Peng Wang}}\textsuperscript{1},
      \href{https://qingqu.engin.umich.edu/}{\textbf{Qing Qu}}
    }

    \affiliation{University of Michigan \quad \quad \textsuperscript{1}University of Macao}

    \authornote{\corrauth\ Corresponding author}

    \abstracttext{Deep neural networks have attained remarkable success across diverse classification tasks. Recent empirical studies have shown that deep networks learn features that are linearly separable across classes. However, these findings often lack rigorous justifications, even under relatively simple settings. In this work, we address this gap by examining the linear separation capabilities of shallow nonlinear networks. Specifically, inspired by the low intrinsic dimensionality of image data, we model inputs as a union of low-dimensional subspaces (UoS) and demonstrate that a single nonlinear layer can transform such data into linearly separable sets. Theoretically, we show that this transformation occurs with high probability when using random weights and quadratic activations. Notably, we prove this can be achieved when the network width scales polynomially with the intrinsic dimension of the data rather than the ambient dimension. Experimental results corroborate these theoretical findings and demonstrate that similar linear separation properties hold in practical scenarios beyond our analytical scope. This work bridges the gap between empirical observations and theoretical understanding of the separation capacity of nonlinear networks, offering deeper insights into model interpretability and generalization.}
    

    \date{\today}
    \correspondence{\href{mailto:alecx@umich.edu}{alecx@umich.edu}}
    \resources{\href{https://github.com/alecxu00/uos-linear-separability}{Code}}

    \headerlogo{Deepthink_landscape_do_not_delete_compressed.png}{https://deepthink-umich.github.io}
\fi

\ifsimods
    \documentclass[review,onefignum,onetabnum]{siamonline220329}
    
\usepackage{lipsum}
\usepackage{amsfonts}
\usepackage{graphicx}
\usepackage{epstopdf}
\usepackage{algorithmic}
\ifpdf
  \DeclareGraphicsExtensions{.eps,.pdf,.png,.jpg}
\else
  \DeclareGraphicsExtensions{.eps}
\fi

\newsiamremark{remark}{Remark}
\newsiamremark{hypothesis}{Hypothesis}
\crefname{hypothesis}{Hypothesis}{Hypotheses}
\newsiamthm{claim}{Claim}

\headers{Understanding How Nonlinear Networks Create Linearly Separable Features for Low-Dimensional Data}{A. S. Xu, C. Yaras, P. Wang, Q. Qu}

\title{Understanding How Nonlinear Networks Create Linearly Separable Features for Low-Dimensional Data\thanks{Submitted to the editors \today.
\funding{This work was funded by NSF CAREER CCF-2143904, NSF IIS 2312842, NSF IIS 2402950, ONR N00014-22-1-2529, and a gift grant from KLA.}}}

\author{ Alec S. Xu\thanks{Electrical Engineering and Computer Science, University of Michigan, Ann Arbor, MI, USA.
  (\email{alecx@umich.edu},
  \email{cjyaras@umich.edu},
  \email{pengwa@umich.edu},
  \email{qingqu@umich.edu}).}
\and Can Yaras\footnotemark[2] 
\and Peng Wang\footnotemark[2] 
\and Qing Qu\footnotemark[2]  }

\usepackage{amsopn}

    \usepackage[utf8]{inputenc} 
\usepackage[T1]{fontenc}    
\usepackage{url}
\usepackage{hyperref}
\usepackage{amsmath,amssymb,amsbsy,amsfonts,amscd,bm}
\usepackage{paralist}
\usepackage{xcolor}
\usepackage{xspace}
\usepackage{color}
\usepackage{graphicx}
\graphicspath{{./figs/}}
\usepackage{algorithm}
\usepackage{algorithmic}
\usepackage{tikz}
\usepackage{tikz-3dplot}
\usetikzlibrary{quotes, angles}
\usepackage{comment}
\usepackage{multirow}
\usepackage{enumitem}
\usepackage{fancyhdr}
\usepackage{cleveref}
\usepackage{subcaption}
\usepackage{wrapfig}
\usepackage[toc, page]{appendix}
\usepackage[numbers]{natbib} 
\usepackage{bbm}
\usepackage{tcolorbox}

\setlist[enumerate]{leftmargin=.5in}
\setlist[itemize]{leftmargin=.5in}

\newtheorem{problem}{Problem}
\newtheorem{assum}{Assumption}
\newcommand{\mtx}[1]{\boldsymbol{#1}}

\newboolean{twoColVersion}
\setboolean{twoColVersion}{false}
\newcommand{\twoCol}[2]{\ifthenelse{\boolean{twoColVersion}} {#1} {#2} }

\newcommand{\calA}{\mathcal{A}}

\newcommand{\calI}{\mathcal{I}}
\newcommand{\calJ}{\mathcal{J}}

\newcommand{\calN}{\mathcal{N}}

\newcommand{\calR}{\mathcal{R}}
\newcommand{\calS}{\mathcal{S}}

\newcommand{\calX}{\mathcal{X}}

\newcommand{\mC}{\mtx{C}}

\newcommand{\mS}{\mtx{S}}

\newcommand{\xmath}[1] {\ensuremath{#1}\xspace}
\newcommand{\blmath}[1] {\xmath{\bm{#1}}}
\newcommand{\A}{\blmath{A}}
\newcommand{\B}{\blmath{B}}
\newcommand{\D}{\blmath{D}}

\newcommand{\I}{\blmath{I}}
\newcommand{\Sb}{\blmath{S}}

\newcommand{\Q}{\blmath{Q}}
\newcommand{\U}{\blmath{U}}
\newcommand{\V}{\blmath{V}}
\newcommand{\W}{\blmath{W}}
\newcommand{\X}{\blmath{X}}
\newcommand{\Y}{\blmath{Y}}

\newcommand{\ab}{\blmath{a}}
\newcommand{\blb}{\blmath{b}}

\newcommand{\w}{\blmath{w}}
\newcommand{\x}{\blmath{x}}

\newcommand{\y}{\blmath{y}}
\newcommand{\z}{\blmath{z}}

\newcommand{\0}{\blmath{0}} 

\newcommand{\ie}{\text{i.e.}}
\newcommand{\eg}{\text{e.g.}}
\renewcommand{\u} {\blmath{u}}
\renewcommand{\v} {\blmath{v}}

\def \reals   { \mathbb{R} }
\newcommand{\eE}{\xmath{\mathbb{E}}}
\newcommand{\rank}{\mathrm{rank}}
\newcommand{\mSigma}{\mtx{\Sigma}}

\newcommand{\exprm}{\mathrm{exp}}
\newcommand{\trrm}{\mathrm{Tr}}

\newcommand{\varrm}{\mathrm{Var}}
\newcommand{\covrm}{\mathrm{Cov}}
\newcommand{\maxrm}{\mathrm{max}}
\newcommand{\minrm}{\mathrm{min}}

    \ifpdf
    \hypersetup{
      pdftitle={Understanding How Nonlinear Networks Create Linearly Separable Features for Low-Dimensional Data},
      pdfauthor={A. S. Xu, C. Yaras, P. Wang, and Q. Qu}
    }
    \fi
\fi

\ifaistats
    \documentclass[twoside]{article}
    \usepackage[accepted]{aistats2026}
    \usepackage[utf8]{inputenc} 
\usepackage[T1]{fontenc}    
\usepackage{url}
\usepackage{xr-hyper}
\usepackage{hyperref}
\usepackage{amsmath,amssymb,amsbsy,amsfonts,amscd,bm,amsthm}
\usepackage{paralist}
\usepackage{xcolor}
\usepackage{xspace}
\usepackage{color}
\usepackage{graphicx}
\graphicspath{{./figs/}}
\usepackage{algorithm}
\usepackage{tikz}
\usepackage{tikz-3dplot}
\usetikzlibrary{quotes, angles}
\usepackage{comment}
\usepackage{multirow}
\usepackage{enumitem}
\usepackage{fancyhdr}
\usepackage{cleveref}
\usepackage{subcaption}
\usepackage{wrapfig}
\usepackage[toc, page]{appendix}
\usepackage{bbm}
\usepackage{tcolorbox}

\setlist[enumerate]{leftmargin=.5in}
\setlist[itemize]{leftmargin=.5in}

\newtheorem{problem}{Problem}
\newtheorem{assum}{Assumption}
\newtheorem{definition}{Definition}
\newtheorem{theorem}{Theorem}
\newtheorem{corollary}{Corollary}
\newtheorem{lemma}{Lemma}
\newcommand{\mtx}[1]{\boldsymbol{#1}}

\newboolean{twoColVersion}
\setboolean{twoColVersion}{false}
\newcommand{\twoCol}[2]{\ifthenelse{\boolean{twoColVersion}} {#1} {#2} }

\newcommand{\calA}{\mathcal{A}}

\newcommand{\calI}{\mathcal{I}}
\newcommand{\calJ}{\mathcal{J}}

\newcommand{\calN}{\mathcal{N}}

\newcommand{\calR}{\mathcal{R}}
\newcommand{\calS}{\mathcal{S}}

\newcommand{\calX}{\mathcal{X}}

\newcommand{\mC}{\mtx{C}}

\newcommand{\mS}{\mtx{S}}

\newcommand{\xmath}[1] {\ensuremath{#1}\xspace}
\newcommand{\blmath}[1] {\xmath{\bm{#1}}}
\newcommand{\A}{\blmath{A}}
\newcommand{\B}{\blmath{B}}
\newcommand{\D}{\blmath{D}}

\newcommand{\I}{\blmath{I}}
\newcommand{\Sb}{\blmath{S}}

\newcommand{\Q}{\blmath{Q}}
\newcommand{\U}{\blmath{U}}
\newcommand{\V}{\blmath{V}}
\newcommand{\W}{\blmath{W}}
\newcommand{\X}{\blmath{X}}
\newcommand{\Y}{\blmath{Y}}

\newcommand{\ab}{\blmath{a}}
\newcommand{\blb}{\blmath{b}}

\newcommand{\w}{\blmath{w}}
\newcommand{\x}{\blmath{x}}

\newcommand{\y}{\blmath{y}}
\newcommand{\z}{\blmath{z}}

\newcommand{\0}{\blmath{0}} 

\newcommand{\ie}{\text{i.e.}}
\newcommand{\eg}{\text{e.g.}}
\renewcommand{\u} {\blmath{u}}
\renewcommand{\v} {\blmath{v}}

\def \reals   { \mathbb{R} }
\newcommand{\eE}{\xmath{\mathbb{E}}}
\newcommand{\rank}{\mathrm{rank}}
\newcommand{\mSigma}{\mtx{\Sigma}}

\newcommand{\exprm}{\mathrm{exp}}
\newcommand{\trrm}{\mathrm{Tr}}

\newcommand{\varrm}{\mathrm{Var}}
\newcommand{\covrm}{\mathrm{Cov}}
\newcommand{\maxrm}{\mathrm{max}}
\newcommand{\minrm}{\mathrm{min}}

    \usepackage[sort]{natbib} 
\fi

\begin{document}

\ifarxiv
    \makeDeepthinkHeader
    \begin{figure*}[h]
        \centering
        \includegraphics[width=0.33\linewidth]{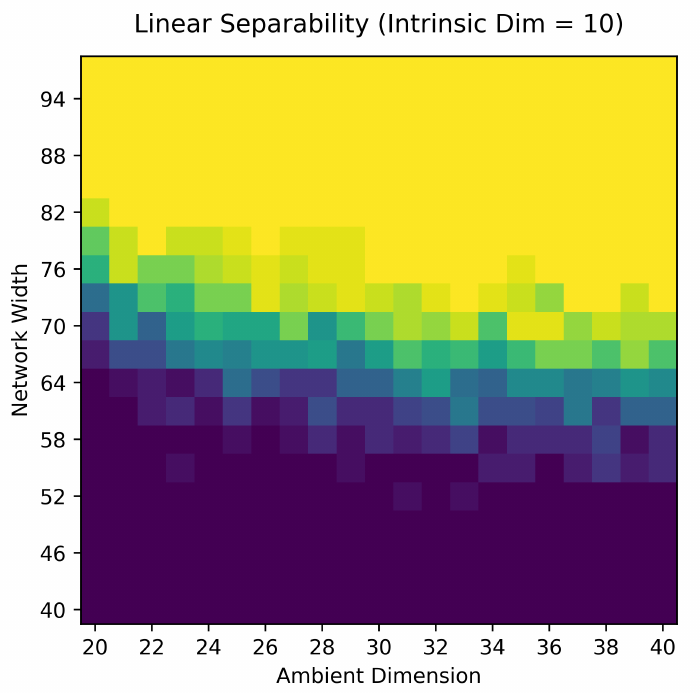}
        \hspace{0.1in}
        \includegraphics[width=0.363\linewidth]{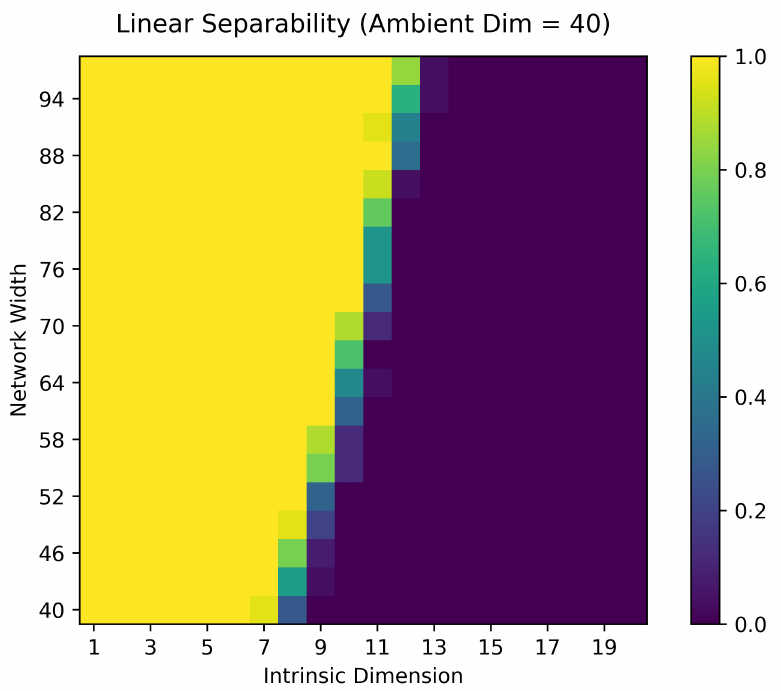}
        \caption{\textbf{Phase transition of linear separability w.r.t. ambient (left) and intrinsic (right) data dimensions vs. network width.} The network width that makes a union of two subspaces linearly separable only scales with the intrinsic dimension. See \Cref{ssec:phase-transition} for experimental details.}  
        \label{fig:separability-d-r}
    \end{figure*}
    
    \newpage 
    
    \tableofcontents 
    \section{Introduction} \label{sec:intro}
Over the past decade, deep neural networks (DNNs) have achieved state-of-the-art performance in a wide range of applications, including computer vision \citep{simonyan2015very, he2016deep} and natural language processing \citep{sutskever2014sequence, vaswani2017attention}. However, despite recent advances \citep{jacot2018neural, mei2018mean, ji2019gradient, arora2018convergence, lampinen2019analytic, papyan2020prevalence, zhu2021geometric,yaras2022neural,zhou2022optimization}, the
theoretical understanding of their empirical success is still primitive, even for relatively basic tasks. For example, in classification problems, the success of deep learning is often attributed to its ability to learn discriminative features that exhibit strong inter-class separation \citep{papyan2020prevalence, alain2017understanding, rangamani2023feature, masarczyk2024tunnel,wang2025understanding,yaras2023law}. 
Despite the remarkable ability of deep networks to achieve linear separation, the underlying mechanisms by which they accomplish this—especially when the input data are initially poorly separated—remain largely unclear. Investigating this phenomenon could significantly improve the interpretability of deep learning models and provide deeper insights into their generalization capabilities. Before presenting our main contribution, we provide a brief review of the existing results --- see \Cref{app:related} for a more detailed discussion. 

\paragraph{Empirical studies on linear separability of early-layer features.} 
Recent empirical studies investigated the role of the intermediate layers in deep nonlinear networks, \eg, \cite{alain2017understanding, ansuini2019intrinsic, recanatesi2019dimensionality, he2023law,zhang2022all,wang2025understanding,yaras2023law,masarczyk2024tunnel,li2024understanding}. These studies indicate that the shallow layers expand the features such that they become linearly separable between classes. 
For instance, in image classification, \cite{alain2017understanding, masarczyk2024tunnel, wang2025understanding} observed linear probing accuracy improves significantly across the early layers of neural networks. This implies the early layers play a critical role in achieving linear separability of the input data.

\paragraph{Theoretical works on linear separability of early-layer features.} To our knowledge, there are limited theoretical studies on the linear separability of features across nonlinear layers in DNNs. Recent works \citep{dirksen2022separation, ghosal2022randomly} studied the separability of features in shallow ReLU networks. These studies rigorously showed the features extracted from a two-layer \citep{dirksen2022separation} and one-layer \citep{ghosal2022randomly} random ReLU network are linearly separable for arbitrary input data. However, a key limitation of these works is that in the worst case, the required network width grows \emph{exponentially} with respect to (w.r.t.) the ambient dimension of the data. Consequently, the network sizes required by theoretical analyses are substantially larger than those typically used in real-world applications, highlighting a gap between theory and practice.

\paragraph{Theoretical studies on representation learning in deep linear networks.} Another line of research has explored how deep \emph{linear} networks (DLNs) progressively compress within-class features and discriminate between-class features \citep{saxe2019mathematical, wang2025understanding}. Building on the empirical observation that linear layers can emulate the behavior of deeper layers in nonlinear networks, \cite{wang2025understanding} provided a theoretical analysis of the progressive feature compression in DLNs, under the assumption that the input data are already linearly separable. However, due to this restrictive assumption, the study cannot fully explain the structures of hierarchical representation in nonlinear networks, particularly \emph{how} the early layers transform input features to achieve linear separability due to the nonlinear operators.

\subsection{Our Contributions} \label{ssec:contributions}
In this work, we investigate the linear separability of features in shallow nonlinear networks for data with low intrinsic dimensions.
Specifically, we show
\begin{tcolorbox}[colframe = red!75!black]
\begin{center}
   \emph{a single nonlinear layer with random weights 
   transforms data from a union of low-dimensional subspaces into linearly separable sets.}
\end{center}
\end{tcolorbox}
\noindent We rigorously prove this result with $K = 2$ subspaces and discuss how the result can be extended to $K > 2$ subspaces. 
In our analysis, we assume that the activation is quadratic and the first-layer weights are random. The resulting width of the network scales \emph{polynomially} w.r.t. the intrinsic dimension of the subspaces. Moreover, our results empirically hold under more generic settings. For example, we can replace the quadratic with other activations, such as ReLU, and still achieve linear separability with similar requirements on the subspace dimensions and number of subspaces  
(see \Cref{fig:rank-K-sweep}). Our findings offer insights into the role of overparameterization in deep representation learning and explain why learning based upon random features can lead to good in-distribution generalization.

\subsection{Notation and Paper Organization} Before delving into the technical discussion, we introduce the notation used throughout the paper and outline its organization.

\paragraph{Notation.} For a positive integer $N$, we use $[N]$ to denote the index set $\{1, 2, \dots, N\}$. We use $\calN(\mu, \sigma^2)$ to denote a Gaussian distribution with mean $\mu$ and variance $\sigma^2$, and $\calN(\bm{\mu}, \bm{\Sigma})$ to denote a multivariate Gaussian distribution with mean $\bm{\mu}$ and covariance $\bm{\Sigma}$. We use $\| \cdot \|$ to denote the Euclidean norm of a vector, $\bm 0_m$ to denote an $m$-dimensional vector of all zeros, $\lambda_i(\cdot)$ to denote the $i^{th}$ largest eigenvalue of a symmetric matrix, and $\sigma_i(\cdot)$ to denote the $i^{th}$ largest singular value of a matrix. With a slight abuse of notation, for some function $\phi$ and set $\calX$, $\phi \big( \calX \big)$ denotes the set $\big\{ \x \in \calX: \phi \big( \x \big) \big\}$. Unless otherwise stated, the term ``subspace'' implies a linear subspace embedded in Euclidean space.

\paragraph{Organization.} The rest of this paper is organized as follows. We motivate the union of subspaces (UoS) 
data model, introduce our problem setting, and motivate our theoretical assumptions in \Cref{sec:prelim}. 
We then state our main theoretical results and provide a proof sketch in \Cref{sec:theoretical}, with the full proof in \Cref{app:thm-1-proof}. In \Cref{sec:empirical}, we provide empirical evidence supporting our theoretical results, and investigate settings not considered in our analysis. 
Finally, we summarize our results and conclude in \Cref{sec:conclusion}.
    \section{Preliminaries} \label{sec:prelim}
In this section, we introduce the basic problem setup and motivations. First, we introduce the UoS model for our input data in \Cref{ssec:uos}, and then discuss the choices of the network in \Cref{ssec:problem}.

\subsection{Assumptions on Input Data} \label{ssec:uos}

\begin{figure}
    \centering
    \includegraphics[width=\linewidth]{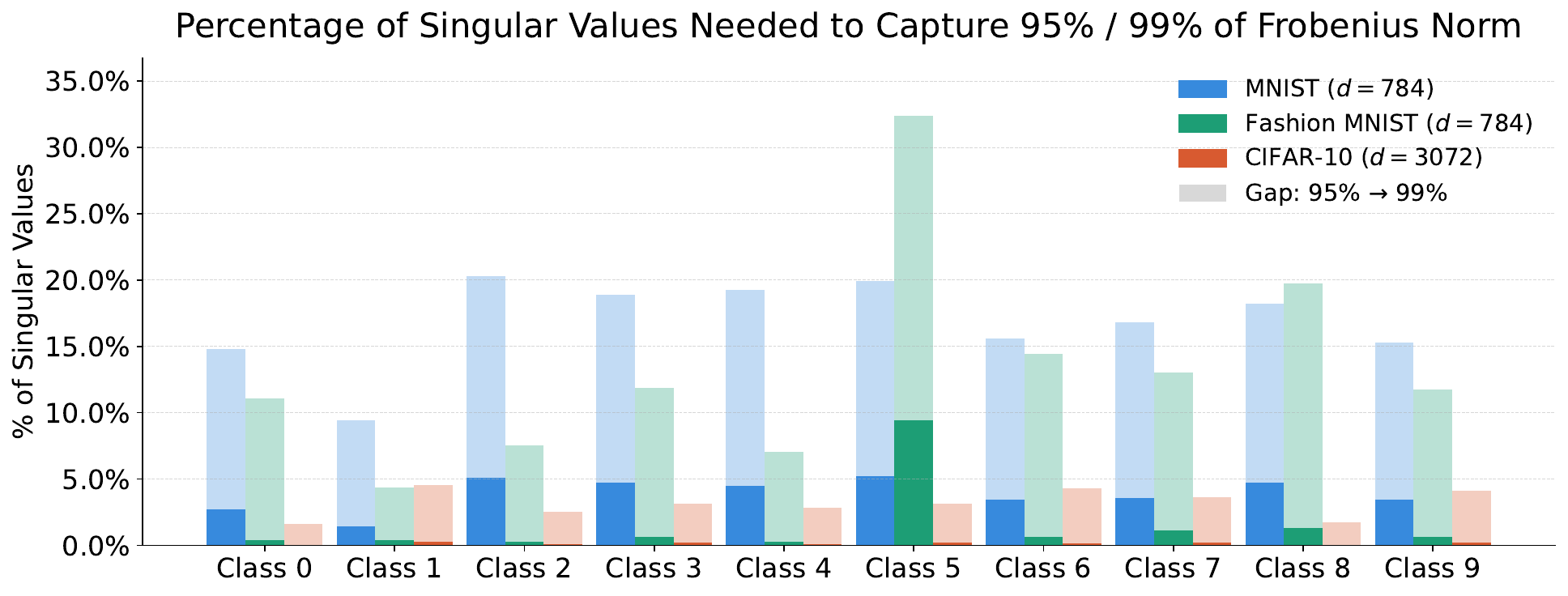}
    \caption{\textbf{In common image datasets, a small proportion of singular values in each class's data matrix captures a large majority of the Frobenius norm.} The darker bars indicate the proportion of singular values needed to reach $95\%$ of that class data matrix's Frobenius norm, while the lighter bars indicate the proportion needed to reach $99\%$. For MNIST, Fashion MNIST, and CIFAR-10 respectively, about the first $15 - 20\%$ (about $115 - 155$ out of $784$), $5 - 10\%$ (about $40 - 80$ out of $784$), and $2 - 4\%$ (about $60 - 120$ out of $3072$) of the singular values account for $99\%$ of most class data matrix's Frobenius norms --- the lone outlier is class $5$ in Fashion MNIST. This implies each class \emph{approximately} lies on its own low-dimensional subspace. In other words, \textbf{these datasets approximately satisfy the UoS data model.} See \Cref{ssec:image-data-svals} for experimental details.}
    \label{fig:image_data_class_svals}
\end{figure}

Recent empirical studies indicate real-world image data typically possess a significantly lower \emph{intrinsic} dimension than their ambient dimension. For instance, \cite{pope2020intrinsic} used a nearest-neighbor approach to estimate the intrinsic dimension of many popular image datasets, including MNIST \citep{lecun1998gradient}, CIFAR-10 \citep{krizhevsky2009learning}, and ImageNet \citep{russakovsky2015imagenet}. They showed the intrinsic dimension of these datasets is at most around $40$, even though the images themselves contain thousands of pixels. Furthermore, \cite{brown2023verifying} used a similar approach to show \emph{each class} has its own low intrinsic dimensionality. These results indicate image data lie on  \emph{a union of low-dimensional manifolds} within high-dimensional space. 

Although low-dimensional manifolds can exhibit complex structures, each manifold can be locally approximated by its tangent space, which is a linear subspace embedded within the ambient space. This motivates us to initiate our study with a simplified model: \textbf{a union of $K$ low-dimensional subspaces (UoS)} that capture the local structures of manifolds. Similar models have recently been explored for understanding generative models \citep{wang2024diffusion,chen2024exploring}. Furthermore, \Cref{fig:image_data_class_svals} shows common image datasets \emph{approximately} lie on a UoS, where each subspace represents a different class. 

For ease of analysis and exposition, we focus on the case where $K = 2$. Nonetheless, our results extend to the case where $K > 2$, as discussed in \Cref{ssec:multiple}. To set the stage for our analysis, we introduce a generic definition of a union of $K$ subspaces.

\begin{tcolorbox}
\begin{definition}[Union of $K$ Low-Dimensional Subspaces] \label{def:UoS}
    Let $\calS_1, \calS_2, \dots \calS_K \subseteq \reals^d$ be $K$ linear subspaces with dimensions $r_1, r_2, \dots, r_K$, respectively. Let $\U_k \in \mathbb{R}^{d\times r_k}$ be an orthonormal basis of $\calS_k$ for all $k\in [K]$. 
    The union of subspaces $\calS_1, \calS_2, \dots, \calS_K$ is defined as such:
    \begin{align*}
        \bigcup\limits_{k=1}^K \calS_k := \Big\{\z \in \reals^d: \exists k \in [K], \bm \alpha \in \mathbb{R}^{r_k} \: \text{s.t.} \: \z = \U_k \bm \alpha \Big\} .
    \end{align*}
\end{definition} 
\end{tcolorbox}

The \emph{principal angles} between two subspaces is a generalization of the angles between two vectors (i.e., two one-dimensional subspaces). For two subspaces $(\calS_1,\calS_2)$ of dimensions $r_1$ and $r_2$, there exist $\minrm\{r_1, r_2\}$
principal angles between them. These angles are formally defined as follows.
\begin{tcolorbox}
\begin{definition}[Principal angles between two subspaces]
     Suppose that the columns of $\U_1 \in \reals^{d \times r_1}$ and $\U_2 \in \reals^{d \times r_2}$ are orthonormal bases for subspaces $\calS_1$ and $\calS_2$, respectively.  Let $r := \minrm\{r_1, r_2\}$. For all $\ell \in [r]$, the $\ell^{th}$ principal angle $\theta_\ell \in [0, \pi/2]$ between $\calS_1$ and $\calS_2$ is defined as 
    \begin{equation*}
        \cos(\theta_\ell) \;:=\; \sigma_\ell(\U_1^\top \U_2).
    \end{equation*}
\end{definition}
\end{tcolorbox}

\begin{wrapfigure}[8]{r}{0.35\textwidth} 
    \centering
    \tdplotsetmaincoords{120}{50}
    \begin{tikzpicture}[scale=1.75]
    \tdplotsetrotatedcoords{90}{0}{0}
    \fill[blue!70, opacity=0.5, tdplot_rotated_coords] (-1, -0.5, 0) -- (1, -0.5, 0) -- (1, 0.5, 0) -- (-1, 0.5, 0) -- cycle; 
    
    \draw[<-, thick, orange] (-0.5, -0.75) -- (-0.17, -0.255);
    \draw[-, thick, orange, opacity=0.3] (-0.17, -0.255) -- (0, 0); 
    
    \draw[->, thick, orange] (0, 0) -- (0.5, 0.75); 
    \draw[<->, densely dotted, black] (-1, 0.45) -- (1, -0.45);
    \draw[<->, densely dotted, black] (0, 1) -- (0, -1);
    \draw[<->, densely dotted, black] (-0.75, -0.6) -- (0.75, 0.6);

    
    \node[fill=black, circle, inner sep=1pt, opacity=0.5] at (0, 0, 0) {};
    
    \coordinate (a) at (0.3, 0.45);
    \coordinate (o) at (0, 0);
    \coordinate (b) at (0.3, -0.15);
    \pic[draw, <->, "$\theta_1$", angle eccentricity=1.5]{angle = b--o--a};
    
    \node[anchor=south] at (0.6, 0.7) {$\calS_1$};
    \node[anchor=west] at (0.5, -0.6) {$\calS_2$};
    
    \end{tikzpicture}
    \caption{\textbf{The principal angle between a one-dimensional subspace $\mathcal S_1$ and two-dimensional subspace $\mathcal S_2$.}}
    \label{fig:princ-angles}
\end{wrapfigure}
\noindent The principal angle is illustrated in \Cref{fig:princ-angles}. By the above definition, since $0 \leq \theta_1 \leq \theta_2 \leq \dots \leq \theta_r \leq \pi/2$, we will sometimes use $\theta_{\min}$ to denote $\theta_1$. 
Building on these definitions, we will make the following assumption on the UoS model for our analysis in \Cref{sec:theoretical}.
\begin{tcolorbox}
\begin{assum} \label{assum:subspaces}
    There are $K = 2$ subspaces $\calS_1, \calS_2$ with equal dimensions, \ie, $r_1 = r_2 := r$. Furthermore, the principal angles between $\calS_1$ and $\calS_2$ are strictly positive, \ie, $0 < \theta_1 \leq \theta_2 \leq \dots \leq \theta_r \leq \pi/2$.
\end{assum}
\end{tcolorbox}

\noindent We discuss \Cref{assum:subspaces} below.

\paragraph{Number of subspaces.} We assume $K=2$ subspaces to simplify both the analysis and exposition. The results can be generalized to consider $K > 2$ subspaces, which we discuss in detail in \Cref{ssec:multiple}. 

\paragraph{Subspace dimensions.} We assume equal dimensionality for each subspace for simplicity. In practice, each subspace in a UoS can have different dimensions. We believe our result can be generalized to this setting, and leave detailed analysis for future work.

\paragraph{Principal angles between subspaces.} We assume none of the principal angles are equal to zero to ensure $\calS_1 \cap \calS_2 = \{\0_d\}$. Otherwise, it is impossible to label the non-zero points in $\calS_1 \cap \calS_2$. We note $\theta_1 > 0$ if and only if $r < d/2$. This assumption is typically satisfied in practice, as usually $r \ll d$.

\begin{figure}[t]
    \centering
    \tdplotsetmaincoords{120}{50}
    \begin{tikzpicture}[scale=2.0]
    
    \begin{scope}[tdplot_main_coords]
        \tdplotsetrotatedcoords{0}{90}{90}
        \fill[orange!70, opacity=0.5, tdplot_rotated_coords] (-1, -0.4, 0) -- (1, -0.4, 0) -- (1, 0.4, 0) -- (-1, 0.4, 0) -- cycle;
    
        \tdplotsetrotatedcoords{0}{0}{0}
        \fill[blue!70, opacity=0.5, tdplot_rotated_coords] (-1, -0.5, 0) -- (1, -0.5, 0) -- (1, 0.5, 0) -- (-1, 0.5, 0) -- cycle;

        \draw[<->, densely dotted, black] (0, 0, 1) -- (0, 0, -1);
        \draw[<->, densely dotted, black] (0, 1.5, 0) -- (0, -1.5, 0);
        \draw[<->, densely dotted, black] (-1.5, 0, 0) -- (1.5, 0, 0); 
    \end{scope}

    \node[fill=black, circle, inner sep=1pt, opacity=0.5] at (0, 0, 0) {};
    
    \node[anchor=south] at (0.5, 0.4) {$\calS_1$};
    \node[anchor=west] at (0.7, -0.2) {$\calS_2$};
    
    \draw[thick,->] (1.5, 0, 0) -- (2.5, 0, 0) node[midway, above] {$f(\cdot)$};
    
    \begin{scope}[xshift=3.3cm, yshift=-0.3cm, scale=0.7]
        \fill[blue!70, opacity=0.7,scale=1.5] plot[smooth cycle, tension=1] coordinates {(0,0) (0.8,0.4) (1,1) (0.2,1.2) (-0.4,0.8)};
        \node at (0.5, 0.9) {$f(\calS_1)$};
    
        \fill[orange!70, opacity=0.7,scale=1.5,xshift=-0.5cm] plot[smooth cycle, tension=1] coordinates {(2,-0.3) (2.7,0) (2.5,0.8) (1.8,0.6)};
        \node at (2.6, 0.3) {$f(\calS_2)$};
        
        \draw[dashed, thick, xshift=0.9cm, yshift=-0.5cm, rotate=-30] (0, 0) -- (0, 2.5);
    \end{scope}
    
    \end{tikzpicture}
    \caption{\textbf{An illustration of \Cref{prob:binary}.} We aim to find conditions on $f$ so a union of subspaces (left) transforms into linearly separable sets (right).}
    \label{fig:lin-sep}
\end{figure}

\subsection{Linear Separability of UoS via Nonlinear Networks} \label{ssec:problem} 


In this work, we investigate how nonlinear neural networks separate the data that follows the UoS model.  Specifically, we consider a shallow neural network $f_{\W}(\bm x): \mathbb R^d \mapsto \mathbb R^D$, which is a feature mapping from the input space $\mathbb R^d$ to a feature space $\mathbb R^D$:
\begin{align}\label{eq:func-NN}
    f_{\W}(\bm x) = \sigma (\bm W \bm x ).
\end{align}
Here, $\W \in \reals^{D \times d}$ is the weight matrix, and $\sigma(\cdot)$ is an entry-wise nonlinear activation function. 
As illustrated in \Cref{fig:lin-sep}, based on the above setup, we are interested in the following problem:
\begin{problem} \label{prob:binary}
   Consider a union of two subspaces $\calS_1$ and $\calS_2$ that satisfy \Cref{assum:subspaces}. Under what conditions does there exist a separating hyperplane $\v \in \reals^D$ such that
    \begin{equation} \label{eq:lin-sep-problem}
        \v^\top f_{\W}\big( \U_1 \bm \alpha \big) > 0 \; \; \text{and} \; \; \v^\top f_{\W}\big( \U_2 \bm \alpha \big) < 0
    \end{equation}
    for all $\bm \alpha \in \reals^r \setminus \{\0_r\}$? 
\end{problem}




\smallskip

\noindent 
\Cref{prob:binary} remains challenging even under the simplified setup considered here. Generally, data from two distinct subspaces are not inherently linearly separable, and neither a linear mapping nor a nonlinear activation alone are sufficient to transform such data into linearly separable sets --- see \Cref{app:problem} for a more detailed discussion. Thus, to tackle \Cref{prob:binary}, we must \emph{jointly} apply a linear mapping and a nonlinear transformation to achieve linear separability of the subspaces. Specifically, we now introduce the following assumptions on the network \eqref{eq:func-NN}, based upon which we characterize the sufficient conditions for achieving linear separability in \Cref{sec:theoretical}.
\begin{tcolorbox}
\begin{assum} \label{assum:network}
    For the mapping $f_{\W}(\bm x) $ in \eqref{eq:func-NN}, we assume that the activation function $\sigma(\cdot)$ is the quadratic (entry-wise square) function, and the entries of $\W$ are independent and identically distributed (iid) standard Gaussian, \ie, $W_{ij} \overset{\text{iid}}{\sim} \calN(0, 1)$ for all $(i, j) \in [D] \times [d]$. 
\end{assum}
\end{tcolorbox}

\noindent 
We briefly discuss \Cref{assum:network} below.

\vspace{-0.3cm}

\paragraph{Quadratic activation.} In this work, we consider the quadratic activation due to its smoothness and simplicity. Such activations have also been considered in many previous theoretical results of analyzing nonlinear networks, \eg, \cite{li2018algorithmic, soltanolkotabi2018theoretical, du2018power,  sarao2020optimization,gamarnik2024stationary}  --- see \Cref{app:related} for a more detailed discussion. Moreover, we believe the results approximately hold for several other nonlinear activations, such as ReLU. In \Cref{sec:empirical}, we empirically show if one replaces the quadratic activation with other activations, the output features from \eqref{eq:func-NN} are still linearly separable under a UoS data model. We also observe the required width to achieve linear separability scales similarly with the intrinsic dimension and the number of classes under both ReLU and quadratic activations --- see \Cref{fig:rank-K-sweep}. 

\vspace{-0.3cm}

\paragraph{Random weights.} \Cref{assum:network}  yields a random feature model, which has been widely studied in the literature, \eg, \citep{rahimi2007random, rahimi2008weighted, rudi2017generalization, bach2017equivalence, li2021towards} (see \citep{liu2021random} for a survey). 
    Moreover, it can also shed light on trained DNNs. For example, in the infinite-width limit \citep{jacot2018neural, arora2019exact, cao2019generalization, allen2019convergence} 
    random networks behave similarly to fully-trained networks. This is called the Neural Tangent Kernel (NTK) \citep{jacot2018neural} regime, where the random initialization determines the NTK, which remains constant during training \citep{jacot2018neural}. Furthermore, the Neural Network Gaussian Process kernel (NNGP) \citep{lee2018deep} is the kernel associated with a network at random initialization. Recently, \citep{kothapalli2024kernel} studied Neural Collapse (NC) \citep{papyan2020prevalence} of nonlinear networks from a kernel perspective. They showed that NNGP and NTK exhibited similar amounts of NC.

    Additionally, for finite-width networks, we empirically observe if the initial-layer features under a UoS data model are linearly separable at random initialization, pushing the layer weights away from their randomly initialized values via training does \emph{not} impact the linearly separability of these features --- see \Cref{fig:linear_probe_depth_3}. Thus, studying the linear separability of the features from random layers provides insight into the linear separability of the features from trained layers in finite-width networks. 

\begin{figure*}[t]
    \centering
    \begin{subfigure}[t]{0.49\textwidth}
        \centering
        \includegraphics[width=\linewidth]{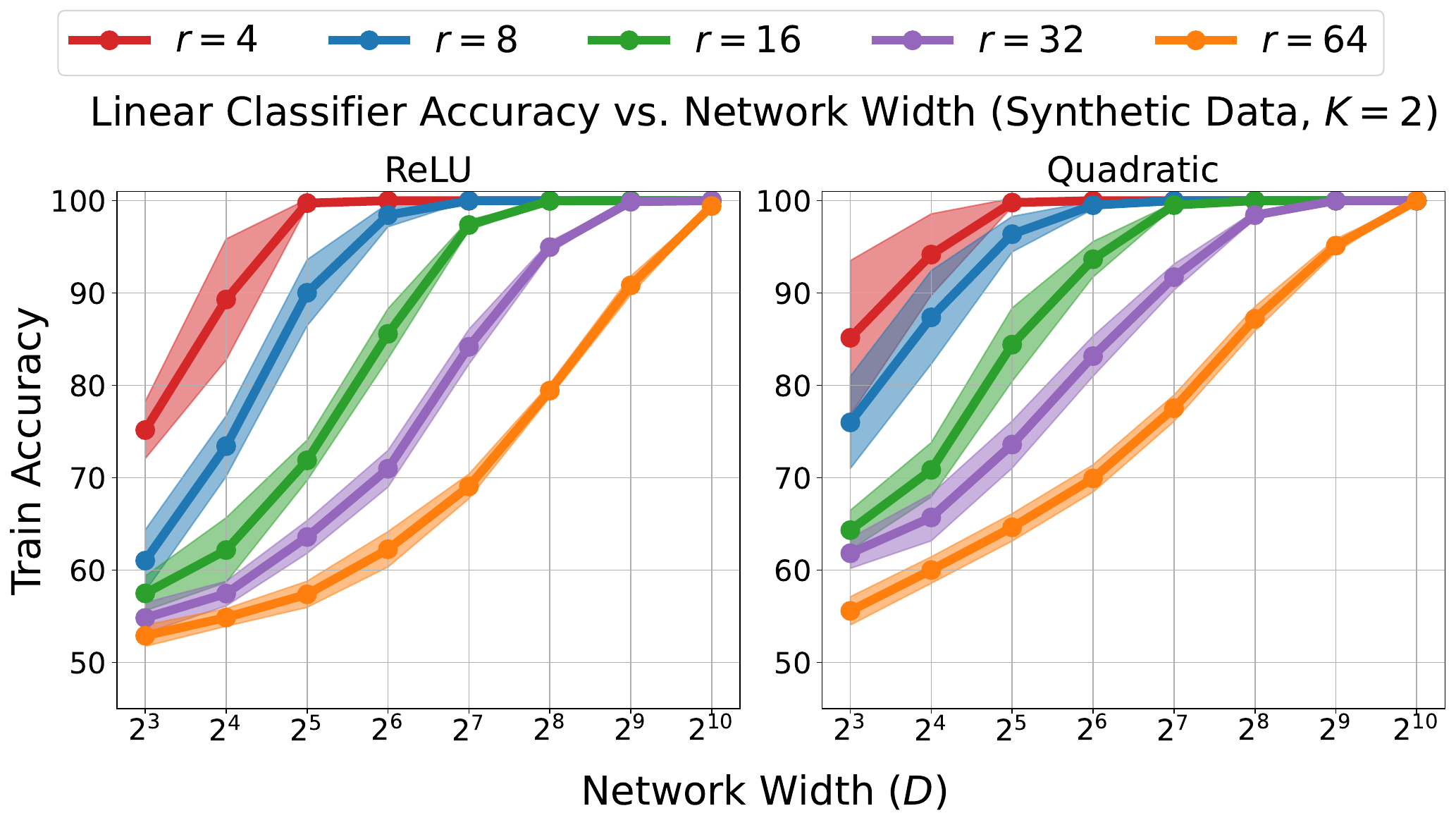}
        \caption{Sweeping $r \in \{4, 8, 16, 32, 64\}$.}
        \label{subfig:rank-sweep}
    \end{subfigure}\hfill
    \begin{subfigure}[t]{0.49\textwidth}
        \centering
        \includegraphics[width=\linewidth]{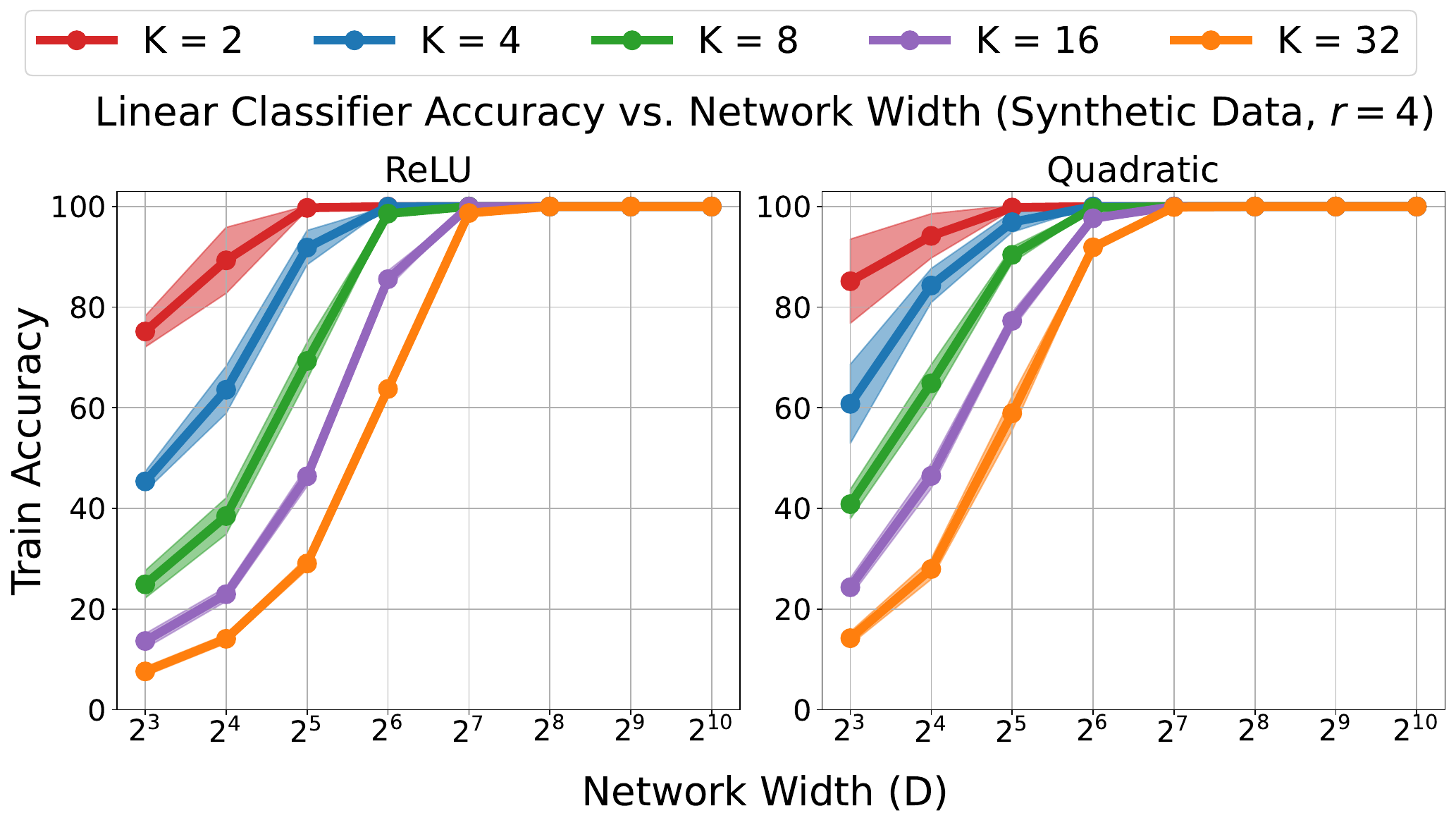}
        \caption{Sweeping $K \in \{2, 4, 8, 16, 32\}$.}
        \label{subfig:K-sweep}
    \end{subfigure}
    \caption{\textbf{ReLU vs. quadratic layers for linear separability.} ReLU and quadratic activations exhibit similar width requirements w.r.t. the intrinsic dimension (left) and number of subspaces (right) for achieving linear separability. See \Cref{sapp:quad-relu-comp} for details.} 
    \label{fig:rank-K-sweep}
\end{figure*}

\begin{figure}[t]
    \centering
    \includegraphics[width=0.7\linewidth]{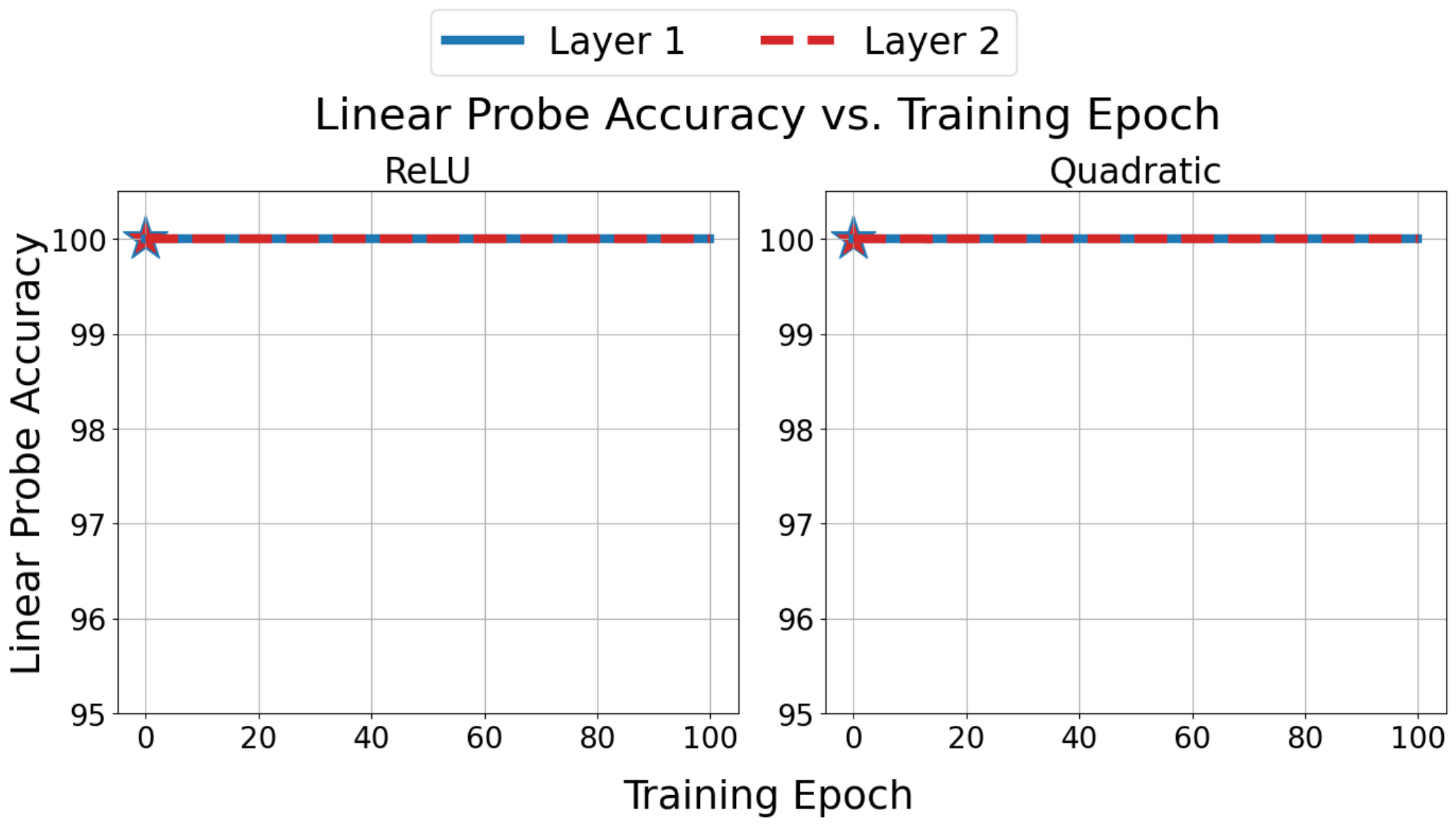}
    \caption{\textbf{Linear separability of hidden-layer features throughout training.} If the initial-layer features are linearly separable at random initialization, they remain linearly separable throughout training. See \Cref{sapp:random_trained_lin_sep} for experimental details.}
    \label{fig:linear_probe_depth_3}
\end{figure}

\section{Theoretical Results} \label{sec:theoretical}
We first state our main theoretical results and their implications in \Cref{ssec:theorem,ssec:multiple}, and correspondingly provide a sketch of the proof in \Cref{ssec:proof-sketch}.



\subsection{Main Results: $K = 2$ Subspaces} \label{ssec:theorem}
First, we state our main theoretical result in the binary case $K=2$. 
\begin{tcolorbox}
\begin{theorem}[Linear Separability of $f(\calS_1)$ and $f(\calS_2)$] \label{thm:binary-lin-sep}
    Suppose Assumptions~\ref{assum:subspaces} and \ref{assum:network} hold, and let $\delta \in (0, 1)$. If the network width $D$ satisfies
    \begin{equation} \label{eq:D-bound}
       D \geq \mathcal{O}\left( \frac{r^3}{\sin^2(\theta_{min})} \cdot \log\left(\frac{r}{\delta}\right) \right), 
    \end{equation}
    then $f(\calS_1)$ and $f(\calS_2)$ are linearly separable with probability at least $1 - \delta$ w.r.t. the randomness of $\W$.
\end{theorem}
\end{tcolorbox}
\Cref{thm:binary-lin-sep} states that the random feature model in \eqref{eq:func-NN} transforms a two subspaces into linearly separable sets, given that the network width scales \emph{polynomially} with the subspaces' intrinsic dimension. We discuss the implications of our result below.

\smallskip

\noindent \textbf{Network width.} 
Our result indicates fewer neurons are needed to achieve linear separability of early-layer features compared to previous studies. Specifically, \citep{dirksen2022separation,ghosal2022randomly} showed one-layer and two-layer random-ReLU networks make arbitrarily structured, nonlinearly separated classes linearly separable. However, under our data model, their network widths scale \emph{exponentially} with the intrinsic dimension. These scaling requirements are much larger than those used in practical DNNs, limiting their applicability. In contrast, \Cref{thm:binary-lin-sep} requires network widths to scale \emph{polynomially} with the intrinsic dimension, aligning more closely with real-world network sizes. For example,  \Cref{fig:image_data_class_svals} shows the CIFAR-10 dataset \citep{krizhevsky2009learning} approximately satisfies a UoS model, where each class subspace is of rank on the order of $10^1$. Previous results \citep{dirksen2022separation,ghosal2022randomly} require a network width on the order of $\exp(10^1)$, whereas our theorem requires a width on the order of $10^3$, which more closely aligns with network sizes used in practice. Therefore, our results provide a more accurate characterization of how the early layers in practical DNNs make low-dimensional data  linearly separable.

\paragraph{Connection to NTK-based results.} Previous work \cite{du2019gradient,huang2020dynamics} leveraged NTK approaches to show global convergence of gradient descent in overparameterized two-layer networks. In their analyses, they showed sufficiently wide networks remain close to their random initializations during training, which is closely related to our random feature model in \Cref{assum:network}. With $N$ training samples, \cite{du2019gradient, huang2020dynamics} showed for two-layer networks of widths $\mathcal{O}\left( \operatorname{poly}(N) \right)$, gradient descent converges to zero training loss, implying perfect linear classifier accuracy under a classification setting. In comparison, our work and \cite{dirksen2022separation,ghosal2022randomly} consider separating \emph{infinitely many points} in the underlying class sets, which are linear subspaces in our case. Under this setting, results from \cite{du2019gradient,huang2006extreme} require \emph{infinitely} wide layers achieve linear separability. However, our result is \emph{independent} of the number of data points, and only depends polynomially on the intrinsic dimension. 

\paragraph{Overparameterization in representation learning.} DNNs are often \emph{overparameterized}, \ie, the number of parameters is larger than the number of training samples $N$. \Cref{thm:binary-lin-sep} states that with probability at least $1 - \delta$, a layer with $\mathcal{O}\left(dr^3 \cdot \log(1 / \delta) \right)$ parameters transforms two subspaces (so an infinite number of data points) into linearly separable sets. In practice, one has a finite number of data points $N$ to classify. In a finite-data setting where the data points lie on a union of two subspaces, by setting the failure probability $\delta = \frac{1}{N}$, a layer with $\mathcal{O}\left(dr^3 \cdot \log(N)\right)$ parameters correctly classifies the data points with probability at least $1 - \frac{1}{N}$. For large $N$, the number of parameters needed to correctly classify all $N$ data points is much smaller than $N$ itself. This implies an \emph{underparameterized} network suffices in separating the data by class, and that overparameterization may serve other purposes in representation learning, e.g., feature compression \citep{wang2025understanding}.

\paragraph{In-distribution generalization.} Our result also provides insight into in-distribution generalization when learning with random features. \Cref{thm:binary-lin-sep} states a single random nonlinear layer makes \emph{all points} in the two subspaces linearly separable. 
Suppose we have a dataset with $N$ train samples lying on a union of two subspaces, apply the random feature map \Cref{eq:func-NN} with $\mathcal{O}(r^3 \cdot \log(Nr))$ features, and train a linear classifier on the random features to classify the train samples. If the test samples lie in the same subspaces as the train samples, then the classifier will also achieve perfect test accuracy with probability at least $1 - \frac{1}{N}$. 

\subsection{Extension to $K > 2$ Subspaces}\label{ssec:multiple}
We now generalize the result from \Cref{thm:binary-lin-sep} to consider $K > 2$ subspaces.
\begin{tcolorbox}
\begin{corollary} \label{cor:K-lin-sep}
    Suppose there are $K > 2$ subspaces each of dimension $r$, where $(K-1)r < d/2$. For all $k \in [K]$, let $\tilde{\calS}_k \supset \calS_k$ and $\overline{\calS}_k \supset \bigcup_{j \in [K], j \neq k} \calS_j$ be $\tilde{r}$-dimensional subspaces with principal angles $\theta_{k, 1}, \theta_{k, 2}, \dots, \theta_{k, \tilde{r}}$ that satisfy \Cref{assum:subspaces}, where $\tilde{r} := (K-1)r$. Also let \Cref{assum:network} hold and $\delta \in (0,1)$. If the network width $D$ satisfies
    \begin{equation}
        D \geq \max\limits_{k \in [K]} \mathcal{O} \left( \frac{\tilde{r}^3}{\sin^2(\theta_{k, min})} \cdot \log \left( \frac{\tilde{r}}{\delta} \right) \right) 
    \end{equation}
     then for all $k \in [K]$, the sets $f(\calS_k)$ and $f\Big( \bigcup_{j \in [K], j \neq k} \calS_j \Big)$ are linearly separable with probability at least $1 - K\delta$ w.r.t. the randomness of $\W$.
\end{corollary}
\end{tcolorbox}
\Cref{cor:K-lin-sep} states if the layer width scales in polynomial order w.r.t. both the intrinsic dimension \emph{and} the number of subspaces, the nonlinear features are one-vs.-all separable: each individual subspace is separated from \emph{all} of the remaining subspaces. In contrast, \Cref{thm:binary-lin-sep} only depends on the intrinsic dimension, as it only considers the binary subspaces setting. 

\subsection{Proof Sketches} \label{ssec:proof-sketch}
In the following, we first provide a proof sketch of \Cref{thm:binary-lin-sep} for binary subspaces $K=2$, and later we generalize the analysis to multiple subspaces $K>2$.

\paragraph{Proof sketch for \Cref{thm:binary-lin-sep}.} We first provide a proof sketch of \Cref{thm:binary-lin-sep}, defering the full proof to \Cref{app:thm-1-proof}. Let $\X := \W \U_1 \in \reals^{D \times r}$ and $\Y := \W \U_2 \in \reals^{D \times r}$, and let $\x_n, \y_n \in \reals^r$ denote the $n^{th}$ row of $\X$ and $\Y$, respectively, written as column vectors. Note $\x_n = \U_1^\top \w_n$ and $\y_n = \U_2^\top \w_n$, where $\w_n \overset{iid}{\sim} \calN(\0_d, \I_d)$ denotes the $n^{th}$ row in $\W$. First, under \Cref{assum:network}, \eqref{eq:lin-sep-problem} holds if and only if there exists a vector $\v \in \reals^D$ such that
\begin{equation} \label{eq:lin-sep-outer-sum}
    \sum\limits_{n=1}^D v_n \x_n \x_n^\top \succ 0 \; \; \text{and} \; \; \sum\limits_{n=1}^D v_n \y_n \y_n^\top \prec 0.
\end{equation}

Next, we are interested in the \emph{existence} of a hyperplane $\v$ that separates the random features, which is not necessarily a max-margin hyperplane. We choose a linear classifier $\v$ with the following entries:

\smallskip

\begin{center}
    \textit{For all $n \in [D]$, $v_n = \mathrm{sign}\big(\|\x_n\|^2 - \|\y_n\|^2\big)$.}
\end{center}

\smallskip

\noindent This choice of $\v$ is a \emph{projection-based classifier}: the subspace onto which $\w_n$ has the largest projection determines the sign of $v_n$. If $\|\U_1^\top \w_n\|^2 > \|\U_2^\top \w_n\|^2$, then we set $v_n = +1$ to push the inner product $\v^\top f_{\W}(\U_1 \bm \alpha)$ to be ``more positive'' for any $\bm \alpha \in \reals^r$. Likewise,  setting $v_n = -1$ when $\|\U_1^\top \w_n\|^2 < \|\U_2^\top \w_n\|^2$ pushes $\v^\top f_{\W}(\U_2 \bm \alpha)$ to be ``more negative'' for any $\bm \alpha \in \reals^r$. Since $\|\U_k^\top \w_n\|^2 \sim \chi^2_r$ for $k \in \{1, 2\}$, $\|\U_1^\top \w_n\|^2 = \|\U_2^\top \w_n\|^2$ occurs with probability zero. With this choice of $\v$, \eqref{eq:lin-sep-outer-sum} is equivalent to

\begin{align} 
    &\mS_1 := \sum\limits_{i \in \calI} \x_i \x_i^\top - \sum\limits_{j \in \calI^c} \x_j \x_j^\top \succ 0, \;  \text{and} \nonumber \\
    &\mS_2 := \sum\limits_{i \in \calI} \y_i\y_i^\top - \sum\limits_{j \in \calI^c} \y_j \y_j^\top \prec 0, \label{eq:S1-S2}
\end{align}
where $\calI := \{n: v_n = +1\}$ and $\calI^c := \{n: v_n = -1\}$.

We now wish to upper bound the failure probability $P\Big( \mS_1 \not \succ 0 \cup \mS_2 \not \prec 0 \Big) = P\Big(\lambda_r\big( \mS_1 \big) \leq 0 \cup \lambda_1\big( \mS_2 \big) \geq 0 \Big)$. Next, we show $\mS_1$ and $\mS_2$ are sums of sub-exponential random matrices, 
which allows us to use Bernstein's matrix inequality \citep[Theorem 6.2]{tropp2012user} to obtain individual bounds on $P\Big(\lambda_r(\mS_1) \leq 0\Big)$ and $P\Big(\lambda_1(\mS_2) \geq 0\Big)$. Applying the union bound 
by some constant $\delta \in (0, 1)$, and then re-arranging the appropriate terms to lower bound $D$, leads to the result in \Cref{thm:binary-lin-sep}. 

\smallskip

\paragraph{Extension to $K > 2$ subspaces in \Cref{cor:K-lin-sep}.} We now present a proof sketch for \Cref{cor:K-lin-sep}, omitting the full details as it directly follows from an application of \Cref{thm:binary-lin-sep}. Note we assume $(K-1)r < d/2$. This assumption is not very limiting when the number of classes is small, since $K$ and $r$ are typically much smaller than $d$ in practice.

Let $k \in [K]$ be arbitrary and $\overline{\calS}_k := \calR\Big( \begin{bmatrix}
    \U_1 & \U_2 & \dots & \U_{k-1} & \U_{k+1} & \dots & \U_K
\end{bmatrix} \Big)$ be an $\tilde{r}$-dimensional subspace, where $\tilde{r} = (K-1)r$ and $\calR(\cdot)$ denotes the column space of a matrix. Note $\overline{\calS}_k \supset \bigcup_{j = 1, j \neq k}^K \calS_j$. Also let $\tilde{\calS}_k$ denote an $\tilde{r}$-dimensional subspace $\tilde{\calS}_k \supset \calS_k$ such that $\tilde{\calS}_k$ and $\overline{\calS}_k$ satisfy \Cref{assum:subspaces}. Such a $\tilde{\calS}_k$ exists iff $(K-1)r < d/2$. Since $\calS_k \subset \tilde{\calS}_k$ and $\bigcup_{j \in [K], j \neq k} \calS_j \subset \overline{\calS}_k$, it suffices to transform $\overline{\calS}_k$ and $\tilde{\calS}_k$ into linearly separable sets. 

We directly apply \Cref{thm:binary-lin-sep} to transform $\overline{\calS}_k$ and $\tilde{\calS}_k$, and thus $\calS_k$ and $\bigcup_{j \in [K], j \neq k} \calS_j$, into linearly separable sets with high probability. Since this is now a problem of separating two $\tilde{r}$-dimensional subspaces,  the $r$ in \Cref{thm:binary-lin-sep} becomes $\tilde{r}$. Applying the union bound over all $k \in [K]$, a nonlinear layer of $\mathcal{O}\left( \mathrm{poly}(Kr) \right)$ width transforms a union of $K$ subspaces into $K$ one-vs-all linearly separable sets with high probability.

    \section{Experimental Results} \label{sec:empirical}
In this section, we empirically verify a single random nonlinear layer makes a UoS linearly separable for both synthetic and real-world data. Specifically, in \Cref{ssec:synthetic-data}, we verify our main results \Cref{thm:binary-lin-sep} and \Cref{cor:K-lin-sep} on synthetic data, and explore settings beyond our assumptions. In \Cref{ssec:cifar10-mcr2,ssec:real-images}, we provide experimental results on real images, which again support our theoretical results. All experiments in this section were conducted on a single NVIDIA A40 GPU.

\subsection{Synthetic Data} \label{ssec:synthetic-data}
In this subsection, we verify the early-layer features are linearly separable for synthetic data generated from the UoS model under various settings, including different nonlinear activations $\sigma(\cdot)$, network widths $D$, and the number of subspaces $K$. 

\smallskip

\paragraph{Synthetic data generation.} We first generated $K$ matrices $\U_1, \U_2, \dots, \U_K$ uniformly at random from the $d \times r$ Stiefel manifold. 
We then generated $N = K \cdot N_k$ training samples as follows, where $N_k = 5 \cdot 10^3$ in all settings. For all $k \in [K]$, we created $N_k$ samples via $\x_{k, i} = \U_k \z_i$, where $\z_i$ were sampled iid from $\calN(\0_r, \I_r)$ for all $i \in [N_k]$. Finally, when applicable, we generated $N$ test samples using the same procedure.

\begin{figure*}[t]
    \centering
    \begin{subfigure}{0.495\textwidth}
        \centering
        \includegraphics[width=\linewidth]{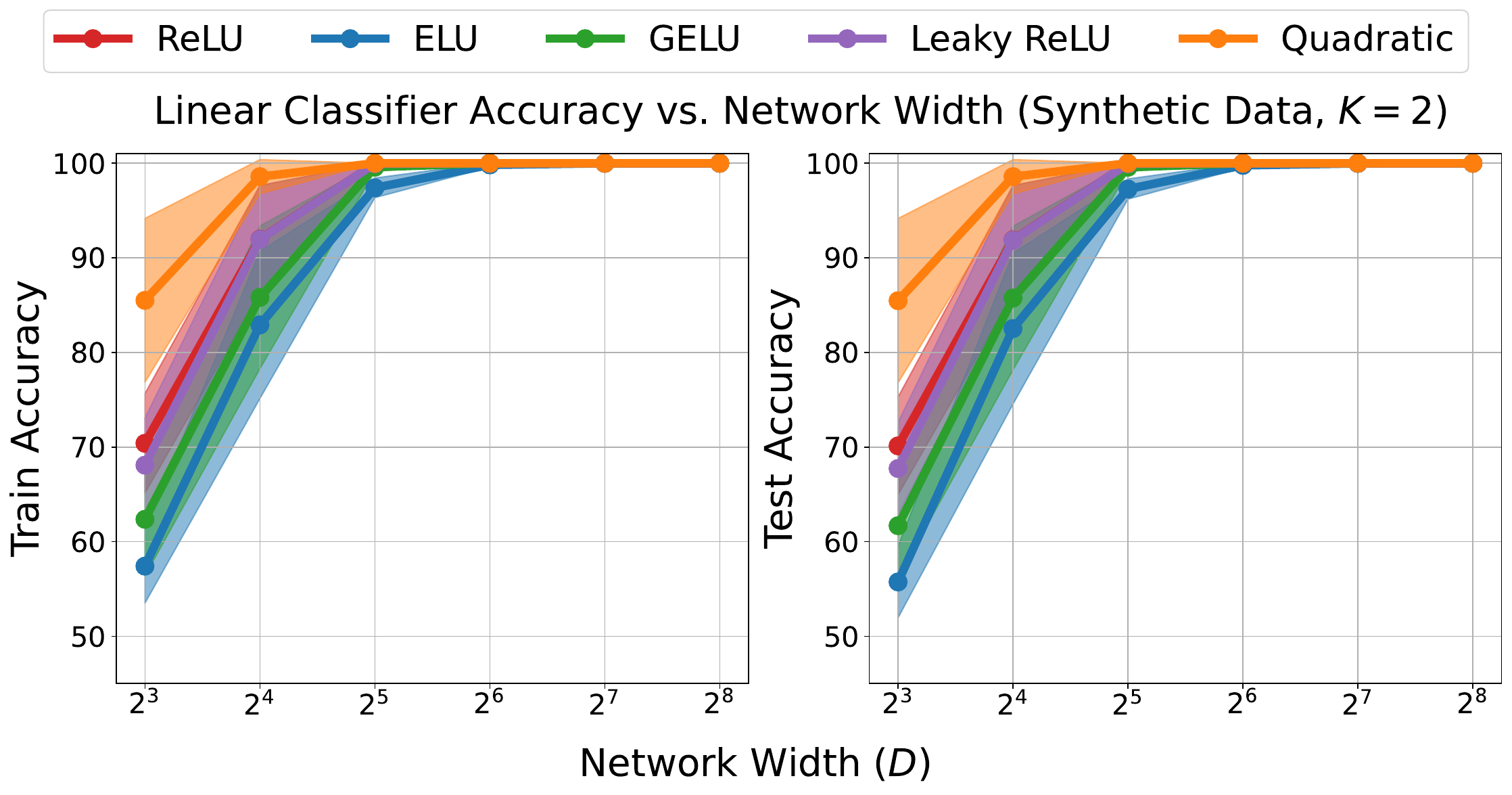}
        \caption{$K=2$ subspaces.}
        \label{subfig:train-test-acc-K2}
    \end{subfigure}\hfill
    \begin{subfigure}{0.495\textwidth}
        \centering
        \includegraphics[width=\linewidth]{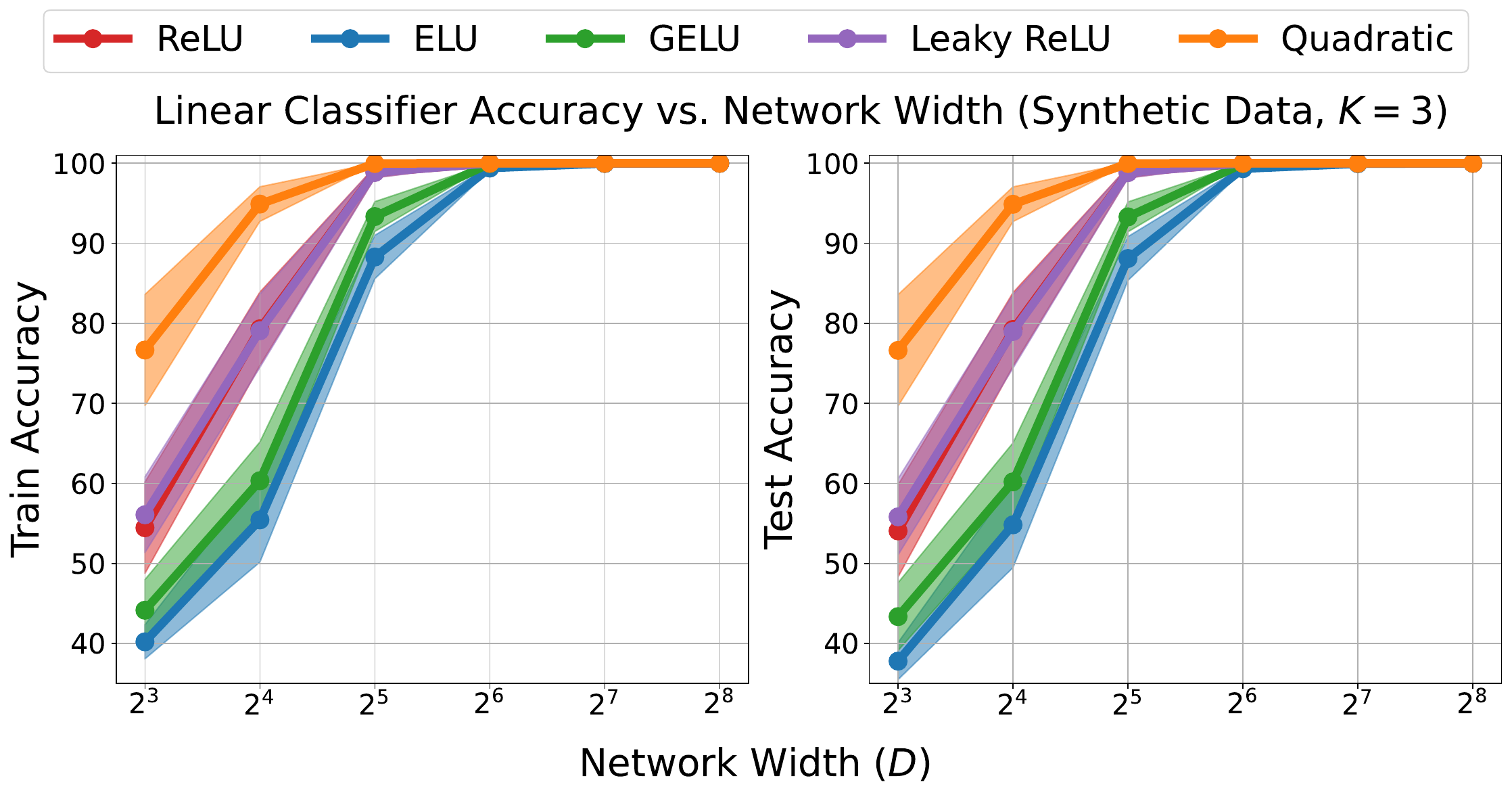}
        \caption{$K=3$ subspaces.}
        \label{subfig:train-test-acc-K3}
    \end{subfigure}
    \caption{ \textbf{Linear separability of random features on synthetic UoS data.} When the input data perfectly lie on a union of $K = 2$ (left) or $K = 3$ (right) subspaces, a linear classifier achieves perfect train and test accuracy when trained on features extracted by a sufficiently wide nonlinear layer with random weights.} 
    \label{fig:train-test-acc}
\end{figure*}

\paragraph{Model training.} 
We trained a linear classifier upon the random feature model in \Cref{eq:func-NN}. Specifically, we sampled the entries of $\W \in \reals^{D \times d}$ iid from $\calN(0, 10^{-2})$, then applied the random feature mapping \eqref{eq:func-NN}  on the train and test samples. Afterwards, we trained a linear classifier $\V \in \reals^{K \times D}$ on the train set random features under cross-entropy loss. After training, we used the trained classifier $\V$ to classify the test samples. We averaged all results over $10$ trials.

\paragraph{Results.} Based upon the above setup, we discuss the results below.

\begin{itemize}
    \item \textbf{Dependence on $K$ and $r$.} \Cref{fig:rank-K-sweep} shows the width of ReLU and quadratic layers have similar dependence w.r.t. the intrinsic dimension and the number of subspaces. At all values of $r$ and $K$, the linear classifier achieved perfect accuracy at similar widths for both activations. Thus, although our analysis assumes a quadratic activation, our empirical findings in \Cref{fig:rank-K-sweep} imply similar results hold under the ReLU activation. Further details on the experimental setup are in \Cref{sapp:quad-relu-comp}.  
    
    \item \textbf{Effects of nonlinear activations.} \Cref{fig:train-test-acc} shows the mean and standard deviation of the train and test accuracies at each network width for every activation function. Regardless of the activation, the linear classifier's mean accuracy across the trials increased as the network width grew, eventually achieving perfect classification performance. Furthermore, the standard deviation of the accuracies approached zero at sufficiently large widths. Although linear classifiers eventually achieve perfect accuracy for all activations, different activations required different widths to do so. Specifically, the quadratic requires noticeably smaller widths to achieve linear separability compared to the other activations. 
\end{itemize}

\subsection{CIFAR-10 MCR\texorpdfstring{$^2$}{} Representations} \label{ssec:cifar10-mcr2}
Second, we validate our results via experiments on the Maximal Coding Rate Reduction (MCR$^2$) representations \citep{yu2020learning} of the CIFAR-10 image dataset \citep{krizhevsky2009learning}. While natural images do not inherently adhere to a UoS model, they can be transformed into a UoS structure through nonlinear transformations. Specifically, MCR$^2$ \citep{yu2020learning} is a framework to learn data representations whose embeddings lie on a UoS \citep{wang2024a}.

\paragraph{Setup.} We trained a ResNet-18 model \citep{he2016deep} to learn MCR$^2$ representations of the CIFAR-10 dataset \citep{krizhevsky2009learning}. We adhered to the same architectural changes, hyperparameter settings, and training procedures as described in \cite{yu2020learning}. The resulting representations reside in a union of $K=10$ subspaces embedded in $\mathbb{R}^d$ with $d = 128$. From \cite{yu2020learning}, for each class $k \in [K]$, the representations of images in the $k^{\text{th}}$ class approximately lie on a 10-dimensional subspace, so $r \approx 10$. Additionally, the learned representations across different classes are nearly orthogonal, so $\theta_\ell \approx \pi/2$ for all $\ell \in [r]$.

We created training and testing sets with the MCR$^2$ representations, where each set contained $N = 10^4$ samples, with $N_k = 10^3$ samples per class. We then sampled a random weight matrix $\W \in \reals^{D \times d}$ with iid $\calN(0, 1)$ entries, and applied the random feature map $f_{\W}(\x)$ to the MCR$^2$ representations. We then trained a linear classifier $\V \in \reals^{K \times D}$ on the random features to classify the MCR$^2$ representations using cross-entropy loss. We employed the same activation functions as specified in \Cref{ssec:synthetic-data}. We varied the network width from $2^5$ to $2^{12}$ in powers of $2$, and averaged all results over $10$ trials.

\paragraph{Results.} \Cref{subfig:cifar10-mcr2} illustrates the mean and standard deviation of train and test accuracies achieved on the CIFAR-10 MCR$^2$ features across different activation functions and network widths. For each activation function, the mean accuracy of the linear classifier increased with the network width, ultimately approaching \emph{near-perfect} accuracy (approximately $99\%$). The standard deviation of accuracy across trials also diminished to nearly zero as the network width increased. We hypothesize that the failure to achieve $100\%$ accuracy is due to the representations not perfectly conforming to subspaces.

\subsection{Fashion MNIST and CIFAR-10}
\label{ssec:real-images}

\begin{figure*}[t]
    \centering
    \begin{subfigure}{0.495\textwidth}
        \centering
        \includegraphics[width=\linewidth]{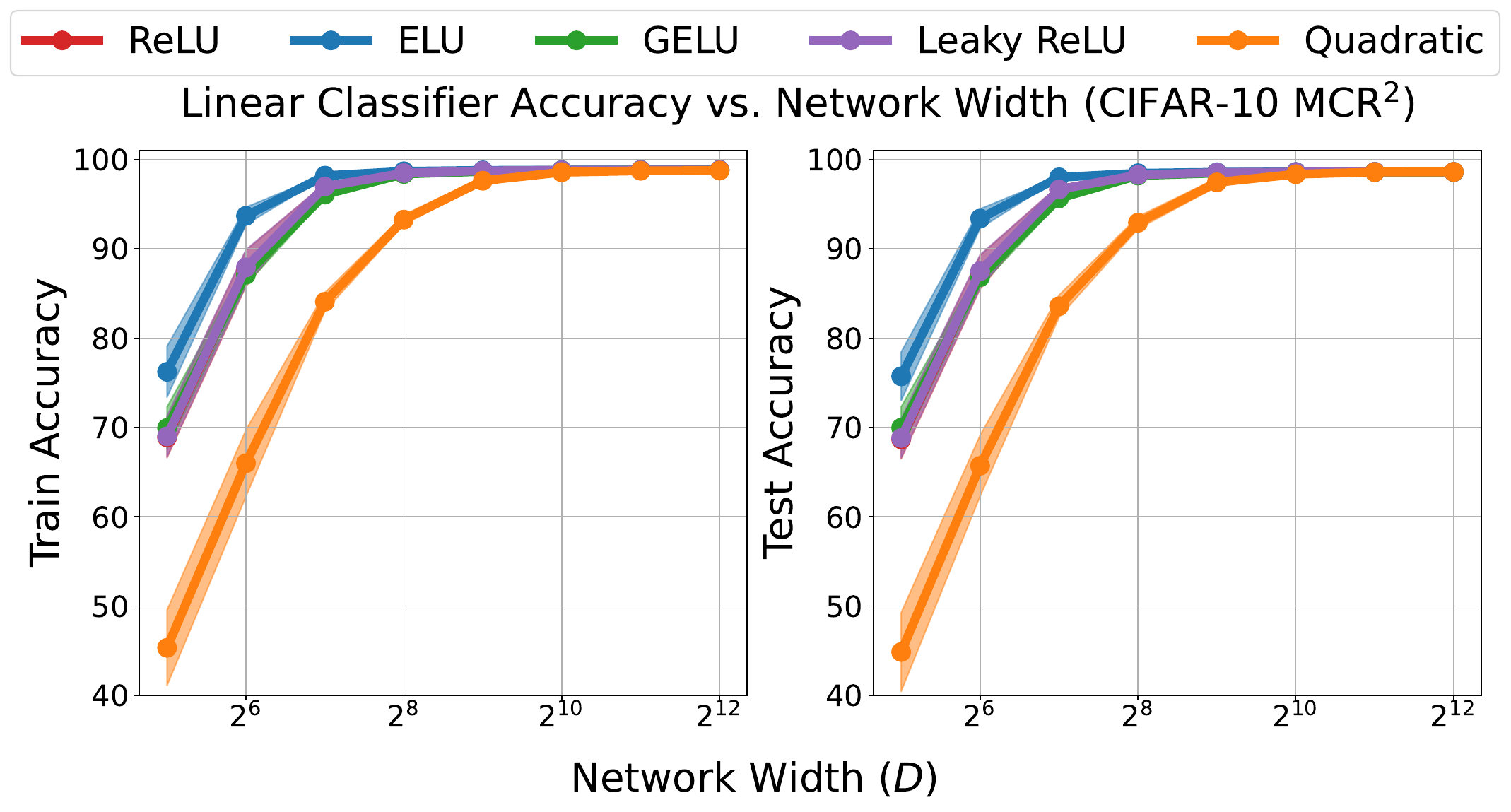}
        \caption{CIFAR-10 MCR$^2$ representations.}
        \label{subfig:cifar10-mcr2}
    \end{subfigure}\hfill
    \begin{subfigure}{0.495\textwidth}
        \centering
        \includegraphics[width=\linewidth]{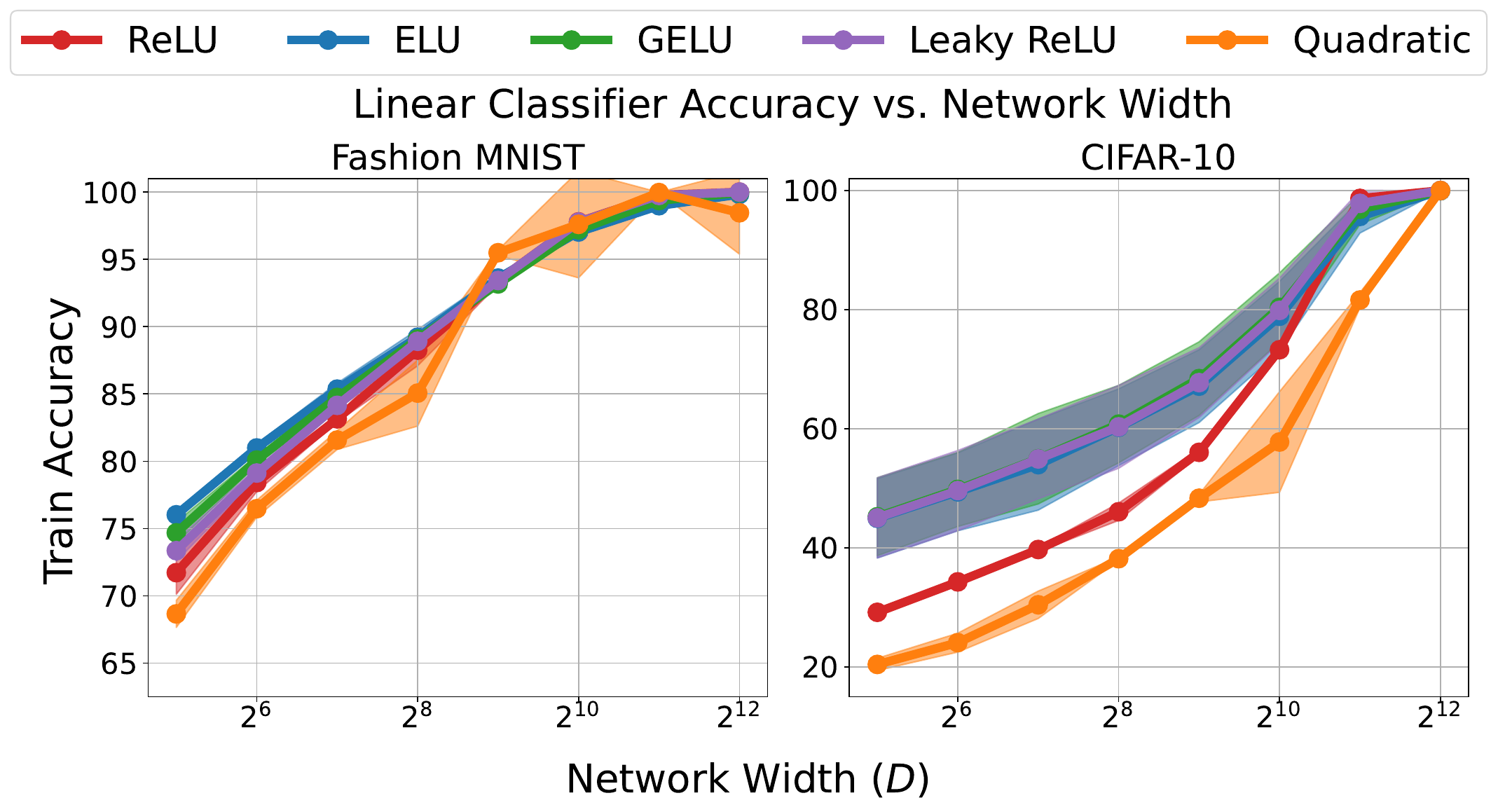}
        \caption{Fashion MNIST (left) and CIFAR-10 (right) images}
        \label{subfig:fashion-mnist-cifar10}
    \end{subfigure}
    \caption{ \textbf{Linear separability of random features on image data.} \textit{Left:} after transforming image data to lie on a UoS using the MCR$^2$ \citep{yu2020learning} framework, sufficiently many random features are linearly separable. \textit{Right:} sufficiently many random features using \emph{raw} image data as input are also linearly separable.} 
    \label{fig:real-images}
\end{figure*}

Finally, we show that our theory approximately holds on the Fashion MNIST \citep{xiao2017fashion} and CIFAR-10 \citep{krizhevsky2009learning} datasets. Both datasets contain $K = 10$ classes. \Cref{fig:image_data_class_svals} shows both datasets approximately satisfy the UoS data model: in both datasets, a small number of singular values account for a large majority each class data matrix's Frobenius norm. In particular, both datasets' per-class intrinsic subspace dimensions are about on the order of $10^1$, even though their ambient dimensions on the order of $10^2$ (Fashion MNIST) or $10^3$ (CIFAR-10).

\paragraph{Setup.} For both datasets, we randomly sampled $N = 10^4$ training images, with $N_k = 10^3$ images per class. Then, we flattened the images into $d$-dimensional vectors, where $d = 784$ for Fashion MNIST, and $d = 3072$ for CIFAR-10. Finally, we followed the exact training procedure as described in \Cref{ssec:cifar10-mcr2}, but averaged all results over $5$ trials instead. 

\paragraph{Results.} \Cref{subfig:fashion-mnist-cifar10} shows the mean and standard deviation of the train accuracies achieved on the Fashion MNIST and CIFAR-10 random features across various activation functions. Recall from \Cref{fig:image_data_class_svals}, the per-class subspace dimensions are on the order of $10^1$ in both datasets. \Cref{subfig:fashion-mnist-cifar10} shows the linear classifier achieves near-perfect accuracy at widths around $2^{11}$ or $2^{12}$ across all activations, which is on the order of $10^3$. Thus, network widths that are polynomial in the \emph{intrinsic} data dimension suffices for linear separability, which aligns with our theoretical results. 

Furthermore, recall that CIFAR-10's ambient dimension is about $4 \times$ that of Fashion MNIST ($d = 3072$ for CIFAR-10, while $d = 784$ for Fashion MNIST). Despite this difference, between $2^{11}$ and $2^{12}$ random features suffice for linear separability in \emph{both} datasets, implying little to no dependence on the data ambient dimension.

    \section{Related Works} \label{app:related}
In this section, we provide a more detailed discussion of the relationship between our results and prior work.

\paragraph{Separation capacity of nonlinear networks.} As discussed in \Cref{ssec:contributions}, \cite{dirksen2022separation, ghosal2022randomly} are most closely related to ours. They analyzed two arbitrarily-structured, nonlinearly-separated classes, and showed the resulting features from two-layer \citep{dirksen2022separation} and one-layer \citep{ghosal2022randomly} random ReLU networks are linearly separable with high probability. In our work, we specifically model the inputs as lying on a union of low-dimensional subspaces, and consider the quadratic activation instead of ReLU. Under our data model, the results in \cite{dirksen2022separation, ghosal2022randomly} require the network widths to scale \emph{exponentially} with the intrinsic dimension of the input data. In contrast, our result requires \emph{polynomial} dependence. Another related work, \cite{an2015can}, also assumes the data is from two arbitrary nonlinearly-separated sets. They prove there exists a \emph{deterministic} two-layer ReLU network that makes these sets linearly separable.

\paragraph{XOR data.}  Previous works on neural network analyses have considered XOR input data \citep{glasgow2024sgd,meng2024benign}, which is a special case of our UoS data model. To visualize this, consider an arbitrary data point $\bm x \in \left\{-1, +1\right\}^2 = \begin{bmatrix}
    x_1 & x_2
\end{bmatrix}$, where $x_i \in \{-1, +1\}$. Then, the corresponding label is $y = \operatorname{XOR}(x_1, x_2) = \begin{cases}
    +1 & x_1 = x_2 \\
    -1 & x_1 \neq x_2
\end{cases}$. \Cref{fig:xor-data} shows these data points lie on a union of two one-dimensional linear subspaces, where each linear subspace represents a different class. 
\begin{wrapfigure}{r}{0.4\textwidth}
    \begin{tikzpicture}[scale=0.8]

  \draw[<->, thick] (-2.2,0) -- (2.4,0);
  \draw[<->, thick] (0,-2.2) -- (0,2.4);

  \draw[dashed, blue!70, line width=1.2pt]
    (-1.8,-1.8) -- (1.8,1.8)
    node[above right, blue!80] {$\mathcal{S}_1$};

  \draw[dashed, orange!70, line width=1.2pt]
    (-1.8,1.8) node[above left, orange!70] {$\mathcal{S}_2$} -- (1.8,-1.8);

  \fill[blue!80]  ( 1.3, 1.3) circle (4pt)
    node[right, blue!90, font=\small] {$\quad y=+1$};
  \fill[blue!80]  (-1.3,-1.3) circle (4pt)
    node[left,  blue!90, font=\small] {$y=+1 \quad$};

  \fill[orange!80]  ( 1.3, -1.3) circle (4pt)
    node[right, orange!80, font=\small] {$\quad y=-1$};
  \fill[orange!80]  (-1.3,1.3) circle (4pt)
    node[left,  orange!80, font=\small] {$y=-1 \quad$};
    
    \end{tikzpicture}
    \caption{Two-dimensional XOR data lies on a union of one-dimensional subspaces.}
    \label{fig:xor-data}
    \vspace{-\intextsep}
    \vspace{-\intextsep}
\end{wrapfigure}

\paragraph{Neural collapse in shallow nonlinear networks.} Recently, \cite{hong2024beyond} studied the Neural Collapse (NC) phenomenon in shallow ReLU networks. Specifically, they identified sufficient data-dependent conditions on when shallow ReLU networks exhibit NC. Although our work and \cite{hong2024beyond} study the properties of the features in nonlinear networks, the settings have fundamental differences. Notably, NC characterizes the structure of the features from the \emph{penultimate} layer. Additionally, \cite{hong2024beyond} consider shallow ReLU networks to analyze NC in more realistic settings compared to previous works. In contrast, we study the linear separability of the features in the \emph{early} layers in DNNs, and study a shallow nonlinear network to facilitate such analysis. 

\paragraph{Learning with random features.} Our theoretical result uses random weights in the nonlinear layer, yielding a random feature map. Learning with random features was introduced in \citep{rahimi2007random} as an alternative to kernel methods, and its generalization properties have been widely studied \citep{rahimi2008weighted, rudi2017generalization, bach2017equivalence, li2021towards, chen2024conditioning}. Although our result does not directly imply broader conclusions about learning with random features, we show that a random feature map can transform subspaces into linearly separable sets with high probability. This implies if train and test samples lie on the same subspaces, a linear classifier can perfectly classify the test samples. 

\paragraph{Neural tangent kernel.} As  discussed in \Cref{ssec:theorem}, previous works have shown the global convergence of gradient descent on overparameterized two-layer networks using NTK techniques \citep{du2019gradient,huang2020dynamics}. These works showed that during training, a highly overparameterized two-layer network remains close to its random initialization, which is closely related to our random feature model. These results showed gradient descent converges to zero training loss for wide two-layer networks. In contrast, our work does not consider any training. Rather, we directly assume a random feature model, showing a nonlinear layer with random weights makes two linear subspaces linearly separable with high probability. 

\paragraph{Analysis of quadratic-activation networks.} 
Our theoretical result assumes the nonlinear activation is the entry-wise quadratic function. While previous works on quadratic activation have focused on the optimization landscape and generalization abilities of overparameterized networks \citep{li2018algorithmic, soltanolkotabi2018theoretical, du2018power,  sarao2020optimization, gamarnik2024stationary}, our contribution lies in demonstrating that data on a union of subspaces can be made linearly separable with high probability under this quadratic activation. This perspective provides new insights into the feature separation properties of quadratic activation networks, complementing the optimization-centric findings of prior studies. 

\paragraph{Rare Eclipse problem.} Finally, our problem shares conceptual similarities with the Rare Eclipse problem studied in \cite{bandeira2017compressive, cambareri2017rare}, which focuses on mapping two linearly separable sets into a lower-dimensional space where they become disjoint with high probability. Using Gordon's Escape through a Mesh \citep{gordon1988milman}, \cite{bandeira2017compressive} demonstrated that a random Gaussian matrix achieves this and provides a lower bound on the required dimension. Similarly, we show that a nonlinear random mapping can \emph{transform} two sets (linear subspaces) into linearly separable sets with high probability. However, beyond this shared goal of increasing separability, the two problems differ fundamentally in approach and context.

    \section{Conclusion} \label{sec:conclusion}
In this work, we studied the linear separability of early-layer features in nonlinear networks for low-dimensional data, using a UoS model motivated by the low intrinsic dimensionality of image data. We rigorously proved that a single nonlinear layer with random weights and quadratic activation can transform $K \geq 2$ subspaces into (one-vs.-all) linearly separable sets with high probability. Notably, our result requires the network width to be polynomial in the  intrinsic dimension, while previous results require exponential dependence. Although our analysis assumes a quadratic activation, our empirical findings on synthetic and real data indicate similar results hold for other activations, such as ReLU. 

\paragraph{Future work.} There are several interesting avenues for future work. First, as discussed in \Cref{ssec:uos}, a union of low-dimensional linear subspaces is a simplified model to capture the local linear structure in nonlinear manifolds. Relaxing the UoS data model to consider the global nonlinear structure in manifolds would be a natural extension to this work. For example, one could model a data sample $\x$ in the $k^{th}$ class as $\x := \phi(\U_k \bm \alpha)$, where $\phi(\cdot)$ is from a class of nonlinear functions, and $\U_k \in \reals^{d \times r_k}$ captures the data's low intrinsic dimensionality. Additionally, our experiments demonstrated replacing the quadratic with other activations, such as ReLU, yield similar results. Extending our analysis to consider other activations is another possible direction for future work. Finally, in this work, we proved \emph{there exists} a hyperplane $\v$ that separates the random features with high probability. However, this is not necessarily a max-margin hyperplane. Thus, considering an optimization over $\v$ would yield tighter bounds on the required width.

\clearpage

    \clearpage
    
    \bibliographystyle{alpha}
    \bibliography{refs}
    
    \clearpage


    \newpage 
    \appendix
    \section{Linear Separability of a Union of Subspaces}\label{app:problem} In this section, we discuss why a nonlinear  layer $f_{\bm W}(\bm x) = \sigma(\bm W \bm x)$ is necessary to transform a UoS into linearly separable sets. We first show two subspaces themselves are not linearly separable. Next, we show applying either a linear transformation or a nonlinear activation \emph{individually} are insufficient in making a UoS linearly separable. 

\paragraph{Subspaces are not linearly separable in general.} Suppose we have two one-dimensional subspaces $\calS_1, \calS_2 \subset \reals^2$ with bases $\u_1 = \begin{bmatrix}
        1 & 1
\end{bmatrix}^\top$ and $\u_2 = \begin{bmatrix}
    -1 & 1
\end{bmatrix}^\top$, respectively. As shown in \Cref{fig:activations-only}, there does not exist any hyperplane (line) that can linearly separate $\calS_1$ and $\calS_2$ because they both pass through the origin.

\paragraph{Linear mapping alone is insufficient for linear separability.} Now suppose $\calS_1$ and $\calS_2$ are arbitrary $r_1$ and $r_2$-dimensional subspaces of $\reals^d$, and define $g_{\W}(\x) := \W \x$, where $\x \in \reals^d$ and $\W \in \reals^{D \times d}$. We show $g_{\W}(\calS_1)$ and $g_{\W}(\calS_2)$ are not linearly separable sets. For any $k \in \{1, 2\}$ and arbitrary $\bm \alpha^{(k)} \in \reals^{r_k}$, we have
\begin{equation*}
    \W \U_k \bm \alpha^{(k)} = \tilde{\U}_k \bm \alpha^{(k)},
\end{equation*}
where $\tilde{\U}_k \in \reals^{D \times r_k}$. Therefore, the point $\W\U_k \bm \alpha^{(k)}$ lies in an $r_k$-dimensional subspace in $\reals^D$. Since this holds for all $\bm \alpha^{(k)} \in \reals^{r_k}$, the sets $g_{\W}(\calS_1)$ and $g_{\W}(\calS_2)$ remain as linear subspaces of $\reals^D$ that pass through the origin, which are not linearly separable in general. 

\paragraph{Nonlinear activations alone are insufficient for linear separability.} Second, for various activation functions, we show $\sigma(\calS_1)$ and $\sigma(\calS_2)$ are not linearly separable sets through counterexamples.  Again suppose the bases of $\calS_1, \calS_2 \subset \reals^2$ are $\u_1 = \begin{bmatrix}
    1 & 1
\end{bmatrix}^\top$ and $\u_2 = \begin{bmatrix}
    -1 & 1
\end{bmatrix}^\top$, respectively. Let us first consider the entry-wise quadratic activation, which we considered in our theoretical analysis. For any $\alpha \in \reals$, we have
    \begin{align*}
        &\sigma(\u_1 \alpha) = 
        \begin{bmatrix}
            1^2 \\
            1^2
        \end{bmatrix} \alpha^2 = 
        \begin{bmatrix}
            \alpha^2 \\
            \alpha^2
        \end{bmatrix} \; \text{and} \;
        \sigma(\u_2 \alpha) = 
        \begin{bmatrix}
            (-1)^2 \\
            1^2
        \end{bmatrix} \alpha^2 = 
        \begin{bmatrix}
            \alpha^2 \\
            \alpha^2
        \end{bmatrix},
    \end{align*}
    so $\sigma(\u_1 \alpha) = \sigma(\u_2 \alpha)$. 
    Therefore, the sets $\sigma(\calS_1)$ and $\sigma(\calS_2)$ are \emph{identical} (see \Cref{fig:activations-only}, left), clearly implying they are not distinguishable.
    
    \smallskip
    
    Next, consider $\sigma(\cdot) = \text{ReLU}(\cdot)$, which is more commonly used in practice. For any nonzero $\alpha \in \reals$, $\sigma(\u_1 \alpha) = \u_1 \alpha$ if $\alpha > 0$, and $\sigma(\u_1 \alpha) = \0_2$ if $\alpha < 0$. Additionally, $\sigma(\u_2 \alpha) = \begin{bmatrix}
        0 & \alpha
    \end{bmatrix}^\top$ if $\alpha > 0$, and $\sigma(\u_2 \alpha) = \begin{bmatrix}
        -\alpha & 0
    \end{bmatrix}^\top$ if $\alpha < 0$. 
    Therefore, $\sigma(\calS_1) = \big\{\x \in \reals^2: x_1 > 0, x_2 > 0\} \cup \{\0_2\}$, while $\sigma(\calS_2) = \big\{ \x \in \reals^2: x_1 > 0, x_2 = 0 \big\} \cup \big\{ \x \in \reals^2: x_1 = 0, x_2 > 0 \big\}$, where $x_1$ and $x_2$ respectively denote the first and second elements of $\x \in \reals^2$. These sets are \emph{not} linearly separable (see \Cref{fig:activations-only}, right), since some points in $\sigma(\mathcal{S}_2)$ are above $\sigma(\mathcal{S}_1)$, and some points are below.

\begin{figure}[t]
    \centering
    \begin{minipage}{0.5\textwidth}
        \centering
        \begin{subfigure}{\textwidth}
        \centering
        \begin{tikzpicture}[scale=0.5]
            \draw[<->, thick] (-2, 0) -- (2, 0); 
            \draw[<->, thick] (0, -2) -- (0, 2); 
            
            \draw[<->, thick, blue] (-1.5, -1.5) -- (1.5, 1.5) node[above right] {$\calS_1$}; 
            \draw[<->, thick, orange] (1.5, -1.5) -- (-1.5, 1.5) node[above left] {$\calS_2$}; 
            
            \draw[->, thick, black] (3.0, 0) -- (4.5, 0) node[midway, above] {Quadratic};
            
            \draw[<->, thick] (5.5, 0) -- (9.5, 0); 
            \draw[<->, thick] (7.5, -2) -- (7.5, 2); 
            
            \draw[->, thick, blue] (7.5, 0) -- (9, 1.5) node[above right] {$\sigma(\calS_1)$}; 
            \draw[->, thick, orange, dashed] (7.5, 0) -- (9, 1.5) node[above left] {$\sigma(\calS_2)$}; 
        \end{tikzpicture}
        \caption{Quadratic applied to $\mathrm{span}(\u_1) \cup \mathrm{span}(\u_2)$.}
        \end{subfigure}
    \end{minipage}%
    \hfill
    \begin{minipage}{0.5\textwidth}
        \centering
        \begin{subfigure}{\textwidth}
            \centering
            \begin{tikzpicture}[scale=0.5]
            \draw[<->, thick] (-2, 0) -- (2, 0); 
            \draw[<->, thick] (0, -2) -- (0, 2); 
            
            \draw[<->, thick, blue] (-1.5, -1.5) -- (1.5, 1.5) node[above right] {$\calS_1$}; 
            \draw[<->, thick, orange] (1.5, -1.5) -- (-1.5, 1.5) node[above left] {$\calS_2$}; 
            
            \draw[->, thick, black] (3.0, 0) -- (4.5, 0) node[midway, above] {ReLU};
            
            \draw[<->, thick] (5.5, 0) -- (9.5, 0); 
            \draw[<->, thick] (7.5, -2) -- (7.5, 2); 
            
            \draw[->, thick, blue] (7.5, 0) -- (9, 1.5) node[above right] {$\sigma(\calS_1)$}; 
            \draw[->, thick, orange, dashed] (7.5, 0) -- (7.5, 1.5) node[above left] {$\sigma(\calS_2)$}; 
            \draw[->, thick, orange, dashed] (7.5, 0) -- (9.0, 0) node[below] {$\sigma(\calS_2)$}; 
        \end{tikzpicture}
        \caption{ReLU applied to $\mathrm{span}(\u_1) \cup \mathrm{span}(\u_2)$.}
        \end{subfigure}
    \end{minipage}
    \caption{\textbf{Activation alone is insufficient for linearly separating two subspaces.} When $\calS_1 = \mathrm{span}(\u_1)$ and $\calS_2 = \mathrm{span}(\u_2)$, the sets $\sigma(\calS_1)$ and $\sigma(\calS_2)$ are not linearly separable for $\sigma(\cdot) = $ quadratic (left) and $\sigma(\cdot) = \mathrm{ReLU}(\cdot)$ (right).} 
    \label{fig:activations-only}
\end{figure}
    \section{Supporting Results} \label{app:supporting}
\vspace{-0.25cm}
We provide supporting Lemmas that are useful in proving Theorem~\ref{thm:binary-lin-sep}. Beforehand, we re-state previous notation here for convenience, and introduce some new notation. We use $\calN(\mu, \sigma^2)$ to denote a Gaussian distribution with mean $\mu$ and variance $\sigma^2$, $\calN(\bm{\mu}, \bm{\Sigma})$ to denote a multivariate Gaussian distribution with mean $\bm{\mu}$ and covariance $\bm{\Sigma}$, and $\chi^2_m$ to denote a chi-squared distribution with $m$ degrees of freedom. We use $Z \: \big| \: \calA$ to denote random variable $Z$ conditioned on an event $\calA$. We denote the pdf of a random variable $Z$ with $f_Z(\cdot)$, and the covariance of a random vector with $\covrm(\cdot)$. 

We use $\| \cdot \|$ to denote the Euclidean norm of a vector, $\sigma_i(\cdot)$ to denote the $i^{th}$ largest singular value of a matrix, and $\lambda_i(\cdot)$ to denote the $i^{th}$ largest eigenvalue of a symmetric matrix. We also use $\0_m$ to denote the $m$-dimensional vector of all zeroes.

For any positive integer $N$, we use $[N]$ to denote the set $\{1, 2, \dots, N\}$. With a slight abuse of notation, for some function $\phi$ and set $\calX$, $\phi\big( \calX \big)$ denotes the set $\big\{ \x \in \calX: \phi \big( \x \big) \big\}$. 

Let $\w \sim \calN(\0_d, \I_d)$, $\U_1, \U_2 \in \reals^{d \times r}$ be such that their columns are orthonormal bases for $\calS_1$ and $\calS_2$, respectively, where the subspaces satisfy \Cref{assum:subspaces}. Also let $\x := \U_1^\top \w$, and $\y := \U_2^\top \w$. Note $\x \sim \calN(\0_r, \I_r)$ and $\y \sim \calN(\0_r, \I_r)$.  Also let $X := \|\x\|^2$ and $Y := \|\y\|^2$, meaning $X, Y \sim \chi^2_r$. Note $\x$ and $\y$, as well as $X$ and $Y$, are \emph{correlated}. Finally, let $\ab$ and $\blb$ be random vectors with the following distributions:
\begin{align*}
    \ab \sim \x \: \big| \: \|\x\|^2 > \|\y\|^2 \; \; \text{and} \; \; \blb \sim \y \: \big| \: \|\x\|^2 > \|\y\|^2.
\end{align*}

\subsection{Expectation of Order Statistics: \texorpdfstring{$\chi^2_m$}{ } Random Variables}

\Cref{lem:expec-max-min-chi-squares} provides exact expressions for the expectation of the maximum and minimum of two iid $\chi^2_m$ random variables.

\begin{lemma}
    \label{lem:expec-max-min-chi-squares}
    Let $X, Y \overset{\text{iid}}{\sim} \chi^2_m$, $A = \maxrm\{X, Y\}$, and $B = \minrm\{X, Y\}$. Then,
    \begin{align*}
        &\eE[A] = m + \frac{2}{\sqrt{\pi}} \frac{\Gamma((m+1)/2)}{\Gamma(m/2)}, \; \; \text{and} \; \; \eE[B] = m - \frac{2}{\sqrt{\pi}} \frac{\Gamma((m+1)/2)}{\Gamma(m/2)},
    \end{align*}
    where $\Gamma(\cdot)$ denotes the Gamma function.
\end{lemma}
\begin{proof}
Note $A + B = X + Y$, so $\eE[A + B] = \eE[X + Y] = 2m$. Therefore, it suffices to compute $\eE[A]$:
\begin{align*}
    \eE[A] &= \int\limits_0^\infty \int\limits_0^\infty \maxrm\{x, y\} f_X(x) f_Y(y) \: dx \: dy \\
    &= \int\limits_0^\infty \int\limits_y^\infty x f_X(x) f_Y(y) \: dx \: dy + \int\limits_0^\infty \int\limits_x^\infty y f_X(x) f_Y(y) \: dy \: dx = 2 \int\limits_0^\infty \int\limits_y^\infty x f_X(x) f_Y(y) \: dx \: dy \\
    &\overset{(a)}{=} \frac{2}{2^m \Gamma(m/2)^2} \int\limits_0^\infty \int\limits_y^\infty x^{m/2}e^{-x/2} y^{m/2 - 1} e^{-y/2} \: dx \: dy, 
\end{align*}
where we substituted the pdf of a $\chi^2_m$ distribution in $(a)$. Letting $t = \frac{x}{2}$ results in
\begin{align*}
    &\frac{2}{2^m \Gamma(m/2)^2} \int\limits_0^\infty \int\limits_y^\infty x^{m/2}e^{-x/2} y^{m/2 - 1} e^{-y/2} \: dx \: dy \\
    &= \frac{4}{2^{m/2} \Gamma(m/2)^2} \int\limits_0^\infty \int\limits_{y/2}^\infty t^{m/2} e^{-t} y^{m/2 - 1} e^{-y/2} \: dt \: dy \\
    &\overset{(b)}{=} \frac{4}{2^{m/2} \Gamma(m/2)^2} \int\limits_0^\infty \Gamma(m/2 + 1, y/2) y^{m/2 - 1}e^{-y/2} \: dy, 
\end{align*}
where in $(b)$, we substituted the definition of the upper incomplete Gamma function, denoted as $\Gamma(p, x)$. Using the recurrence relation $\Gamma(p + 1, x) = p\Gamma(p, x) + x^p e^{-x}$ yields
\begin{align*}
    &\frac{4}{2^{m/2} \Gamma(m/2)^2} \int\limits_0^\infty \Gamma(m/2 + 1, y/2) y^{m/2 - 1}e^{-y/2} \: dy \\
    &= \underbrace{\frac{m}{\Gamma(m/2)^2} \int\limits_0^\infty \Gamma(m/2, y/2) (y/2)^{m/2 - 1} e^{-y/2} \: dy}_{\text{$(c)$}} + \underbrace{\frac{1}{2^{m-2} \Gamma(m/2)^2} \int\limits_0^\infty y^{m-1} e^{-y} \: dy}_{\text{$(d)$}}. 
\end{align*}
We first simplify $(c)$. Letting $s = y / 2$, $(c)$ becomes
\begin{equation*}
    \frac{2m}{\Gamma(m/2)^2}\int\limits_0^\infty \Gamma(m/2, s) s^{m/2 - 1} e^{-s} \: ds.
\end{equation*}
From pg. 137, eq. (8) in \cite{bateman1953higher}:
\begin{equation*}
    \int\limits_0^\infty \Gamma(m/2, s) s^{m/2 - 1} e^{-s} \: ds = \frac{\Gamma(m)}{(m/2) \cdot 2^m} {}_2F_1(1, m; m/2 + 1; 1/2)
\end{equation*}
where ${}_2F_1(a, b; c, d)$ denotes the ordinary hypergeometric function. By Gauss's Second Summation Theorem \cite{slater1966generalized}:
\begin{equation} \label{eq:gauss-second-sum}
    \frac{\Gamma(m)}{(m/2) \cdot 2^m} {}_2F_1(1, m; m/2 + 1; 1/2) = \frac{\Gamma(m) \Gamma(1/2) \Gamma(m/2 + 1)}{(m/2) \cdot 2^m \cdot \Gamma((m+1)/2)}.
\end{equation}
By Legendre's duplication formula, $\Gamma(m) = \frac{\Gamma(m/2) \Gamma((m + 1)/2)}{2^{1-m} \sqrt{\pi}}$. Additionally, the Gamma function satisfies the recurrence relation $\Gamma(z + 1) = z\Gamma(z)$ for all $z > 0$. Substituting these expressions into \eqref{eq:gauss-second-sum} leads to
\begin{equation*}
    \frac{\Gamma(m) \Gamma(1/2) \Gamma(m/2 + 1)}{(m/2) \cdot 2^m \cdot \Gamma((m+1)/2)} 
    = \frac{\Gamma(m/2)\Gamma(m/2 + 1)}{m} = \frac{\Gamma(m/2)^2}{2}.
\end{equation*}
Therefore, $(c)$ fully simplifies to the following:
\begin{equation*} 
    \frac{m}{\Gamma(m/2)^2} \int\limits_0^\infty \Gamma(m/2, y/2) (y/2)^{m/2 - 1} e^{-y/2} \: dy = \frac{2m}{\Gamma(m/2)^2}\frac{\Gamma(m/2)^2}{2} = m.
\end{equation*}
We now simplify $(d)$:
\begin{equation*}
    \frac{1}{2^{m-2} \Gamma(m/2)^2} \int\limits_0^\infty y^{m-1} e^{-y} \: dy \overset{(e)}{=} \frac{\Gamma(m)}{2^{m-2} \Gamma(m/2)^2} \overset{(f)}{=} \frac{2\Gamma((m+1)/2)}{\Gamma(m/2)\sqrt{\pi}},
\end{equation*}
where $(e)$ is by the definition of the Gamma function, and $(f)$ is by Legendre's duplication formula. Thus,
\begin{equation*}
    \eE[A] = m + \frac{2}{\sqrt{\pi}}\frac{\Gamma((m+1)/2)}{\Gamma(m/2)}.
\end{equation*}
We then use the property $\eE[A + B] = \eE[A] + \eE[B] = 2m$ to obtain $\eE[B]$:
\begin{equation*}
    \eE[B] = m - \frac{2}{\sqrt{\pi}}\frac{\Gamma((m+1)/2)}{\Gamma(m/2)}.
\end{equation*}
\end{proof}

\subsection{Eigenvalues of Difference between Projection Matrices}
Next, we provide a result about the eigenvalues of $\U_1\U_1^\top - \U_2\U_2^\top$.
\begin{lemma}
\label{lem:proj-mat-diff-eigvals}
    Let $\U_1, \U_2 \in \reals^{d \times r}$ s.t. $\U_1^\top \U_1 = \U_2^\top \U_2 = \I_r$, and $\sigma_\ell(\U_1^\top \U_2) = \cos(\theta_\ell)$ for all $\ell \in [r]$, where $\theta_1 := \theta_{min} > 0$. Then, $\U_1\U_1^\top - \U_2\U_2^\top$ has $r$ eigenvalues equal to $\sin(\theta_1), \sin(\theta_2), \dots, \sin(\theta_r)$, $r$ eigenvalues equal to $-\sin(\theta_1), -\sin(\theta_2), \dots, -\sin(\theta_r)$, and $d - 2r$ eigenvalues equal to $0$.
\end{lemma}

    \begin{proof}
        Let $\Phi := \U_1\U_1^\top - \U_2\U_2^\top \in \reals^{d \times d}$. We derive an exact expression for the characteristic polynomial $\det\Big(\Phi - \lambda \I_d  \Big)$. First, note 
        \begin{equation*}
            \Phi = \begin{bmatrix}
                \U_1 & \U_2
            \end{bmatrix} \begin{bmatrix}
                \U_1^\top \\
                -\U_2^\top
            \end{bmatrix},
        \end{equation*}
        and let $\U \mSigma \V^\top$ be a singular value decomposition of $\U_1^\top \U_2  \in \reals^{r \times r}$. Then, assuming $\lambda \neq 0$,
        \begin{align*}
            \det&\Big(\Phi - \lambda \I_d \Big) = (-1)^d \lambda^d \det\Big(\I_d - \frac{1}{\lambda}\Phi\Big) =  (-1)^d \lambda^d \det\bigg(\I_d - \frac{1}{\lambda} \begin{bmatrix}
                \U_1 & \U_2
            \end{bmatrix} \begin{bmatrix}
                \U_1^\top \\
                -\U_2^\top
            \end{bmatrix} \bigg) \\
            &\overset{(a)}{=}  (-1)^d \lambda^d \det\bigg(\I_{2r} - \frac{1}{\lambda} \begin{bmatrix}
                \I_r & \U \mSigma \V^\top \\
                -\V \mSigma \U^\top & -\I_r
            \end{bmatrix} \bigg) \\
            &= (-1)^d \lambda^d \det\bigg( \begin{bmatrix}
                (1 - 1/\lambda) \I_r & -(1/\lambda) \U \mSigma \V^\top \\
                (1/\lambda)\V \mSigma \U^\top & (1 + 1/\lambda) \I_r
            \end{bmatrix} \bigg) \\
            &\overset{(b)}{=} (-1)^d \lambda^d (1 - 1/\lambda)^r \det\bigg((1 + 1/\lambda)\I_r + \frac{(1 / \lambda^2)}{1 - 1/\lambda} \V \mSigma^2 \V^\top  \bigg) \\
            &= (-1)^d \lambda^d (1 - 1/\lambda)^r \det(\V) \det\bigg( (1 + 1/\lambda) \I_r + \frac{(1/\lambda^2)}{1 - 1/\lambda} \mSigma^2 \bigg) \det(\V^\top) \\
             &= (-1)^d \lambda^d \det\bigg( (1 - 1/\lambda^2) \I_r + (1/\lambda^2) \mSigma^2 \bigg) = (-1)^d \lambda^{d-2r} \prod_{\ell}^r \Big[ \lambda^2 - 1 + \cos^2(\theta_\ell) \Big] \\
            &= (-1)^d \lambda^{d-2r} \prod_{\ell=1}^r \Big[ \big( \lambda + \sin(\theta_\ell) \big) \big(\lambda - \sin(\theta_\ell) \big)
            \Big] 
        \end{align*}
        where $(a)$ is from Sylvester's Determinant Identity, and $(b)$ is from the fact that $$\det\bigg( \begin{bmatrix}
            \A & \B \\ \mC & \D
        \end{bmatrix} \bigg) = \det(\A)\det(\D - \mC \A^{-1} \B)$$ for invertible $\A$.
        Solving for the roots of $\det\Big(\Phi - \lambda\I_d\Big) = 0$ yields $\lambda = \pm \sin(\theta_\ell)$ for all $\ell \in [r]$. Therefore, $\Phi$ has $2r$ eigenvalues equal to $\pm \sin(\theta_1), \pm \sin(\theta_2), \dots, \pm \sin(\theta_r)$. Although we also have $\lambda = 0$ with multiplicity $d - 2r$, we initially assumed $\lambda \neq 0$, so these roots are invalid.
        
        We now show the remaining $d - 2r$ eigenvalues must be $0$. We showed there are at least $2r$ eigenvalues that are non-zero, so $2r \leq \rank(\Phi).$ Additionally, we have $$\rank(\Phi) = \rank(\U_1\U_1^\top - \U_2\U_2^\top) \leq \rank(\U_1 \U_1^\top) + \rank(-\U_2 \U_2^\top) = 2r.$$
        Thus, $2r \leq \rank(\Phi) \leq 2r,$ which implies $\rank(\Phi) = 2r$. Therefore, $\Phi$ must have \emph{exactly} $2r$ non-zero eigenvalues, implying the remaining $d - 2r$ eigenvalues must all be equal to $0$.
    \end{proof}

\subsection{Expectation of Random Symmetric Rank-\texorpdfstring{$1$}{ } Matrices} \label{ssec:expec-symm-rank-1}
\vspace{-0.15cm}

We provide upper and lower bounds for $\eE[\ab\ab^\top]$ and $\eE[\blb\blb^\top]$. We first show $\eE[\ab\ab^\top]$ and $\eE[\blb\blb^\top]$ are isotropic matrices.
\begin{lemma}
\label{lem:aa-bb-theta-isotropic}
    Let $\w \sim \mathcal{N}(\0_d, \I_d)$, $\x := \U_1^\top \w$, $\y := \U_2^\top \w$, $\ab \sim \x \: \big| \: \|\x\|^2 > \|\y\|^2$, and $\blb \sim \y \: \big| \: \|\x\|^2 > \|\y\|^2$. Then, $\eE\big[\ab \ab^\top\big]$ and $\eE\big[\blb \blb^\top\big]$ are both isotropic matrices.
\end{lemma}
\begin{proof}
    Since $\covrm(\x) = \I_r$ and $\covrm(\y) = \I_r$, which are isotropic matrices, $\covrm(\ab)$ and $\covrm(\blb)$ are also isotropic matrices. Thus, it suffices to show $\eE[\ab] = \0_r$ and $\eE[\blb] = \0_r$.
    \begin{align*}
        \eE&[\ab] = \eE_{\x, \y \sim \calN(\0_r, \I_r)}\big[ \x \: | \: \|\x\|^2 > \|\y\|^2 \big] \\
        &= \int\limits_0^\infty \int\limits_y^\infty \eE_{\x \sim \calN(\0_r, \I_r)}\big[ \x \: | \: \|\x\|^2 = x \big] f_{X, Y}(x, y) \: dx \: dy \overset{(a)}{=} \0_r,
    \end{align*}
    where $X, Y \sim \chi^2_r$, and $(a)$ is because $\x \: \big| \: \|\x\|^2 = x$ is distributed uniformly on the sphere of radius $\sqrt{x}$, so $\eE\big[\x \: | \: \|\x\|^2 = x \big] = \0_r$. We can use the same argument to show $\eE[\blb] = \0_r$. Therefore, $\eE[\ab \ab^\top] = \covrm(\ab)$ and $\eE[\blb \blb^\top] = \covrm(\blb)$, which are both isotropic matrices.
\end{proof}

\bigskip

\noindent The next result provides upper and lower bounds for $\eE[\ab \ab^\top]$ and $\eE[\blb \blb^\top]$.
\begin{lemma}
\label{lem:expec-aa-bb-theta}
    Let $\w \sim \calN(\0_d, \I_d)$, $\x := \U_1^\top \w$, $\y := \U_2^\top \w$, $\ab \sim \x \: \big| \: \|\x\|^2 > \|\y\|^2$, and $\blb \sim \y \: \big| \: \|\x\|^2 > \|\y\|^2$. Then, we have
    \begin{align*}
       &\Bigg(1 + \sqrt{\frac{2}{\pi}} \cdot \frac{\sin(\theta_1)}{\sqrt{r+1}}\Bigg) \I_r \preceq \eE\big[ \ab \ab^\top \big] \preceq \Bigg(1 + \frac{1}{r} \cdot \sqrt{\sum\limits_{\ell=1}^r \sin^2(\theta_\ell)} \Bigg) \I_r, \; \; \text{and} \\
        &\Bigg( 1 - \frac{1}{r} \cdot \sqrt{\sum\limits_{\ell=1}^r \sin^2(\theta_\ell)} \Bigg) \I_r \preceq \eE\big[ \blb \blb^\top \big] \preceq \Bigg(1 - \sqrt{\frac{2}{\pi}} \cdot \frac{\sin(\theta_1)}{\sqrt{r+1}}\Bigg) \I_r.
    \end{align*}
\end{lemma}
\begin{proof}
    By Lemma~\ref{lem:aa-bb-theta-isotropic}, $\eE[\ab\ab^\top]$ is an isotropic matrix, so it suffices to upper and lower bound  $\trrm\big(\eE[\ab\ab^\top]\big) = \eE\big[\trrm(\ab\ab^\top)\big] = \eE[\|\ab\|^2]$.  By definition of $\ab$, $\|\ab\|^2 \sim \maxrm\{X, Y\}$, where $X, Y \sim \chi^2_r$ are not necessarily independent. We first note
    \begin{equation*}
        \|\ab\|^2 = \frac{1}{2} \Big( \|\x\|^2 + \|\y\|^2 + \big| \|\x\|^2 - \|\y\|^2 \big| \Big).
    \end{equation*}
     Therefore,
     \begin{equation} \label{eq:norm_ab_theta}
         \eE[\|\ab\|^2] = \frac{1}{2} \Big(\eE[\|\x\|^2] + \eE[\|\y\|^2] + \eE\big[ | \|\x\|^2 - \|\y\|^2 | \big] \Big) = r + \frac{1}{2}\Big(\eE\big[ | \|\x\|^2 - \|\y\|^2 | \big] \Big),
     \end{equation}
     so it suffices to upper and lower bound $\eE\big[ | \|\x\|^2 - \|\y\|^2 | \big]$. First, note 
     \begin{equation} \label{eq:expect-abs-norm-diff}
        \eE\Big[ \big| \|\x\|^2 - \|\y\|^2 \big| \Big] = \eE\Big[ \big| \|\U_1^\top \w\|^2 - \|\U_2^\top \w\|^2 \big| \Big] 
        = \eE\Big[ \big| \w^\top (\U_1 \U_1^\top - \U_2 \U_2^\top) \w \big| \Big]. 
     \end{equation}
     Let $\Phi := \U_1\U_1^\top - \U_2\U_2^\top$. We first establish an upper bound as such:
     \begin{align*}
         \eE&\Big[ \big| \w^\top (\U_1 \U_1^\top - \U_2 \U_2^\top) \w \big| \Big] = \eE\Big[ \sqrt{(\w^\top \Phi \w)^2 } \Big] 
         \overset{(a)}{\leq} \sqrt{\eE\Big[ (\w^\top \Phi \w)^2 \Big]} \\ 
         &= \sqrt{\varrm(\w^\top \Phi \w)} \overset{(b)}{=} \sqrt{2\trrm\Big( \Phi^2 \Big)} = 2 \sqrt{\sum\limits_{\ell=1}^r \sin^2(\theta_\ell)},
     \end{align*}
     where $(a)$ is from Jensen's inequality, $(b)$ is from Eq. (381) in \cite{petersen2008matrix}, and the last equality is due to \Cref{lem:proj-mat-diff-eigvals}. Therefore, 
     \begin{equation*}
         \eE[\|\ab\|^2] \leq r + \sqrt{\sum\limits_{\ell=1}^r \sin^2(\theta_\ell)} \implies \eE[\ab \ab^\top] \preceq \Bigg(1 + \frac{1}{r} \cdot \sqrt{\sum\limits_{\ell=1}^r \sin^2(\theta_\ell)} \Bigg) \I_r.
     \end{equation*}
     We now establish a lower bound for $\eE\Big[ \big| \|\x\|^2 - \|\y\|^2 \big| \Big]$. Note $\Phi$ is symmetric, so there exists an eigendecomposition $\Phi = \Q \Lambda \Q^\top$ where $\Q \in \reals^{d \times d}$ is an orthogonal matrix, and $\Lambda$ is a diagonal matrix consisting of the eigenvalues of $\Phi$. We assume the eigenvalues are listed in descending order in $\Lambda$. By \Cref{lem:proj-mat-diff-eigvals}, $\Phi$ has $2r$ non-zero eigenvalues equal to $\pm \sin(\theta_1), \pm\sin(\theta_2), \dots, \pm\sin(\theta_r)$. Therefore:
     \begin{equation} 
         \w^\top \Phi \w = \w^\top \Q \Lambda \Q^\top \w := \z^\top \Lambda \z = \sum\limits_{\ell=1}^r \sin(\theta_\ell) \big[ z_\ell^2 - z_{d - \ell + 1}^2 \big],
     \end{equation}
    where $\z := \Q^\top \w \sim \calN(\0_d, \I_d)$. Therefore, 
    \begin{equation*}
        \eE\Big[ \big| \|\x\|^2 - \|\y\|^2 \big| \Big] = \eE\Big[ | \z^\top \Lambda \z | \Big] = \eE \bigg[ \Big| \sum\limits_{\ell=1}^r \sin(\theta_\ell) \big[ z_\ell^2 - z_{d - \ell + 1}^2 \big] \Big| \bigg].
    \end{equation*}
    Before we proceed, we first note $0 < \sin(\theta_1) \leq \sin(\theta_\ell) \leq 1$ for all $\ell \in [r]$, so
    \begin{equation*}
        \Big| \sin(\theta_\ell) \big[ z_\ell^2 - z_{d - \ell + 1}^2 \big] \Big| \geq \Big| \sin(\theta_1) \big[ z_\ell^2 - z_{d - \ell + 1}^2 \big] \Big| = \sin(\theta_1) \big| z_\ell^2 - z_{d - \ell + 1}^2  \big|
    \end{equation*}
    for all $\ell \in [r]$. Thus, we have
    \begin{align*}
        \sum\limits_{\ell=1}^r &\Big| \sin(\theta_\ell) \big[ z_\ell^2 - z_{d - \ell + 1}^2 \big] \Big| \geq \sin(\theta_1) \sum\limits_{\ell=1}^r \big| z_\ell^2 - z_{d - \ell + 1}^2 \big| \overset{(c)}{\geq} \sin(\theta_1) \bigg| \sum\limits_{\ell=1}^r  z_\ell^2 - z_{d - \ell + 1}^2 \bigg|,
    \end{align*}
    where $(c)$ is from the Triangle Inequality. Let $Z_1 := \sum\limits_{i=1}^r z_i^2$ and $Z_2 := \sum\limits_{\ell=1}^r z_{d - \ell + 1}^2$. Note $Z_1, Z_2 \overset{\text{iid}}{\sim} \chi^2_r$ random variables, and $|Z_1 - Z_2| = \max\{Z_1, Z_2\} - \min\{Z_1, Z_2\}$. Substituting this lower bound into \eqref{eq:expect-abs-norm-diff} yields
    \begin{align*}
        \eE\Big[ \big| \|\x\|^2 - \|\y\|^2 \big| \Big] &\geq \sin(\theta_1) \eE\Big[ |Z_1 - Z_2| \Big] = \sin(\theta_1) \Big( \eE\big[ \max\{Z_1, Z_2\} \big] - \eE\big[ \min\{Z_1, Z_2\} \big] \Big) \\
        &\overset{(d)}{=} \frac{4\sin(\theta_1)}{\sqrt{\pi}} \frac{\Gamma((r+1)/2)}{\Gamma(r/2)},
    \end{align*}
    where $(d)$ is from \Cref{lem:expec-max-min-chi-squares}. We can then lower bound $\frac{\Gamma((r+1)/2)}{\Gamma(r/2)}$ as such. First, let $x := r/2$. Then, by Wendel's Inequality,
    \begin{align*}
        &\frac{\Gamma(x + 1/2)}{x^{1/2}\Gamma(x)} \geq \bigg( \frac{x}{x + 1/2} \bigg)^{1/2} \iff \frac{\Gamma((r + 1)/2)}{\Gamma(r/2)} \geq \frac{r}{\sqrt{2(r+1)}},
    \end{align*}
    so
    \begin{equation*}
        \eE\Big[ \big| \|\x\|^2 - \|\y\|^2 \big| \Big] \geq \frac{4r\sin(\theta_1)}{\sqrt{2\pi(r+1)}}.
    \end{equation*}
    Substituting this lower bound into \eqref{eq:norm_ab_theta} yields
    \begin{equation*}
        \eE\big[ \|\ab\|^2 \big] \geq r + \sqrt{\frac{2}{\pi}} \cdot \frac{r\sin(\theta_1)}{\sqrt{r+1}} \implies \eE\big[ \ab \ab^\top \big] \succeq \Bigg(1 + \sqrt{\frac{2}{\pi}} \cdot \frac{\sin(\theta_1)}{\sqrt{r+1}}\Bigg) \I_r.
    \end{equation*}
    We can then use the fact $\eE\big[ \|\ab\|^2 + \|\blb\|^2 \big] = 2r$ to show
    \begin{equation*}
         \Bigg( 1 - \frac{1}{r} \cdot \sqrt{\sum\limits_{\ell=1}^r \sin^2(\theta_\ell)} \Bigg) \I_r \preceq \eE\big[ \blb \blb^\top \big] \preceq \Bigg(1 - \sqrt{\frac{2}{\pi}} \cdot \frac{\sin(\theta_1)}{\sqrt{r+1}}\Bigg) \I_r,
    \end{equation*}
    which completes the proof.
\end{proof}

\subsection{Matrix Bernstein's Inequality}
\vspace{-0.15cm}
We use Bernstein's matrix inequality to bound the largest and smallest eigenvalues of sums of independent, random symmetric matrices.
\begin{lemma}[Bernstein's inequality, adapted from Theorem 6.2 in \cite{tropp2012user}]
\label{lem:bernstein-inequality}
    Let $\X_1, \dots, \X_n$ be independent random symmetric matrices of dimension $m$. Assume that there exist a positive number $R$ and matrices $\A_i$ such that
    \begin{equation*}
        \eE[\X_i^p] \preceq \frac{p!}{2} \cdot R^{p-2} \cdot \A_i^2 
    \end{equation*}
    for all $i \in [n]$ and integers $p \geq 2$. Then, for all $t \geq 0$:
    \begin{equation*}
        P\Bigg( \lambda_1\bigg( \sum\limits_{i=1}^n \X_i - \eE[\X_i] \bigg) \geq t \Bigg) \leq m \cdot \exprm\bigg( -\frac{t^2}{2(\sigma^2 + Rt)} \bigg),
    \end{equation*}
    where $\sigma^2 = \sigma_1\bigg( \sum\limits_{i=1}^n \A_i^2 \bigg)$.
\end{lemma}
\noindent We refer to the condition $\eE[\X_i^p] \preceq \frac{p!}{2} \cdot R^{p-2} \cdot \A_i^2$ as \emph{Bernstein's condition}. We show $\ab\ab^\top$ and $\blb\blb^\top$ satisfy Bernstein's condition.
\begin{lemma}
\label{lem:aa-bb-theta-bernstein-condition}
    Let $\w \sim \calN(\0_d, \I_d)$ $\x := \U_1^\top \w$, $\y := \U_2^\top \w$, $\ab \sim \x \: \big| \: \|\x\|^2 > \|\y\|^2$, and $\blb \sim \y \: \big| \: \|\x\|^2 > \|\y\|^2$. Then, we have
    \vspace{-0.25cm}
    \begin{align*}
        &\eE\big[ (\ab \ab^\top)^{^p} \big] \preceq \frac{p!}{2} \cdot (2r)^{p-2} \cdot 8r^2 \I_r, \; \; \text{and} \; \; \eE\big[ (\blb \blb^\top)^{^p} \big] \preceq \frac{p!}{2} \cdot (2r)^{p-2} \cdot 8r^2 \I_r
    \end{align*}
    for all integers $p \geq 1$.
\end{lemma}
\begin{proof}
    We first focus on $\eE\big[(\ab \ab^\top)^{^p}\big]$. It suffices to upper bound $\lambda_1\Big(\eE\big[ (\ab \ab^\top)^{^p} \big] \Big)$:
    \begin{equation*}
        \lambda_1\Big(\eE\big[ (\ab \ab^\top) \big] \Big) \overset{(a)}{\leq} \eE\Big[ \lambda_1\big( (\ab \ab^\top)^{^p} \big) \Big] \overset{(b)}{=} \eE\big[ (\|\ab\|^2)^{^p} \big],
    \end{equation*}
    where $(a)$ is due to Jensen's inequality, and $(b)$ is because $(\ab \ab^\top)^{^p}$ is a rank-$1$ matrix for all integers $p \geq 1$. Recall $\|\ab\|^2 \sim \maxrm\{X, Y\}$, where $X := \|\x\|^2$ and $Y := \|\y\|^2$. Therefore,
    \begin{align*}
        &\eE\big[ (\|\ab\|^2)^{^p} \big] =  \int\limits_0^\infty \int\limits_0^\infty \maxrm\{x, y\}^p f_{X, Y}(x, y) \: dx \: dy = \int\limits_0^\infty \int\limits_0^\infty \maxrm\{x^p, y^p\} f_{X, Y}(x, y) \: dx \: dy \\
        &= 2 \int\limits_0^\infty \int\limits_{y}^\infty x^p f_{X, Y}(x, y) \: dx \: dy \overset{(a)}{\leq} 2 \int\limits_0^\infty \int\limits_0^\infty x^p f_{X | Y}(x | y) f_{Y}(y) \: dx \: dy = 2 \int\limits_0^\infty \eE_{X \sim \chi^2_r}\big[ X^p \: | \: Y \big] f_{Y}(y) \: dy \\ &= 2 \eE_{Y \sim \chi^2_r} \Big[ \eE_{X \sim \chi^2_r} \big[ X^p \: | \: Y \big] \Big]
        = 2 \eE\big[ X^p \big] \overset{(b)}{\leq} p! (2r)^p = \frac{p!}{2} \cdot (2r)^{p-2} \cdot 8r^2,
    \end{align*}
    where $(a)$ is because $X$ and $Y$ have non-negative support, and $(b)$ is from Lemma A.6 in \cite{qu2014finding}. Therefore:
    \begin{equation*}
        \eE\big[ (\ab \ab^\top)^{^p} \big] \preceq \frac{p!}{2} \cdot (2r)^{p-2} \cdot 8r^2 \I_r = \frac{p!}{2} \cdot R_a^{p-2} \cdot \A^2, 
    \end{equation*}
    where $R_a = 2r$ and $\A^2 = 8r^2 \I_r$. We can bound $\eE\big[(\blb \blb^\top)^{^p}\big]$ in a similar manner to obtain:
    \begin{equation*}
        \eE\big[ (\blb \blb^\top)^{^p} \big] \preceq \frac{p!}{2} \cdot (2r)^{p-2} \cdot 8r^2 \I_r = \frac{p!}{2} \cdot R_b^{p-2} \cdot \B^2, 
    \end{equation*}
    where $R_b = 2r$ and $\B^2 = 8r^2 \I_r$.
\end{proof}

    \section{Proof of Theorem~\ref{thm:binary-lin-sep}} \label{app:thm-1-proof}
\vspace{-0.25cm}
We now provide the full proof of \Cref{thm:binary-lin-sep}. For ease of exposition, let $\X := \W \U_1$ and $\Y := \W \U_2$, and $\x_n$ and $\y_n$ denote the $n^{th}$ row in $\X$ and $\Y$, respectively, written as column vectors. Note $\x_n = \U_1^\top \w_n$ and $\y_n = \U_2^\top \w_n$, where $\w_n \sim \calN(\0_d, \I_d)$. 

\subsection{Conditions for Linear Separability} \label{ssec:lin-sep-conditions}
\vspace{-0.15cm}
We first identify necessary and sufficient conditions to achieve linear separability between $f(\calS_1)$ and $f(\calS_2)$. By definition of linear separability, we aim to show there exists a $\v \in \reals^D$ such that \eqref{eq:lin-sep-problem} holds for all $\bm \alpha \in \reals^r \setminus \{\0_r\}$. Focusing only on $\U_1$, we can re-write \eqref{eq:lin-sep-problem} under \Cref{assum:network} as such:
\begin{align*}
    \v^\top &f\big( \U_1 \bm \alpha \big) = \sum\limits_{n=1}^D v_n (\w_n^\top \U_1 \bm \alpha)^2 = \sum\limits_{n=1}^D v_n (\w_n^\top \U_1 \bm \alpha) (\w_n^\top \U_1 \bm \alpha) \\
    &= \sum\limits_{n=1}^D v_n (\bm \alpha^\top \U_1^\top \w_n)(\w_n^\top \U_1 \bm \alpha) = \bm \alpha^\top \bigg( \sum\limits_{n=1}^D v_n \U_1^\top \w_n \w_n^\top \U_1 \bigg) \bm \alpha \\
    &= \bm \alpha^\top \bigg( \sum\limits_{n=1}^D v_n \x_n \x_n^\top \bigg) \bm \alpha > 0 \iff \sum\limits_{n=1}^D v_n \x_n \x_n^\top \succ 0. \label{eq:X-outer-pd}
\end{align*}
We can re-write the $\U_2$ part of \eqref{eq:lin-sep-problem} similarly to obtain the following necessary and sufficient conditions for linear separability:
\begin{equation} \label{eq:suff-nec-conditions}
    \sum\limits_{n=1}^D v_n \x_n \x_n^\top \succ 0 \; \; \text{and} \; \; \sum\limits_{n=1}^D v_n \y_n \y_n^\top \prec 0.
\end{equation}

\noindent We then construct the linear classifier $\v$ with the following entries:

\smallskip

\begin{center}
    \textit{For all $n \in [D]$, $v_n = \mathrm{sign}\big( \|\x_n\|^2 - \|\y_n\|^2 \big)$.}
\end{center}

\smallskip

\noindent With this choice of $\v$, \eqref{eq:suff-nec-conditions} becomes
\begin{equation*} 
    \mS_1 := \sum\limits_{i \in \calI} \x_i \x_i^\top - \sum\limits_{j \in \calI^c} \x_j \x_j^\top \succ 0 \; \; \text{and} \; \; \mS_2 := \sum\limits_{i \in \calI} \y_i \y_i^\top - \sum\limits_{j \in \calI^c} \y_j \y_j^\top \prec 0,
\end{equation*}
where $\calI := \{n \in [D]: v_n = +1\}$ and $\calI^c := \{n \in [D]: v_n = -1\}$. We now upper bound the failure probability $P\Big( \mS_1 \not \succ 0 \cup \mS_2 \not \prec 0 \Big)$.

\subsection{Bounding the Failure Probability}

We aim to upper bound $P\Big(\mS_1 \not \succ 0 \cup \mS_2 \not \prec 0\Big)$ by some (arbitrarily) small $\delta \in (0, 1)$.
We first upper bound $P\Big(\mS_1 \not \succ 0\Big)$ and $P\Big(\mS_2 \not \prec 0\Big)$ individually. Let $\gamma_1 := \sqrt{\frac{2}{\pi}} \cdot \frac{\sin(\theta_1)}{\sqrt{r+1}}$ and $\gamma_2 := \frac{1}{r} \cdot \sqrt{\sum\limits_{\ell=1}^r \sin^2(\theta_\ell)}$. Also let $\alpha_1 := 1 + \gamma_1$, $\alpha_2 := 1 + \gamma_2$, $\beta_1 := 1 - \gamma_1$, and $\beta_2 := 1 - \gamma_2$. 

We first upper bound $P\Big(\mS_1 \not \succ 0\Big).$ Note $\mS_1 \not \succ 0$ if and only if $\lambda_r(\mS_1) \leq 0$. By Lemma~\ref{lem:aa-bb-theta-bernstein-condition}, $\mS_1$ and $\mS_2$ are sums of random matrices that satisfy Bernstein's condition. Therefore, we can upper bound $P\Big(\mS_1 \not \succ 0\Big) = P\Big(\lambda_r(\mS_1) \leq 0\Big)$ using Bernstein's inequality:
\begin{align}
    P&\Big(\Sb_1 \not \succ 0\Big) = P\Big(\lambda_r(\Sb_1) \leq 0\Big) \nonumber = P\Big(\lambda_r(\Sb_1) - \lambda_r\big( \eE[\Sb_1] \big) \leq -\lambda_r\big( \eE[\Sb_1]\big) \Big) \nonumber \\
    &\overset{(a)}{\leq} P\Big( \lambda_r\big(\Sb_1 - \eE[\Sb_1]\big) \leq -\lambda_r\big(\eE[\Sb_1]\big) \Big) \nonumber = P\Big( \lambda_1\big(-\Sb_1 - \eE[-\Sb_1]\big) \geq \lambda_r\big(\eE[\Sb_1]\big) \Big) \nonumber \\
    &\overset{(b)}{\leq} r \cdot \exprm\Bigg( -\frac{\lambda_r\big(\eE[\Sb_1]\big)^2}{16r^2D + 4r\lambda_r\big(\eE[\Sb_1]\big)} \Bigg), \label{eq:S1-bernstein}
\end{align}
where $(a)$ is due to Weyl's inequality, and $(b)$ is from Lemma~\ref{lem:bernstein-inequality}. We now upper and lower bound $\eE[\mS_1]$ as follows. First, using \Cref{lem:expec-aa-bb-theta},
\begin{equation*}
   (2Q - \beta_1)D \I_r \preceq \eE\Big[ \mS_1 \: | \: Q \Big] \preceq (2Q - \beta_2) D \I_r. 
\end{equation*}
Then, taking the expectation over $Q$ yields
\begin{equation} \label{eq:E-S1}
    \gamma_1 D \I_r \preceq \eE[\mS_1] \preceq \gamma_2 D \I_r.
\end{equation}
Therefore, $\gamma_1 D \leq \lambda_r\big(\eE[\mS_1]\big) \leq \gamma_2 D $. Substituting \eqref{eq:E-S1} into \eqref{eq:S1-bernstein} leads to
\begin{equation*}
    P\Big(\Sb_1 \not \succ 0\Big) \leq r \cdot \exprm\Bigg( -\frac{\gamma_1^2 D}{16r^2 + 4\gamma_2r} \Bigg).
\end{equation*}
By similar argument, we can show by $-\gamma_2D \I_r \preceq \eE[\mS_2] \preceq -\gamma_1D \I_r$ that
\begin{equation*}
    P\Big( \mS_2 \not \prec 0 \Big) \leq r \cdot \exprm\bigg( -\frac{\gamma_1^2 D}{16r^2 + 4\gamma_2 r} \bigg).
\end{equation*}

\smallskip


\noindent Finally, we apply the Union Bound to obtain the following upper bound on the failure probability:
\begin{equation} \label{eq:failure-prob-bound}
    P\Big(\mS_1 \not \succ 0 \cup \mS_2 \not \prec 0\Big) \leq 2r \cdot \exprm\bigg( -\frac{\gamma_1^2 D}{16r^2 + 4\gamma_2 r} \bigg).
\end{equation}

\subsection{Final Result}
\vspace{-0.15cm}
Upper bounding \eqref{eq:failure-prob-bound} by some (arbitrarily small) $\delta \in (0, 1)$, and then re-arranging the terms to lower bound $D$, results in
\begin{equation} \label{eq:D-gammas}
    D \geq \frac{16r^2 + 4\gamma_2 r}{\gamma_1^2} \cdot \log\bigg(\frac{2r}{\delta}\bigg).
\end{equation}
Substituting the definitions of $\gamma_1$ and $\gamma_2$, as well as $\theta_{min} := \theta_1$, into \eqref{eq:D-gammas} leads to our final result. Let $\delta \in (0, 1)$. Then, $\mS_1 \succ 0$ and $\mS_2 \prec 0$, and thus $f(\calS_1)$ and $f(\calS_2)$ are linearly separable, if the network width $D$ satisfies
\begin{equation} \label{eq:width-bound}
    D \geq \frac{2\pi  \Bigg(4 r^2 + \sqrt{\sum\limits_{\ell=1}^r\sin^2(\theta_\ell)} \ \Bigg)  (r+1)}{\sin^2(\theta_{min})} \cdot \log\bigg(\frac{2r}{\delta}\bigg).
\end{equation}

    \section{Additional Experimental Details}
In this section, we discuss the experimental setup and results for \Cref{fig:separability-d-r,fig:image_data_class_svals,fig:linear_probe_depth_3}.

\subsection{Phase Transition in Terms of Intrinsic Dimension} \label{ssec:phase-transition} 
In this subsection, we describe the setup and results in \Cref{fig:separability-d-r}, which verifies the required network width to achieve linear separability of the initial-layer features grows polynomially w.r.t. the intrinsic dimension. This experiment was run on a MacBook Air with an Apple M3 chip.

\paragraph{Setup.} Over 25 trials, we randomly sampled two matrices $\U_1, \U_2$ from the $d \times r$ Stiefel manifold, and a weight matrix $\W \in \reals^{D \times d}$ with iid standard Gaussian entries.
We varied the ambient dimension $d$ while keeping the intrinsic dimension $r$ fixed, and also varied $r$ while keeping $d$ fixed. In both settings, we tested different layer widths $D$. For each combination of $(D,d)$ and $(D,r)$, we checked for linear separability using the necessary and sufficient conditions \eqref{eq:lin-sep-outer-sum}, and  recorded the proportion of successful trials.

\paragraph{Results.} As seen in \Cref{fig:separability-d-r}, when $d$ increases for a fixed $r$, the values of $D$ at which the proportion of successful trials transitions from $0$ to $1$, or the \emph{phase transition}, remains constant. In contrast, as $r$ increases for a fixed $d$, this phase transition region clearly increases. Thus, \Cref{fig:separability-d-r} verifies the required width to achieve linear separability of the random features only depends on the intrinsic dimension of the subspaces.

\subsection{Singular Values of Class Data Matrices in Image Datasets}
\label{ssec:image-data-svals}
Here, we describe the setup of \Cref{fig:image_data_class_svals}. This experiment was done on a Macbook Air with an Apple M3 chip. 

\paragraph{Setup.} We considered the MNIST, Fashion MNIST \citep{xiao2017fashion}, and CIFAR-10 \citep{krizhevsky2009learning} datasets. In each dataset, we first flattened the images so that they are $d$-dimensional vectors, where $d$ is the number channels multiplied by the pixels. For MNIST and Fashion MNIST, $d = 1 \times 28 \times 28  = 784$, while for CIFAR-10, $d = 3 \times 32 \times 32 = 3072$. Then, for each class, we create a $d \times N$ data matrix, where $N = 5000$ is the number of data points in each class. Finally, for each class data matrix, we compute the number of singular values that account for $95\%$ and $99\%$ of the Frobenius norm in each class's data matrix.

\subsection{Dependence on Dimension and Number of Classes: Quadratic vs. ReLU} \label{sapp:quad-relu-comp}
Here, we describe the setup and results in \Cref{fig:rank-K-sweep}, which shows the required widths of quadratic and ReLU random feature models have similar dependence on the subspace dimension $r$, and number of classes $K$, to achieve linear separability of a UoS. All experiments were run on a single NVIDIA A40 GPU.

\paragraph{Setup.} For all experiments, we set $d = 128$. In \Cref{subfig:rank-sweep}, we set $K = 2$ and swept through $r$ from $2^2$ to $2^6$ by powers of $2$. In \Cref{subfig:K-sweep}, we fixed $r = 4$ and swept through the number of subspaces $K$ from $2$ to $2^5$, again by powers of $2$. In both sweeps, we varied the network width $D$ from $2^5$ to $2^{10}$ by powers of $2$.

\subsection{Linear Separability of Features: Random vs. Trained Weights} \label{sapp:random_trained_lin_sep} 
We first describe the setup and results in \Cref{fig:linear_probe_depth_3}, which investigates how training the network weights away from their random initialization impacts the linear separability of the initial-layer features. Here, all experiments were run on a single NVIDIA V100 GPU.

\paragraph{Setup.} We first created a training set using the above data generation process with $K = 2$, $d = 16$, and $r = 4$. We then trained two 3-layer MLPs of width $D=128$ for $100$ epochs. One MLP had ReLU activations, and the other had quadratic activations. After each training epoch, we performed a linear probing on the features extracted by the two hidden layers. At initialization (marked by a star), all weights were sampled i.i.d. from a zero-mean Gaussian distribution. We averaged the results over $5$ trials. 

\paragraph{Results.} Across all $5$ trials in both MLPs, the features from the hidden layers were linearly separable at random initialization, as evidenced by the perfect linear probing accuracy. After each epoch, the linear probe accuracy remained perfect, implying the features from the hidden layers remained linearly separable during training. Thus, training the weights away from the random initialization does not impact the linear separability of the features.
\fi 


\ifaistats
    \runningtitle{Linearly Separable Features in Shallow Nonlinear Networks}
    \runningauthor{Alec S. Xu, Can Yaras, Peng Wang, Qing Qu}
    
    \twocolumn[
    
    \aistatstitle{Linearly Separable Features in Shallow Nonlinear Networks: Width Scales Polynomially with Intrinsic Data Dimension}

    \aistatsauthor{ Alec S. Xu \And Can Yaras \And  Peng Wang \And Qing Qu }

    \aistatsaddress{University of Michigan \And  University of Michigan \And University of Macao \And University of Michigan } ]

    \begin{abstract}
\vspace{-0.3cm}
     Deep neural networks have attained remarkable success across diverse classification tasks. Recent empirical studies have shown that deep networks learn features that are linearly separable across classes. However, these findings often lack rigorous justifications, even under relatively simple settings. In this work, we address this gap by examining the linear separation capabilities of shallow nonlinear networks. Specifically, inspired by the low intrinsic dimensionality of image data, we model inputs as a union of low-dimensional subspaces (UoS) and demonstrate that a single nonlinear layer can transform such data into linearly separable sets. Theoretically, we show that this transformation occurs with high probability when using random weights and quadratic activations. Notably, we prove this can be achieved when the network width scales polynomially with the intrinsic dimension of the data rather than the ambient dimension. Experimental results corroborate these theoretical findings and demonstrate that similar linear separation properties hold in practical scenarios beyond our analytical scope. This work bridges the gap between empirical observations and theoretical understanding of the separation capacity of nonlinear networks, offering deeper insights into model interpretability and generalization.
    
\end{abstract}
    \section{INTRODUCTION} \label{sec:intro}
Over the past decade, deep neural networks (DNNs) have achieved state-of-the-art performance in a wide range of applications, including computer vision \citep{simonyan2015very, he2016deep} and natural language processing \citep{sutskever2014sequence, vaswani2017attention}. However, despite recent advances \citep{jacot2018neural, mei2018mean, ji2019gradient, arora2018convergence, lampinen2019analytic, papyan2020prevalence, zhu2021geometric,yaras2022neural,zhou2022optimization}, the
theoretical understanding of their empirical success is still primitive, even for relatively basic tasks. For example, in classification problems, the success of deep learning is often attributed to its ability to learn discriminative features that exhibit strong inter-class separation \citep{papyan2020prevalence, alain2017understanding, rangamani2023feature, masarczyk2024tunnel,wang2025understanding,yaras2023law}. 
Despite the remarkable ability of deep networks to achieve linear separation, the underlying mechanisms by which they accomplish this—especially when the input data are initially poorly separated—remain largely unclear. Investigating this phenomenon could significantly improve the interpretability of deep learning models and provide deeper insights into their generalization capabilities. Before presenting our main contribution, we provide a brief review of the existing results --- see \Cref{app:related} for a more detailed discussion. 

\smallskip

\noindent \textbf{Empirical studies on linear separability of early-layer features.} 
Recent empirical studies investigated the role of the intermediate layers in deep nonlinear networks, \eg, \cite{alain2017understanding, ansuini2019intrinsic, recanatesi2019dimensionality, he2023law,zhang2022all,wang2025understanding,yaras2023law,masarczyk2024tunnel,li2024understanding}. These studies indicate that the shallow layers expand the features such that they become linearly separable between classes. 
For instance, in image classification, \cite{alain2017understanding, masarczyk2024tunnel, wang2025understanding} observed linear probing accuracy improves significantly across the early layers of neural networks. This implies the early layers play a critical role in achieving linear separability of the input data.

\noindent \textbf{Theoretical works on linear separability of early-layer features.} To our knowledge, there are limited theoretical studies on the linear separability of features across nonlinear layers in DNNs. Recent works \citep{dirksen2022separation, ghosal2022randomly} studied the separability of features in shallow ReLU networks. These studies rigorously showed the features extracted from a two-layer \citep{dirksen2022separation} and one-layer \citep{ghosal2022randomly} random ReLU network are linearly separable for arbitrary input data. However, a key limitation of these works is that in the worst case, the required network width grows \emph{exponentially} with respect to (w.r.t.) the ambient dimension of the data. Consequently, the network sizes required by theoretical analyses are substantially larger than those typically used in real-world applications, highlighting a gap between theory and practice.

\noindent \textbf{Theoretical studies on representation learning in deep linear networks.} Another line of research has explored how deep \emph{linear} networks (DLNs) progressively compress within-class features and discriminate between-class features \citep{saxe2019mathematical, wang2025understanding}. Building on the empirical observation that linear layers can emulate the behavior of deeper layers in nonlinear networks, \cite{wang2025understanding} provided a theoretical analysis of the progressive feature compression in DLNs, under the assumption that the input data are already linearly separable. However, due to this restrictive assumption, the study cannot fully explain the structures of hierarchical representation in nonlinear networks, particularly \emph{how} the early layers transform input features to achieve linear separability due to the nonlinear operators.




\subsection{OUR CONTRIBUTIONS} \label{ssec:contributions}
In this work, we investigate the linear separability of features in shallow nonlinear networks for data with low intrinsic dimensions.
Specifically, we show
\begin{tcolorbox}[colframe = red!75!black]
\begin{center}
   \emph{a single nonlinear layer 
   transforms data from a union of low-dimensional subspaces into linearly separable sets.}
\end{center}
\end{tcolorbox}
\noindent We rigorously prove this result with $K = 2$ subspaces and discuss how the result can be extended to $K > 2$ subspaces. 
In our analysis, we assume that the activation is quadratic and the first-layer weights are random. The resulting width of the network scales \emph{polynomially} w.r.t. the intrinsic dimension of the subspaces. Moreover, our results empirically hold under more generic settings. For example, we can replace the quadratic with other activations, such as ReLU, and still achieve linear separability with similar requirements on the subspace dimensions and number of subspaces  
(see \Cref{fig:rank-K-sweep}).  

Our findings offer insights into the role of overparameterization in deep representation learning and explain why learning based upon random features can lead to good in-distribution generalization.

\begin{figure*}[t]
    \centering
    \includegraphics[width=0.36\linewidth]{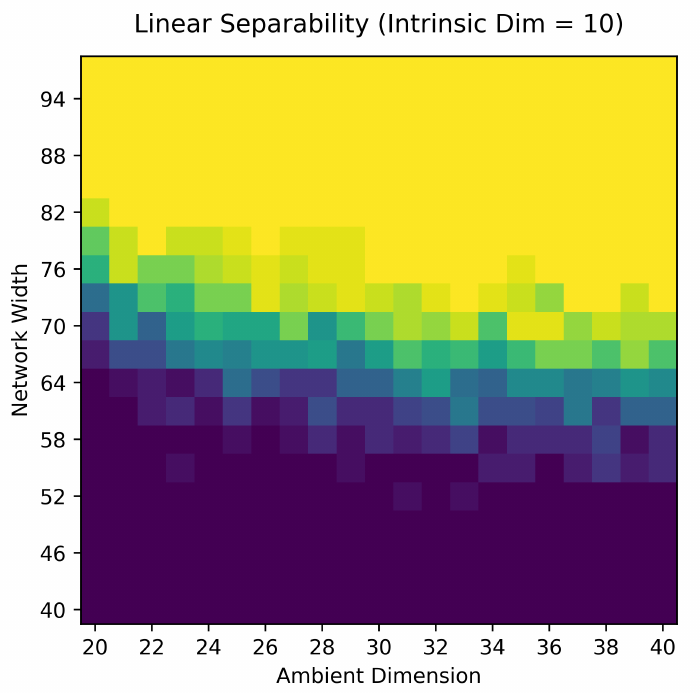}
    \hspace{0.1in}
    \includegraphics[width=0.4\linewidth]{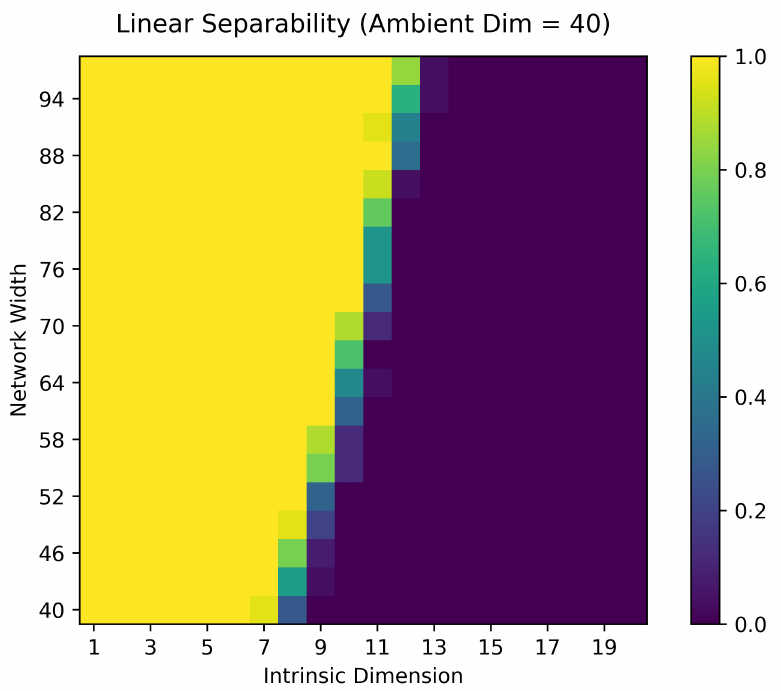}
    \caption{\textbf{Phase transition of linear separability w.r.t. dimensions $(d,r)$ and network width $D$.}  We demonstrate that the network width required to achieve linear separability of a union of two subspaces scales polynomially with the intrinsic dimension. See \Cref{ssec:phase-transition} for experimental details.}  
    \label{fig:separability-d-r}
\end{figure*}


\subsection{NOTATION AND PAPER ORGANIZATION} Before delving into the technical discussion, we introduce the notation used throughout the paper and outline its organization.

\smallskip
\noindent \textbf{Notation.} For a positive integer $N$, we use $[N]$ to denote the index set $\{1, 2, \dots, N\}$. We use $\calN(\mu, \sigma^2)$ to denote a Gaussian distribution with mean $\mu$ and variance $\sigma^2$, and $\calN(\bm{\mu}, \bm{\Sigma})$ to denote a multivariate Gaussian distribution with mean $\bm{\mu}$ and covariance $\bm{\Sigma}$. We use $\| \cdot \|$ to denote the Euclidean norm of a vector, $\bm 0_m$ to denote an $m$-dimensional vector of all zeros, $\lambda_i(\cdot)$ to denote the $i^{th}$ largest eigenvalue of a symmetric matrix, and $\sigma_i(\cdot)$ to denote the $i^{th}$ largest singular value of a matrix. With a slight abuse of notation, for some function $\phi$ and set $\calX$, $\phi \big( \calX \big)$ denotes the set $\big\{ \x \in \calX: \phi \big( \x \big) \big\}$. Unless otherwise stated, the term ``subspace'' implies a linear subspace embedded in Euclidean space.

\smallskip
\noindent \textbf{Organization.} The rest of this paper is organized as follows. We motivate the union of subspaces (UoS) 
data model, introduce our problem setting, and motivate our theoretical assumptions in \Cref{sec:prelim}. 
We then state our main theoretical results and provide a proof sketch in \Cref{sec:theoretical}, with the full proof in \Cref{app:thm-1-proof}. In \Cref{sec:empirical}, we provide empirical evidence supporting our theoretical results, and investigate settings not considered in our analysis. 
Finally, we summarize our results and conclude in \Cref{sec:conclusion}.
    \section{PRELIMINARIES} \label{sec:prelim}
In this section, we introduce the basic problem setup and motivations. First, we introduce the UoS model for our input data in \Cref{ssec:uos}, and then discuss the choices of the network in \Cref{ssec:problem}.



\subsection{ASSUMPTIONS ON INPUT DATA} \label{ssec:uos}


Recent empirical studies indicate real-world image data typically possess a significantly lower \emph{intrinsic} dimension than their ambient dimension. For instance, \cite{pope2020intrinsic} used a nearest-neighbor approach to estimate the intrinsic dimension of many popular image datasets, including MNIST \citep{lecun1998gradient}, CIFAR-10 \citep{krizhevsky2009learning}, and ImageNet \citep{russakovsky2015imagenet}. They showed the intrinsic dimension of these datasets is at most around $40$, even though the images themselves contain thousands of pixels. Furthermore, \cite{brown2023verifying} used a similar approach to show \emph{each class} has its own low intrinsic dimensionality. These results indicate image data lie on  \emph{a union of low-dimensional manifolds} within high-dimensional space. 

Although low-dimensional manifolds can exhibit complex structures, each manifold can be locally approximated by its tangent space, which is a linear subspace embedded within the ambient space. This motivates us to initiate our study with a simplified model: \textbf{a union of $K$ low-dimensional subspaces (UoS)} that capture the local structures of manifolds. Similar models have recently been explored for understanding generative models \citep{wang2024diffusion,chen2024exploring}. Furthermore, \Cref{fig:image_data_class_svals} in \Cref{app:uos-image-data} shows common image datasets \emph{approximately} lie on a UoS, where each subspace represents a different class. 

For ease of analysis and exposition, we focus on the case where $K = 2$. Nonetheless, our results extend to the case where $K > 2$, as discussed in \Cref{ssec:multiple}. To set the stage for our analysis, we introduce a generic definition of a union of $K$ subspaces.

\begin{definition}[Union of $K$ Low-Dimensional Subspaces] \label{def:UoS}
    Let $\calS_1, \calS_2, \dots \calS_K \subseteq \reals^d$ be $K$ linear subspaces with dimensions $r_1, r_2, \dots, r_K$, respectively. Let $\U_k \in \mathbb{R}^{d\times r_k}$ be an orthonormal basis of $\calS_k$ for all $k\in [K]$. 
    The union of subspaces $\calS_1, \calS_2, \dots, \calS_K$ is defined as such:
    \begin{align*}
        \bigcup\limits_{k=1}^K \calS_k := \Big\{\z \in \reals^d: \exists k \in [K], \bm \alpha \in \mathbb{R}^{r_k} \: \text{s.t.} \: \z = \U_k \bm \alpha \Big\} .
    \end{align*}
\end{definition} 


The \emph{principal angles} between two subspaces is a generalization of the angles between two vectors (i.e., two one-dimensional subspaces). For two subspaces $(\calS_1,\calS_2)$ of dimensions $r_1$ and $r_2$, there exist $\minrm\{r_1, r_2\}$
principal angles between them. These angles are formally defined as follows.
\begin{definition}[Principal angles between two subspaces]
     Suppose that the columns of $\U_1 \in \reals^{d \times r_1}$ and $\U_2 \in \reals^{d \times r_2}$ are orthonormal bases for subspaces $\calS_1$ and $\calS_2$, respectively.  Let $r := \minrm\{r_1, r_2\}$. For all $\ell \in [r]$, the $\ell^{th}$ principal angle $\theta_\ell \in [0, \pi/2]$ between $\calS_1$ and $\calS_2$ is defined as 
    \begin{equation*}
        \cos(\theta_\ell) \;:=\; \sigma_\ell(\U_1^\top \U_2).
    \end{equation*}
\end{definition}

\begin{figure}
    \centering
    \tdplotsetmaincoords{120}{50}
    \begin{tikzpicture}[scale=1.75]
    \tdplotsetrotatedcoords{90}{0}{0}
    \fill[blue!70, opacity=0.5, tdplot_rotated_coords] (-1, -0.5, 0) -- (1, -0.5, 0) -- (1, 0.5, 0) -- (-1, 0.5, 0) -- cycle; 
    
    \draw[<-, thick, orange] (-0.5, -0.75) -- (-0.17, -0.255);
    \draw[-, thick, orange, opacity=0.3] (-0.17, -0.255) -- (0, 0); 
    
    \draw[->, thick, orange] (0, 0) -- (0.5, 0.75); 
    \draw[<->, densely dotted, black] (-1, 0.45) -- (1, -0.45);
    \draw[<->, densely dotted, black] (0, 1) -- (0, -1);
    \draw[<->, densely dotted, black] (-0.75, -0.6) -- (0.75, 0.6);

    
    \node[fill=black, circle, inner sep=1pt, opacity=0.5] at (0, 0, 0) {};
    
    \coordinate (a) at (0.3, 0.45);
    \coordinate (o) at (0, 0);
    \coordinate (b) at (0.3, -0.15);
    \pic[draw, <->, "$\theta_1$", angle eccentricity=1.5]{angle = b--o--a};
    
    \node[anchor=south] at (0.6, 0.7) {$\calS_1$};
    \node[anchor=west] at (0.5, -0.6) {$\calS_2$};
    
    \end{tikzpicture}
    \caption{\textbf{The principal angle between a one-dimensional subspace $\mathcal S_1$ and two-dimensional subspace $\mathcal S_2$.}}
    \label{fig:princ-angles}
\end{figure}
\noindent The principal angle is illustrated in \Cref{fig:princ-angles}. By the above definition, since $0 \leq \theta_1 \leq \theta_2 \leq \dots \leq \theta_r \leq \pi/2$, we will sometimes use $\theta_{\min}$ to denote $\theta_1$. 
Building on these definitions, we will make the following assumption on the UoS model for our analysis in \Cref{sec:theoretical}.
\begin{assum} \label{assum:subspaces}
    There are $K = 2$ subspaces $\calS_1, \calS_2$ with equal dimensions, \ie, $r_1 = r_2 := r$. Furthermore, the principal angles between $\calS_1$ and $\calS_2$ are strictly positive, \ie, $0 < \theta_1 \leq \theta_2 \leq \dots \leq \theta_r \leq \pi/2$.
\end{assum}

\noindent We discuss \Cref{assum:subspaces} below.

\paragraph{Number of subspaces.} We assume $K=2$ subspaces to simplify both the analysis and exposition. The results can be generalized to consider $K > 2$ subspaces, which we discuss in detail in \Cref{ssec:multiple}. 

\paragraph{Subspace dimensions.} We assume equal dimensionality for each subspace for simplicity. In practice, each subspace in a UoS can have different dimensions. We believe our result can be generalized to this setting, and leave detailed analysis for future work.

\smallskip

\textbf{Principal angles between subspaces.} We assume none of the principal angles are equal to zero to ensure $\calS_1 \cap \calS_2 = \{\0_d\}$. Otherwise, it is impossible to label the non-zero points in $\calS_1 \cap \calS_2$. We note $\theta_1 > 0$ if and only if $r < d/2$. This assumption is typically satisfied in practice, as usually $r \ll d$.

\begin{figure}[t]
    \centering
    \tdplotsetmaincoords{120}{50}
    \begin{tikzpicture}[scale=1.2]
    
    \begin{scope}[tdplot_main_coords]
        \tdplotsetrotatedcoords{0}{90}{90}
        \fill[orange!70, opacity=0.5, tdplot_rotated_coords] (-1, -0.4, 0) -- (1, -0.4, 0) -- (1, 0.4, 0) -- (-1, 0.4, 0) -- cycle;
    
        \tdplotsetrotatedcoords{0}{0}{0}
        \fill[blue!70, opacity=0.5, tdplot_rotated_coords] (-1, -0.5, 0) -- (1, -0.5, 0) -- (1, 0.5, 0) -- (-1, 0.5, 0) -- cycle;

        \draw[<->, densely dotted, black] (0, 0, 1) -- (0, 0, -1);
        \draw[<->, densely dotted, black] (0, 1.5, 0) -- (0, -1.5, 0);
        \draw[<->, densely dotted, black] (-1.5, 0, 0) -- (1.5, 0, 0); 
    \end{scope}

    \node[fill=black, circle, inner sep=1pt, opacity=0.5] at (0, 0, 0) {};
    
    \node[anchor=south] at (0.5, 0.4) {$\calS_1$};
    \node[anchor=west] at (0.7, -0.2) {$\calS_2$};
    
    \draw[thick,->] (1.5, 0, 0) -- (2.5, 0, 0) node[midway, above] {$f(\cdot)$};
    
    \begin{scope}[xshift=3.3cm, yshift=-0.3cm, scale=0.7]
        \fill[blue!70, opacity=0.7,scale=1.5] plot[smooth cycle, tension=1] coordinates {(0,0) (0.8,0.4) (1,1) (0.2,1.2) (-0.4,0.8)};
        \node at (0.5, 0.9) {$f(\calS_1)$};
    
        \fill[orange!70, opacity=0.7,scale=1.5,xshift=-0.5cm] plot[smooth cycle, tension=1] coordinates {(2,-0.3) (2.7,0) (2.5,0.8) (1.8,0.6)};
        \node at (2.6, 0.3) {$f(\calS_2)$};
        
        \draw[dashed, thick, xshift=0.9cm, yshift=-0.5cm, rotate=-30] (0, 0) -- (0, 2.5);
    \end{scope}
    
    \end{tikzpicture}
    \caption{\textbf{An illustration of \Cref{prob:binary}.} We aim to find conditions on $f$ so a union of subspaces (left) transforms into linearly separable sets (right).}
    \label{fig:lin-sep}
\end{figure}

\subsection{LINEAR SEPARABILITY OF A UOS VIA NONLINEAR NETWORKS} \label{ssec:problem} 


In this work, we investigate how nonlinear neural networks separate the data that follows the UoS model.  Specifically, we consider a shallow neural network $f_{\W}(\bm x): \mathbb R^d \mapsto \mathbb R^D$, which is a feature mapping from the input space $\mathbb R^d$ to a feature space $\mathbb R^D$:
\begin{align}\label{eq:func-NN}
    f_{\W}(\bm x) = \sigma (\bm W \bm x ).
\end{align}
Here, $\W \in \reals^{D \times d}$ is the weight matrix, and $\sigma(\cdot)$ is an entry-wise nonlinear activation function. 
As illustrated in \Cref{fig:lin-sep}, based on the above setup, we are interested in the following problem:
\begin{problem} \label{prob:binary}
   Consider a union of two subspaces $\calS_1$ and $\calS_2$ that satisfy \Cref{assum:subspaces}. Under what conditions does there exist a separating hyperplane $\v \in \reals^D$ such that
    \begin{equation} \label{eq:lin-sep-problem}
        \v^\top f_{\W}\big( \U_1 \bm \alpha \big) > 0 \; \; \text{and} \; \; \v^\top f_{\W}\big( \U_2 \bm \alpha \big) < 0
    \end{equation}
    for all $\bm \alpha \in \reals^r \setminus \{\0_r\}$? 
\end{problem}




\smallskip

\noindent 
\Cref{prob:binary} remains challenging even under the simplified setup considered here. Generally, data from two distinct subspaces are not inherently linearly separable, and neither a linear mapping nor a nonlinear activation alone are sufficient to transform such data into linearly separable sets --- see \Cref{app:problem} for a more detailed discussion. Thus, to tackle \Cref{prob:binary}, we must \emph{jointly} apply a linear mapping and a nonlinear transformation to achieve linear separability of the subspaces. Specifically, we now introduce the following assumptions on the network \eqref{eq:func-NN}, based upon which we characterize the sufficient conditions for achieving linear separability in \Cref{sec:theoretical}.
\begin{assum} \label{assum:network}
    For the mapping $f_{\W}(\bm x) $ in \eqref{eq:func-NN}, we assume that the activation function $\sigma(\cdot)$ is the quadratic (entry-wise square) function, and the entries of $\W$ are independent and identically distributed (iid) standard Gaussian, \ie, $W_{ij} \overset{\text{iid}}{\sim} \calN(0, 1)$ for all $(i, j) \in [D] \times [d]$. 
\end{assum}

\noindent 
We briefly discuss \Cref{assum:network} below.

\vspace{-0.3cm}

\paragraph{Quadratic activation.} In this work, we consider the quadratic activation due to its smoothness and simplicity. Such activations have also been considered in many previous theoretical results of analyzing nonlinear networks, \eg, \cite{li2018algorithmic, soltanolkotabi2018theoretical, du2018power,  sarao2020optimization,gamarnik2024stationary}  --- see \Cref{app:related} for a more detailed discussion. Moreover, we believe the results approximately hold for several other nonlinear activations, such as ReLU. In \Cref{sec:empirical}, we empirically show if one replaces the quadratic activation with other activations, the output features from \eqref{eq:func-NN} are still linearly separable under a UoS data model. We also observe the required width to achieve linear separability scales similarly with the intrinsic dimension and the number of classes under both ReLU and quadratic activations --- see \Cref{fig:rank-K-sweep}. 

\vspace{-0.3cm}

\paragraph{Random weights.} \Cref{assum:network}  yields a random feature model, which has been widely studied in the literature, \eg, \citep{rahimi2007random, rahimi2008weighted, rudi2017generalization, bach2017equivalence, li2021towards} (see \citep{liu2021random} for a survey). 
    Moreover, it can also shed light on trained DNNs. For example, in the infinite-width limit \citep{jacot2018neural, arora2019exact, cao2019generalization, allen2019convergence} 
    random networks behave similarly to fully-trained networks. This is called the Neural Tangent Kernel (NTK) \citep{jacot2018neural} regime, where the random initialization determines the NTK, which remains constant during training \citep{jacot2018neural}. Furthermore, the Neural Network Gaussian Process kernel (NNGP) \citep{lee2018deep} is the kernel associated with a network at random initialization. Recently, \citep{kothapalli2024kernel} studied Neural Collapse (NC) \citep{papyan2020prevalence} of nonlinear networks from a kernel perspective. They showed that NNGP and NTK exhibited similar amounts of NC.

    Additionally, for finite-width networks, we empirically observe if the initial-layer features under a UoS data model are linearly separable at random initialization, pushing the layer weights away from their randomly initialized values via training does \emph{not} impact the linearly separability of these features --- see \Cref{fig:linear_probe_depth_3}. Thus, studying the linear separability of the features from random layers provides insight into the linear separability of the features from trained layers in finite-width networks. 

\begin{figure*}
    \centering
    \begin{subfigure}[t]{0.49\textwidth}
        \centering
        \includegraphics[width=\linewidth]{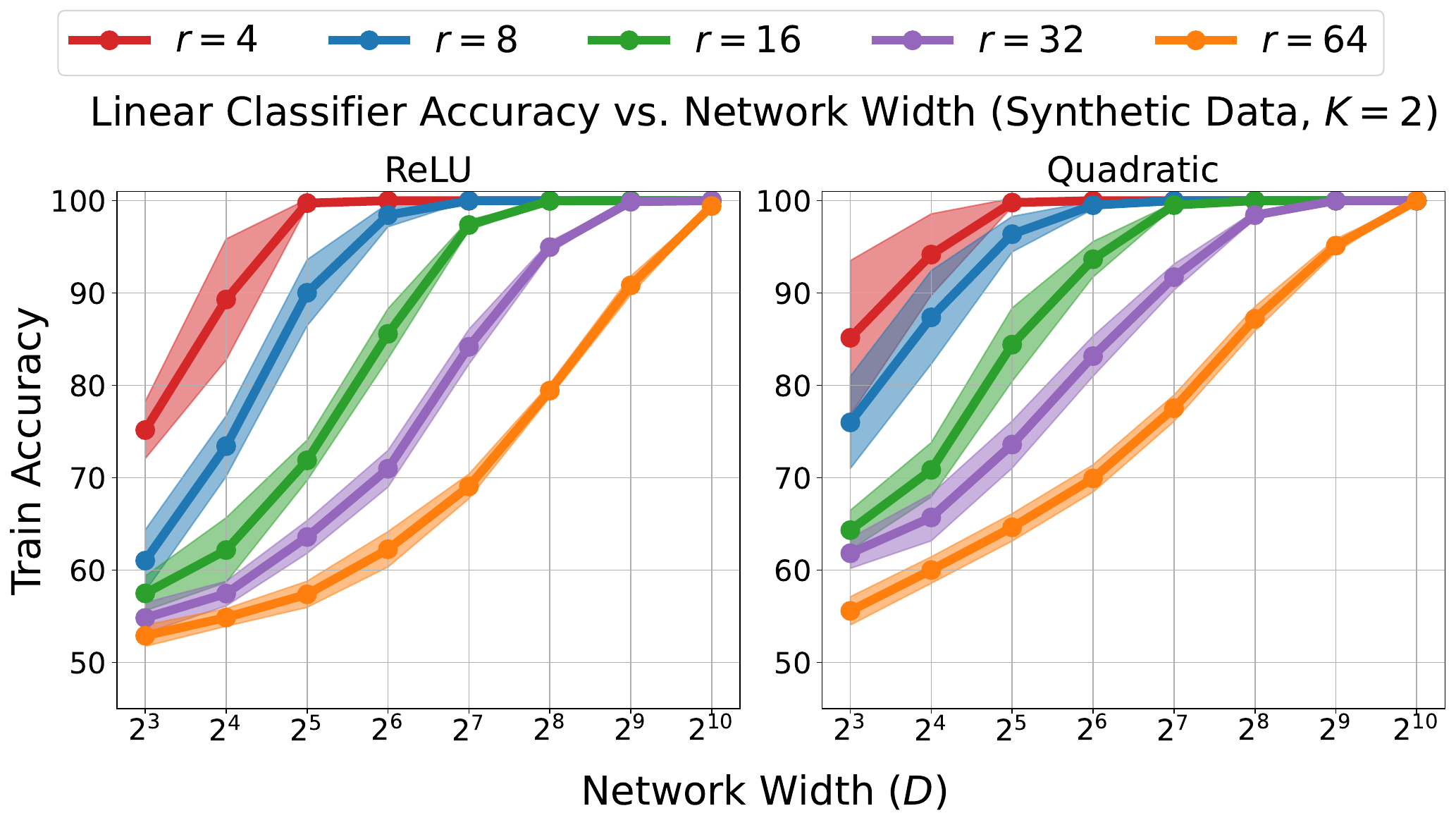}
        \caption{Sweeping $r \in \{4, 8, 16, 32, 64\}$.}
        \label{subfig:rank-sweep}
    \end{subfigure}\hfill
    \begin{subfigure}[t]{0.49\textwidth}
        \centering
        \includegraphics[width=\linewidth]{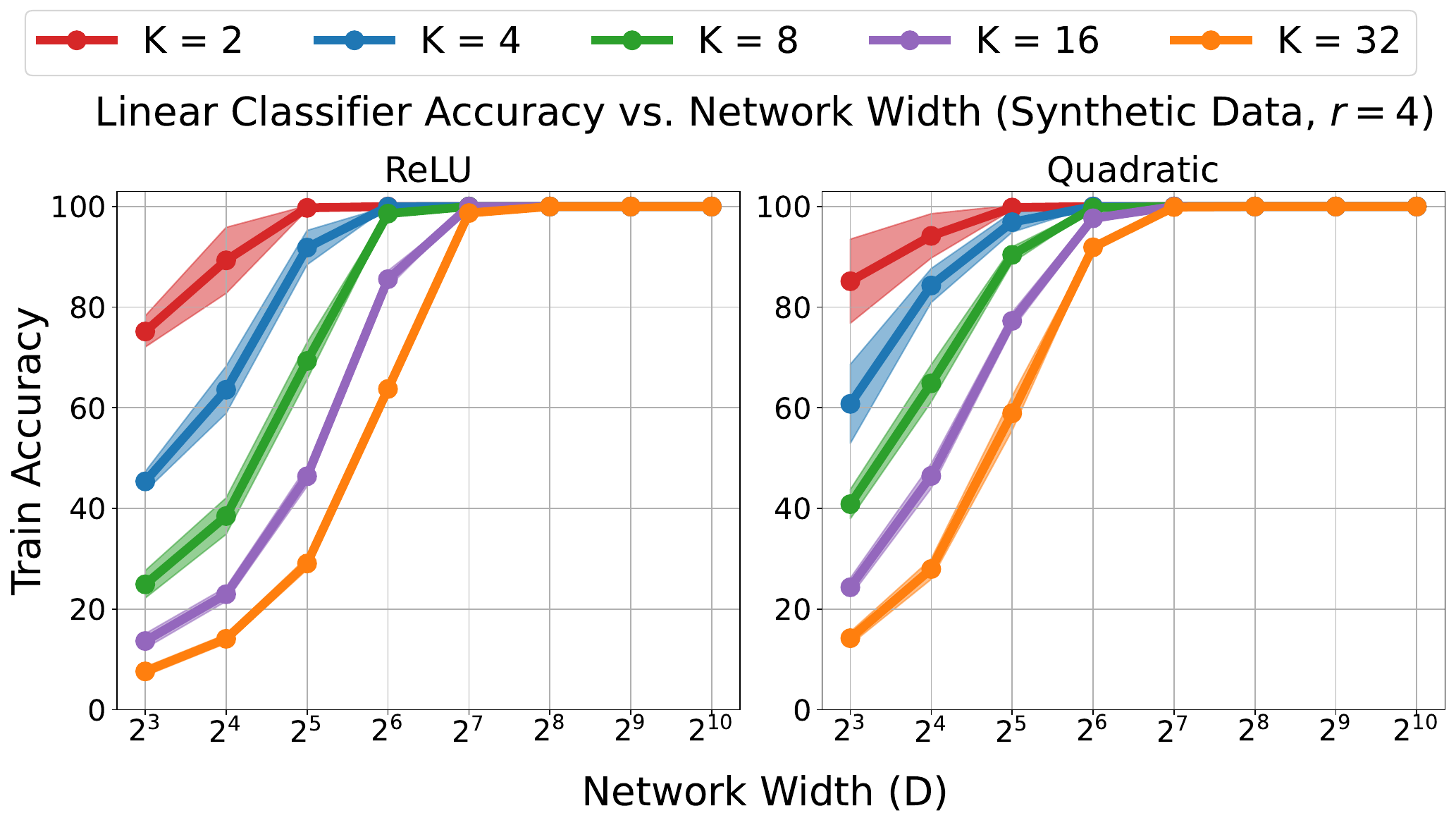}
        \caption{Sweeping $K \in \{2, 4, 8, 16, 32\}$.}
        \label{subfig:K-sweep}
    \end{subfigure}
    \caption{\textbf{ReLU vs. quadratic layers for linear separability.} ReLU and quadratic activations exhibit similar width requirements w.r.t. the intrinsic dimension (left) and number of subspaces (right) for achieving linear separability. See \Cref{sapp:quad-relu-comp} for experimental details.} 
    \label{fig:rank-K-sweep}
\end{figure*}

\begin{figure}[ht]
    \centering
    \includegraphics[width=\linewidth]{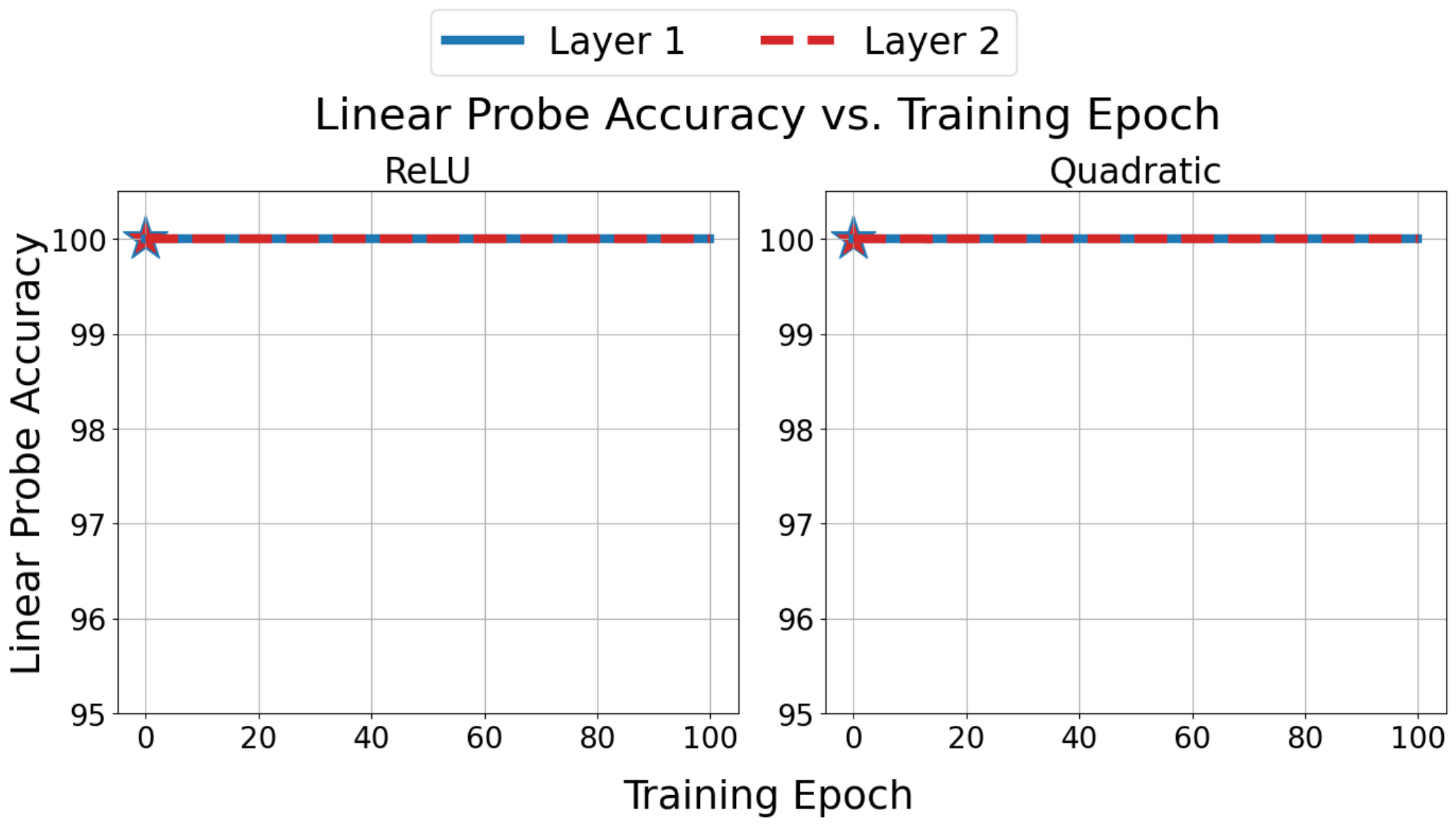}
    \caption{\textbf{Linear separability of hidden-layer features throughout training.} If the initial-layer features are linearly separable at random initialization, they remain linearly separable throughout training. See \Cref{sapp:random_trained_lin_sep} for experimental details.}
    \label{fig:linear_probe_depth_3}
\end{figure}

\section{THEORETICAL RESULTS} \label{sec:theoretical}
We first state our main theoretical results and their implications in \Cref{ssec:theorem,ssec:multiple}, and correspondingly provide a sketch of the proof in \Cref{ssec:proof-sketch}.



\subsection{MAIN RESULTS: $K = 2$ SUBSPACES} \label{ssec:theorem}
First, we state our main theoretical result in the binary case $K=2$. 
\begin{theorem}[Linear Separability of $f(\calS_1)$ and $f(\calS_2)$] \label{thm:binary-lin-sep}
    Suppose Assumptions~\ref{assum:subspaces} and \ref{assum:network} hold, and let $\delta \in (0, 1)$. If the network width $D$ satisfies
    \begin{equation} \label{eq:D-bound}
       D \geq \mathcal{O}\left( \frac{r^3}{\sin^2(\theta_{min})} \cdot \log\left(\frac{r}{\delta}\right) \right), 
    \end{equation}
    then $f(\calS_1)$ and $f(\calS_2)$ are linearly separable with probability at least $1 - \delta$ w.r.t. the randomness of $\W$.
\end{theorem}
\Cref{thm:binary-lin-sep} states that the random feature model in \eqref{eq:func-NN} transforms a two subspaces into linearly separable sets, given that the network width scales \emph{polynomially} with the subspaces' intrinsic dimension. We discuss the implications of our result below.

\smallskip

\noindent \textbf{Network width.} 
Our result indicates fewer neurons are needed to achieve linear separability of early-layer features compared to previous studies. Specifically, \citep{dirksen2022separation,ghosal2022randomly} showed one-layer and two-layer random-ReLU networks make arbitrarily structured, nonlinearly separated classes linearly separable. However, under our data model, their network widths scale \emph{exponentially} with the intrinsic dimension. These scaling requirements are much larger than those used in practical DNNs, limiting their applicability. In contrast, \Cref{thm:binary-lin-sep} requires network widths to scale \emph{polynomially} with the intrinsic dimension, aligning more closely with real-world network sizes. For example,  \Cref{fig:image_data_class_svals} in \Cref{app:uos-image-data} shows the CIFAR-10 dataset \citep{krizhevsky2009learning} approximately satisfies a UoS model, where each class subspace is of rank on the order of $10^1$. Previous results \citep{dirksen2022separation,ghosal2022randomly} require a network width on the order of $\exp(10^1)$, whereas our theorem requires a width on the order of $10^3$, which more closely aligns with network sizes used in practice. Therefore, our results provide a more accurate characterization of how the early layers in practical DNNs make low-dimensional data  linearly separable.

\paragraph{Connection to NTK-based results.} Previous work \citep{du2019gradient,huang2020dynamics} leveraged NTK approaches to show global convergence of gradient descent in overparameterized two-layer networks. In their analyses, they showed sufficiently wide networks remain close to their random initializations during training, which is closely related to our random feature model in \Cref{assum:network}. With $N$ training samples, \citet{du2019gradient, huang2020dynamics} showed for two-layer networks of widths $\mathcal{O}\left( \operatorname{poly}(N) \right)$, gradient descent converges to zero training loss, implying perfect linear classifier accuracy under a classification setting. In comparison, our work and \citet{dirksen2022separation,ghosal2022randomly} consider separating \emph{infinitely many points} in the underlying class sets, which are linear subspaces in our case. Under this setting, results from \citet{du2019gradient,huang2006extreme} require \emph{infinitely} wide layers achieve linear separability. However, our result is \emph{independent} of the number of data points, and only depends polynomially on the intrinsic dimension. 

\paragraph{Overparameterization in representation learning.} DNNs are often \emph{overparameterized}, \ie, the number of parameters is larger than the number of training samples $N$. \Cref{thm:binary-lin-sep} states that with probability at least $1 - \delta$, a layer with $\mathcal{O}\left(dr^3 \cdot \log(1 / \delta) \right)$ parameters transforms two subspaces (so an infinite number of data points) into linearly separable sets. In practice, one has a finite number of data points $N$ to classify. In a finite-data setting where the data points lie on a union of two subspaces, by setting the failure probability $\delta = \frac{1}{N}$, a layer with $\mathcal{O}\left(dr^3 \cdot \log(N)\right)$ parameters correctly classifies the data points with probability at least $1 - \frac{1}{N}$. For large $N$, the number of parameters needed to correctly classify all $N$ data points is much smaller than $N$ itself. This implies an \emph{underparameterized} network suffices in separating the data by class, and that overparameterization may serve other purposes in representation learning, e.g., feature compression \citep{wang2025understanding}.


\smallskip

\noindent \textbf{In-distribution generalization.} Our result also provides insight into in-distribution generalization when learning with random features. \Cref{thm:binary-lin-sep} states a single random nonlinear layer makes \emph{all points} in the two subspaces linearly separable. 
Suppose we have a dataset with $N$ train samples lying on a union of two subspaces, apply the random feature map \Cref{eq:func-NN} with $\mathcal{O}(r^3 \cdot \log(Nr))$ features, and train a linear classifier on the random features to classify the train samples. If the test samples lie in the same subspaces as the train samples, then the classifier will also achieve perfect test accuracy with probability at least $1 - \frac{1}{N}$. 

\subsection{EXTENSION TO $K > 2$ SUBSPACES} \label{ssec:multiple}
We now generalize the result from \Cref{thm:binary-lin-sep} to consider $K > 2$ subspaces.
\begin{corollary} \label{cor:K-lin-sep}
    Suppose there are $K > 2$ subspaces each of dimension $r$, where $(K-1)r < d/2$. For all $k \in [K]$, let $\tilde{\calS}_k \supset \calS_k$ and $\overline{\calS}_k \supset \bigcup_{j \in [K], j \neq k} \calS_j$ be $\tilde{r}$-dimensional subspaces with principal angles $\theta_{k, 1}, \theta_{k, 2}, \dots, \theta_{k, \tilde{r}}$ that satisfy \Cref{assum:subspaces}, where $\tilde{r} := (K-1)r$. Also let \Cref{assum:network} hold and $\delta \in (0,1)$. If the network width $D$ satisfies
    \begin{equation}
        D \geq \max\limits_{k \in [K]} \mathcal{O} \left( \frac{\tilde{r}^3}{\sin^2(\theta_{k, min})} \cdot \log \left( \frac{\tilde{r}}{\delta} \right) \right) 
    \end{equation}
     then for all $k \in [K]$, the sets $f(\calS_k)$ and $f\Big( \bigcup_{j \in [K], j \neq k} \calS_j \Big)$ are linearly separable with probability at least $1 - K\delta$ w.r.t. the randomness of $\W$.
\end{corollary}
\Cref{cor:K-lin-sep} states if the layer width scales in polynomial order w.r.t. both the intrinsic dimension \emph{and} the number of subspaces, the nonlinear features are one-vs.-all separable: each individual subspace is separated from \emph{all} of the remaining subspaces. In contrast, \Cref{thm:binary-lin-sep} only depends on the intrinsic dimension, as it only considers the binary subspaces setting. 

\subsection{PROOF SKETCHES} \label{ssec:proof-sketch}
In the following, we first provide a proof sketch of \Cref{thm:binary-lin-sep} for binary subspaces $K=2$, and later we generalize the analysis to multiple subspaces $K>2$.

\smallskip

\noindent \textbf{Proof sketch for \Cref{thm:binary-lin-sep}.} We first provide a proof sketch of \Cref{thm:binary-lin-sep}, defering the full proof to \Cref{app:thm-1-proof}. Let $\X := \W \U_1 \in \reals^{D \times r}$ and $\Y := \W \U_2 \in \reals^{D \times r}$, and let $\x_n, \y_n \in \reals^r$ denote the $n^{th}$ row of $\X$ and $\Y$, respectively, written as column vectors. Note $\x_n = \U_1^\top \w_n$ and $\y_n = \U_2^\top \w_n$, where $\w_n \overset{iid}{\sim} \calN(\0_d, \I_d)$ denotes the $n^{th}$ row in $\W$. First, under \Cref{assum:network}, \eqref{eq:lin-sep-problem} holds if and only if there exists a vector $\v \in \reals^D$ such that
\begin{equation} \label{eq:lin-sep-outer-sum}
    \sum\limits_{n=1}^D v_n \x_n \x_n^\top \succ 0 \; \; \text{and} \; \; \sum\limits_{n=1}^D v_n \y_n \y_n^\top \prec 0.
\end{equation}

Next, we are interested in the \emph{existence} of a hyperplane $\v$ that separates the random features, which is not necessarily a max-margin hyperplane. We choose a linear classifier $\v$ with the following entries:

\smallskip

\begin{center}
    \textit{For all $n \in [D]$, $v_n = \mathrm{sign}\big(\|\x_n\|^2 - \|\y_n\|^2\big)$.}
\end{center}

\smallskip

\noindent This choice of $\v$ is a \emph{projection-based classifier}: the subspace onto which $\w_n$ has the largest projection determines the sign of $v_n$. If $\|\U_1^\top \w_n\|^2 > \|\U_2^\top \w_n\|^2$, then we set $v_n = +1$ to push the inner product $\v^\top f_{\W}(\U_1 \bm \alpha)$ to be ``more positive'' for any $\bm \alpha \in \reals^r$. Likewise,  setting $v_n = -1$ when $\|\U_1^\top \w_n\|^2 < \|\U_2^\top \w_n\|^2$ pushes $\v^\top f_{\W}(\U_2 \bm \alpha)$ to be ``more negative'' for any $\bm \alpha \in \reals^r$. Since $\|\U_k^\top \w_n\|^2 \sim \chi^2_r$ for $k \in \{1, 2\}$, $\|\U_1^\top \w_n\|^2 = \|\U_2^\top \w_n\|^2$ occurs with probability zero. With this choice of $\v$, \eqref{eq:lin-sep-outer-sum} is equivalent to

\begin{align} 
    &\mS_1 := \sum\limits_{i \in \calI} \x_i \x_i^\top - \sum\limits_{j \in \calI^c} \x_j \x_j^\top \succ 0, \;  \text{and} \nonumber \\
    &\mS_2 := \sum\limits_{i \in \calI} \y_i\y_i^\top - \sum\limits_{j \in \calI^c} \y_j \y_j^\top \prec 0, \label{eq:S1-S2}
\end{align}
where $\calI := \{n: v_n = +1\}$ and $\calI^c := \{n: v_n = -1\}$.

We now wish to upper bound the failure probability $P\Big( \mS_1 \not \succ 0 \cup \mS_2 \not \prec 0 \Big) = P\Big(\lambda_r\big( \mS_1 \big) \leq 0 \cup \lambda_1\big( \mS_2 \big) \geq 0 \Big)$. Next, we show $\mS_1$ and $\mS_2$ are sums of sub-exponential random matrices, 
which allows us to use Bernstein's matrix inequality \citep[Theorem 6.2]{tropp2012user} to obtain individual bounds on $P\Big(\lambda_r(\mS_1) \leq 0\Big)$ and $P\Big(\lambda_1(\mS_2) \geq 0\Big)$. Applying the union bound 
by some constant $\delta \in (0, 1)$, and then re-arranging the appropriate terms to lower bound $D$, leads to the result in \Cref{thm:binary-lin-sep}. 

\smallskip

\noindent \textbf{Extension to $K > 2$ subspaces in \Cref{cor:K-lin-sep}.} We now present a proof sketch for \Cref{cor:K-lin-sep}, omitting the full details as it directly follows from an application of \Cref{thm:binary-lin-sep}. Note we assume $(K-1)r < d/2$. This assumption is not very limiting when the number of classes is small, since $K$ and $r$ are typically much smaller than $d$ in practice.

Let $k \in [K]$ be arbitrary and $\overline{\calS}_k := \calR\Big( \begin{bmatrix}
    \U_1 & \U_2 & \dots & \U_{k-1} & \U_{k+1} & \dots & \U_K
\end{bmatrix} \Big)$ be an $\tilde{r}$-dimensional subspace, where $\tilde{r} = (K-1)r$ and $\calR(\cdot)$ denotes the column space of a matrix. Note $\overline{\calS}_k \supset \bigcup_{j = 1, j \neq k}^K \calS_j$. Also let $\tilde{\calS}_k$ denote an $\tilde{r}$-dimensional subspace $\tilde{\calS}_k \supset \calS_k$ such that $\tilde{\calS}_k$ and $\overline{\calS}_k$ satisfy \Cref{assum:subspaces}. Such a $\tilde{\calS}_k$ exists iff $(K-1)r < d/2$. Since $\calS_k \subset \tilde{\calS}_k$ and $\bigcup_{j \in [K], j \neq k} \calS_j \subset \overline{\calS}_k$, it suffices to transform $\overline{\calS}_k$ and $\tilde{\calS}_k$ into linearly separable sets. 

We directly apply \Cref{thm:binary-lin-sep} to transform $\overline{\calS}_k$ and $\tilde{\calS}_k$, and thus $\calS_k$ and $\bigcup_{j \in [K], j \neq k} \calS_j$, into linearly separable sets with high probability. Since this is now a problem of separating two $\tilde{r}$-dimensional subspaces,  the $r$ in \Cref{thm:binary-lin-sep} becomes $\tilde{r}$. Applying the union bound over all $k \in [K]$, a nonlinear layer of $\mathcal{O}\left( \mathrm{poly}(Kr) \right)$ width transforms a union of $K$ subspaces into $K$ one-vs-all linearly separable sets with high probability.

    \section{EXPERIMENTAL RESULTS} \label{sec:empirical}
In this section, we empirically verify a single random nonlinear layer makes a UoS linearly separable for both synthetic and real-world data. Specifically, in \Cref{ssec:synthetic-data}, we verify our main results \Cref{thm:binary-lin-sep} and \Cref{cor:K-lin-sep} on synthetic data, and explore settings beyond our assumptions. In \Cref{ssec:cifar10-mcr2,ssec:real-images}, we provide experimental results on real images, which again support our theoretical results. All experiments in this section were conducted on a single NVIDIA A40 GPU. Our code is available at \url{https://github.com/alecxu00/uos-linear-separability}.

\subsection{SYNTHETIC DATA} \label{ssec:synthetic-data}
In this subsection, we verify the early-layer features are linearly separable for synthetic data generated from the UoS model under various settings, including different nonlinear activations $\sigma(\cdot)$, network widths $D$, and the number of subspaces $K$. 

\smallskip

\paragraph{Synthetic data generation.} We first generated $K$ matrices $\U_1, \U_2, \dots, \U_K$ uniformly at random from the $d \times r$ Stiefel manifold. 
We then generated $N = K \cdot N_k$ training samples as follows, where $N_k = 5 \cdot 10^3$ in all settings. For all $k \in [K]$, we created $N_k$ samples via $\x_{k, i} = \U_k \z_i$, where $\z_i$ were sampled iid from $\calN(\0_r, \I_r)$ for all $i \in [N_k]$. Finally, when applicable, we generated $N$ test samples using the same procedure.

\begin{figure*}[t]
    \centering
    \begin{subfigure}{0.495\textwidth}
        \centering
        \includegraphics[width=\linewidth]{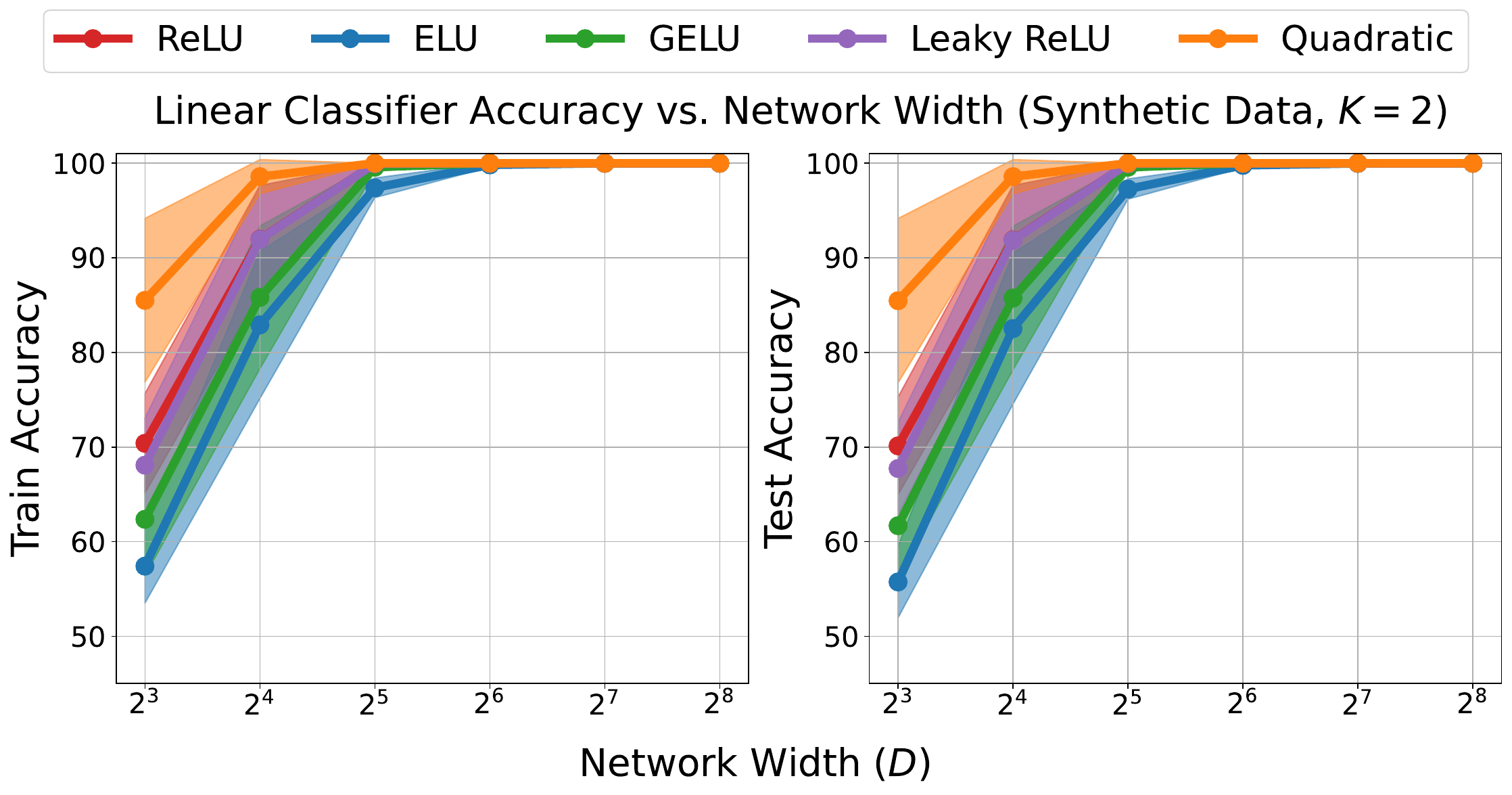}
        \caption{$K=2$ subspaces.}
        \label{subfig:train-test-acc-K2}
    \end{subfigure}\hfill
    \begin{subfigure}{0.495\textwidth}
        \centering
        \includegraphics[width=\linewidth]{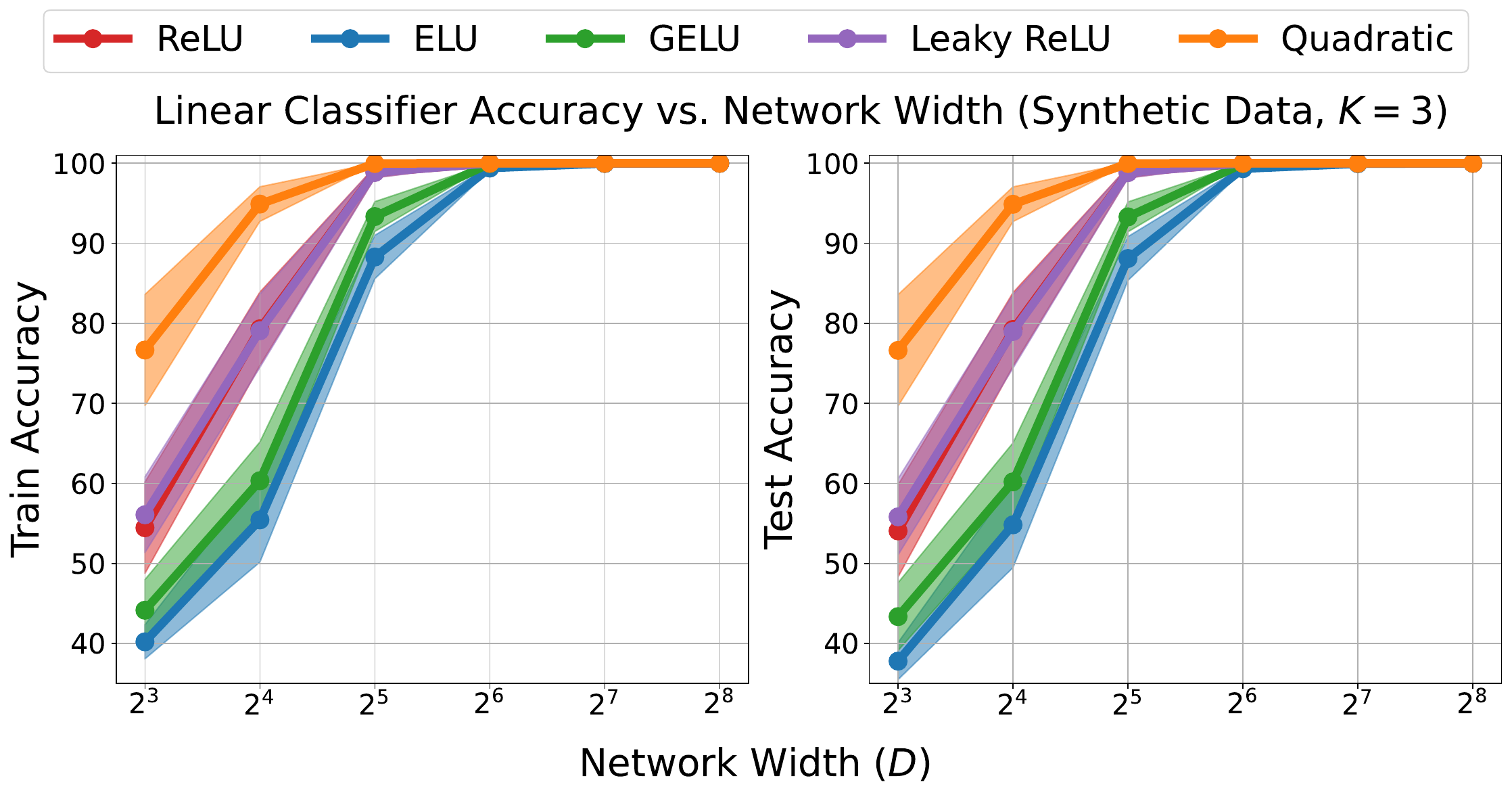}
        \caption{$K=3$ subspaces.}
        \label{subfig:train-test-acc-K3}
    \end{subfigure}
    \caption{ \textbf{Linear separability of random features on synthetic UoS data.} When the input data perfectly lie on a union of $K = 2$ (left) or $K = 3$ (right) subspaces, a linear classifier achieves perfect train and test accuracy when trained on features extracted by a sufficiently wide nonlinear layer with random weights.} 
    \label{fig:train-test-acc}
\end{figure*}

\paragraph{Model training.} 
We trained a linear classifier upon the random feature model in \Cref{eq:func-NN}. Specifically, we sampled the entries of $\W \in \reals^{D \times d}$ iid from $\calN(0, 10^{-2})$, then applied the random feature mapping \eqref{eq:func-NN}  on the train and test samples. Afterwards, we trained a linear classifier $\V \in \reals^{K \times D}$ on the train set random features under cross-entropy loss. After training, we used the trained classifier $\V$ to classify the test samples. We averaged all results over $10$ trials.

\paragraph{Results.} Based upon the above setup, we discuss the results below.

\begin{itemize}[leftmargin=*]
    \item \textbf{Dependence on $K$ and $r$.} \Cref{fig:rank-K-sweep} shows the width of ReLU and quadratic layers have similar dependence w.r.t. the intrinsic dimension and the number of subspaces. At all values of $r$ and $K$, the linear classifier achieved perfect accuracy at similar widths for both activations. Thus, although our analysis assumes a quadratic activation, our empirical findings in \Cref{fig:rank-K-sweep} imply similar results hold under the ReLU activation. Further details on the experimental setup are in \Cref{sapp:quad-relu-comp}.  
    
    \item \textbf{Effects of nonlinear activations.} \Cref{fig:train-test-acc} shows the mean and standard deviation of the train and test accuracies at each network width for every activation function. Regardless of the activation, the linear classifier's mean accuracy across the trials increased as the network width grew, eventually achieving perfect classification performance. Furthermore, the standard deviation of the accuracies approached zero at sufficiently large widths. Although linear classifiers eventually achieve perfect accuracy for all activations, different activations required different widths to do so. Specifically, the quadratic requires noticeably smaller widths to achieve linear separability compared to the other activations. 
\end{itemize}

\subsection{CIFAR-10 MCR\texorpdfstring{$^2$}{} REPRESENTATIONS} \label{ssec:cifar10-mcr2}
Second, we validate our results via experiments on the Maximal Coding Rate Reduction (MCR$^2$) representations \citep{yu2020learning} of the CIFAR-10 image dataset \citep{krizhevsky2009learning}. While natural images do not inherently adhere to a UoS model, they can be transformed into a UoS structure through nonlinear transformations. Specifically, MCR$^2$ \citep{yu2020learning} is a framework to learn data representations whose embeddings lie on a UoS \citep{wang2024a}.

\paragraph{Setup.} We trained a ResNet-18 model \citep{he2016deep} to learn MCR$^2$ representations of the CIFAR-10 dataset \citep{krizhevsky2009learning}. We adhered to the same architectural changes, hyperparameter settings, and training procedures as described in \cite{yu2020learning}. The resulting representations reside in a union of $K=10$ subspaces embedded in $\mathbb{R}^d$ with $d = 128$. From \cite{yu2020learning}, for each class $k \in [K]$, the representations of images in the $k^{\text{th}}$ class approximately lie on a 10-dimensional subspace, so $r \approx 10$. Additionally, the learned representations across different classes are nearly orthogonal, so $\theta_\ell \approx \pi/2$ for all $\ell \in [r]$.

We created training and testing sets with the MCR$^2$ representations, where each set contained $N = 10^4$ samples, with $N_k = 10^3$ samples per class. We then sampled a random weight matrix $\W \in \reals^{D \times d}$ with iid $\calN(0, 1)$ entries, and applied the random feature map $f_{\W}(\x)$ to the MCR$^2$ representations. We then trained a linear classifier $\V \in \reals^{K \times D}$ on the random features to classify the MCR$^2$ representations using cross-entropy loss. We employed the same activation functions as specified in \Cref{ssec:synthetic-data}. We varied the network width from $2^5$ to $2^{12}$ in powers of $2$, and averaged all results over $10$ trials.

\noindent \textbf{Results.} \Cref{subfig:cifar10-mcr2} illustrates the mean and standard deviation of train and test accuracies achieved on the CIFAR-10 MCR$^2$ features across different activation functions and network widths. For each activation function, the mean accuracy of the linear classifier increased with the network width, ultimately approaching \emph{near-perfect} accuracy (approximately $99\%$). The standard deviation of accuracy across trials also diminished to nearly zero as the network width increased. We hypothesize that the failure to achieve $100\%$ accuracy is due to the representations not perfectly conforming to subspaces.

\subsection{FASHION MNIST AND CIFAR-10}
\label{ssec:real-images}

\begin{figure*}[t]
    \centering
    \begin{subfigure}{0.495\textwidth}
        \centering
        \includegraphics[width=\linewidth]{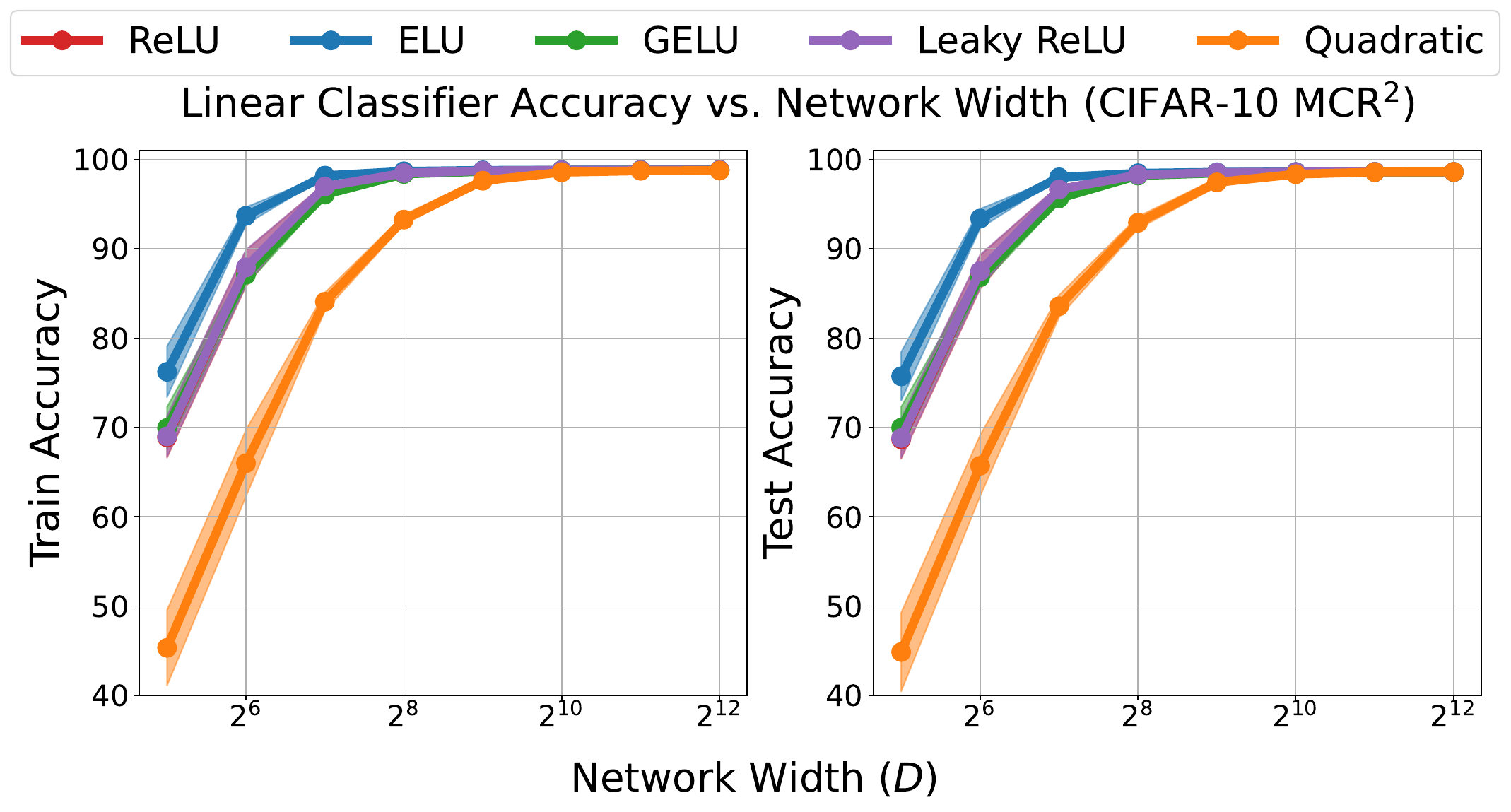}
        \caption{CIFAR-10 MCR$^2$ representations.}
        \label{subfig:cifar10-mcr2}
    \end{subfigure}\hfill
    \begin{subfigure}{0.495\textwidth}
        \centering
        \includegraphics[width=\linewidth]{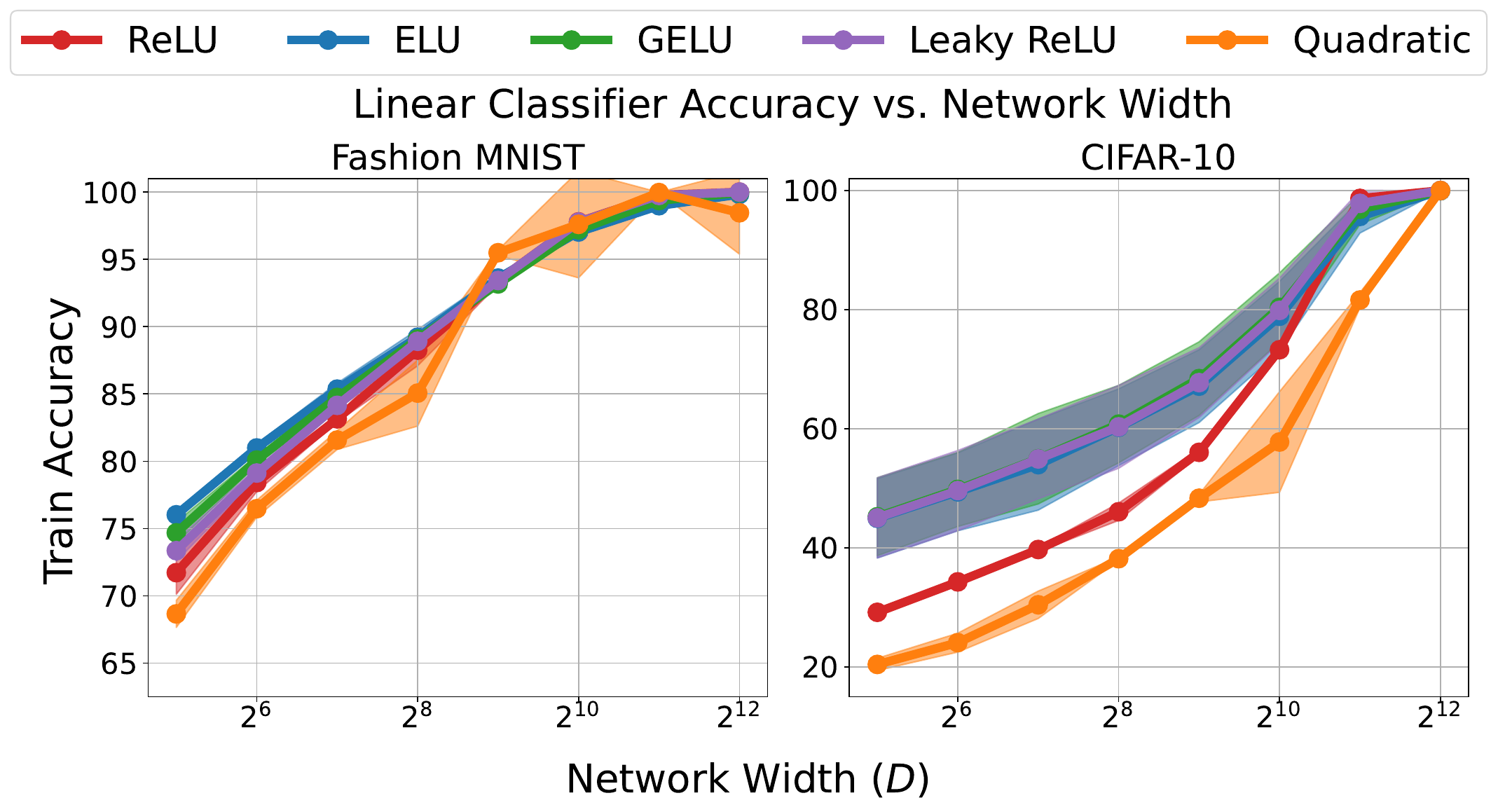}
        \caption{Fashion MNIST (left) and CIFAR-10 (right) images}
        \label{subfig:fashion-mnist-cifar10}
    \end{subfigure}
    \caption{ \textbf{Linear separability of random features on image data.} \textit{Left:} after transforming image data to lie on a UoS using the MCR$^2$ \citep{yu2020learning} framework, sufficiently many random features are linearly separable. \textit{Right:} sufficiently many random features using \emph{raw} image data as input are also linearly separable.} 
    \label{fig:real-images}
\end{figure*}

Finally, we show that our theory approximately holds on the Fashion MNIST \citep{xiao2017fashion} and CIFAR-10 \citep{krizhevsky2009learning} datasets. Both datasets contain $K = 10$ classes. \Cref{fig:image_data_class_svals} in \Cref{app:uos-image-data} shows both datasets approximately satisfy the UoS data model: in both datasets, a small number of singular values account for a large majority each class data matrix's Frobenius norm. In particular, both datasets' per-class intrinsic subspace dimensions are about on the order of $10^1$, even though their ambient dimensions on the order of $10^2$ (Fashion MNIST) or $10^3$ (CIFAR-10). See \Cref{app:uos-image-data} for more details.

\paragraph{Setup.} For both datasets, we randomly sampled $N = 10^4$ training images, with $N_k = 10^3$ images per class. Then, we flattened the images into $d$-dimensional vectors, where $d = 784$ for Fashion MNIST, and $d = 3072$ for CIFAR-10. Finally, we followed the exact training procedure as described in \Cref{ssec:cifar10-mcr2}, but averaged all results over $5$ trials instead. 

\paragraph{Results.} \Cref{subfig:fashion-mnist-cifar10} shows the mean and standard deviation of the train accuracies achieved on the Fashion MNIST and CIFAR-10 random features across various activation functions. Recall from \Cref{app:uos-image-data}, the per-class subspace dimensions are on the order of $10^1$ in both datasets. \Cref{subfig:fashion-mnist-cifar10} shows the linear classifier achieves near-perfect accuracy at widths around $2^{11}$ or $2^{12}$ across all activations, which is on the order of $10^3$. Thus, network widths that are polynomial in the \emph{intrinsic} data dimension suffices for linear separability, which aligns with our theoretical results. 

Furthermore, recall that CIFAR-10's ambient dimension is about $4 \times$ that of Fashion MNIST ($d = 3072$ for CIFAR-10, while $d = 784$ for Fashion MNIST). Despite this difference, between $2^{11}$ and $2^{12}$ random features suffice for linear separability in \emph{both} datasets, implying little to no dependence on the data ambient dimension.

    \section{CONCLUSION} \label{sec:conclusion}

In this work, we studied the linear separability of early-layer features in nonlinear networks for low-dimensional data, using a UoS model motivated by the low intrinsic dimensionality of image data. We rigorously proved that a single nonlinear layer with random weights and quadratic activation can transform $K \geq 2$ subspaces into (one-vs.-all) linearly separable sets with high probability. Notably, our result requires the network width to be polynomial in the  intrinsic dimension, while previous results require exponential dependence. Although our analysis assumes a quadratic activation, our empirical findings on synthetic and real data indicate similar results hold for other activations, such as ReLU.

\section*{ACKNOWLEDGEMENTS}
AX, CY, PW, and QQ acknowledge support from NSF CAREER CCF-2143904, NSF IIS 2312842,
NSF IIS 2402950, ONR N00014-22-1-2529, and a gift grant from KLA. The authors would also like
to thank Dr. Samet Oymak (University of Michigan) and Dr. Boris Hanin (Princeton University)
for their valuable insights.

\clearpage

    \bibliography{refs}

    \clearpage

    \section*{Checklist}
\begin{enumerate}

  \item For all models and algorithms presented, check if you include:
  \begin{enumerate}
    \item A clear description of the mathematical setting, assumptions, algorithm, and/or model. [\textcolor{blue}{Yes}/No/Not Applicable]
    \item An analysis of the properties and complexity (time, space, sample size) of any algorithm. [Yes/No/\textcolor{blue}{Not Applicable}]
    \item (Optional) Anonymized source code, with specification of all dependencies, including external libraries. [\textcolor{blue}{Yes}/No/Not Applicable]
  \end{enumerate}

  \item For any theoretical claim, check if you include:
  \begin{enumerate}
    \item Statements of the full set of assumptions of all theoretical results. [\textcolor{blue}{Yes}/No/Not Applicable]
    \item Complete proofs of all theoretical results. [\textcolor{blue}{Yes}/No/Not Applicable]
    \item Clear explanations of any assumptions. [\textcolor{blue}{Yes}/No/Not Applicable]     
  \end{enumerate}

  \item For all figures and tables that present empirical results, check if you include:
  \begin{enumerate}
    \item The code, data, and instructions needed to reproduce the main experimental results (either in the supplemental material or as a URL). [\textcolor{blue}{Yes}/No/Not Applicable]
    \item All the training details (e.g., data splits, hyperparameters, how they were chosen). [\textcolor{blue}{Yes}/No/Not Applicable]
    \item A clear definition of the specific measure or statistics and error bars (e.g., with respect to the random seed after running experiments multiple times). [\textcolor{blue}{Yes}/No/Not Applicable]
    \item A description of the computing infrastructure used. (e.g., type of GPUs, internal cluster, or cloud provider). [\textcolor{blue}{Yes}/No/Not Applicable]
  \end{enumerate}

  \item If you are using existing assets (e.g., code, data, models) or curating/releasing new assets, check if you include:
  \begin{enumerate}
    \item Citations of the creator If your work uses existing assets. [\textcolor{blue}{Yes}/No/Not Applicable]
    \item The license information of the assets, if applicable. [Yes/No/\textcolor{blue}{Not Applicable}]
    \item New assets either in the supplemental material or as a URL, if applicable. [Yes/No/\textcolor{blue}{Not Applicable}]
    \item Information about consent from data providers/curators. [Yes/No/\textcolor{blue}{Not Applicable}]
    \item Discussion of sensible content if applicable, e.g., personally identifiable information or offensive content. [Yes/No/\textcolor{blue}{Not Applicable}]
  \end{enumerate}

  \item If you used crowdsourcing or conducted research with human subjects, check if you include:
  \begin{enumerate}
    \item The full text of instructions given to participants and screenshots. [Yes/No/\textcolor{blue}{Not Applicable}]
    \item Descriptions of potential participant risks, with links to Institutional Review Board (IRB) approvals if applicable. [Yes/No/\textcolor{blue}{Not Applicable}]
    \item The estimated hourly wage paid to participants and the total amount spent on participant compensation. [Yes/No/\textcolor{blue}{Not Applicable}]
  \end{enumerate}

\end{enumerate}

    \clearpage 


%
%






%
\runningtitle{Understanding How Nonlinear Networks Create Linearly-Separable Features for Low-Dimensional Data}

%

\onecolumn
\appendix

\aistatstitle{SUPPLEMENTARY MATERIALS}

\paragraph{Notation.} We re-state previous notation here for convenience, and introduce some new notation. We use $\calN(\mu, \sigma^2)$ to denote a Gaussian distribution with mean $\mu$ and variance $\sigma^2$, $\calN(\bm{\mu}, \bm{\Sigma})$ to denote a multivariate Gaussian distribution with mean $\bm{\mu}$ and covariance $\bm{\Sigma}$, and $\chi^2_m$ to denote a chi-squared distribution with $m$ degrees of freedom. We use $Z \: \big| \: \calA$ to denote random variable $Z$ conditioned on an event $\calA$. We denote the pdf of a random variable $Z$ with $f_Z(\cdot)$, and the covariance of a random vector with $\covrm(\cdot)$. 

We use $\| \cdot \|$ to denote the Euclidean norm of a vector, $\sigma_i(\cdot)$ to denote the $i^{th}$ largest singular value of a matrix, and $\lambda_i(\cdot)$ to denote the $i^{th}$ largest eigenvalue of a symmetric matrix. We also use $\0_m$ to denote the $m$-dimensional vector of all zeroes.

For any positive integer $N$, we use $[N]$ to denote the set $\{1, 2, \dots, N\}$. With a slight abuse of notation, for some function $\phi$ and set $\calX$, $\phi\big( \calX \big)$ denotes the set $\big\{ \x \in \calX: \phi \big( \x \big) \big\}$. 

\section{RELATED WORKS} \label{app:related}
We provide a more detailed discussion of the relationship between our results and prior work.

\smallskip
\noindent \textbf{Separation capacity of nonlinear networks.} As discussed in \Cref{ssec:contributions}, \cite{dirksen2022separation, ghosal2022randomly} are most closely related to ours. They analyzed two arbitrarily-structured, nonlinearly-separated classes, and showed the resulting features from two-layer \citep{dirksen2022separation} and one-layer \citep{ghosal2022randomly} random ReLU networks are linearly separable with high probability. In our work, we specifically model the inputs as lying on a union of low-dimensional subspaces, and consider the quadratic activation instead of ReLU. Under our data model, the results in \cite{dirksen2022separation, ghosal2022randomly} require the network widths to scale \emph{exponentially} with the intrinsic dimension of the input data. In contrast, our result requires \emph{polynomial} dependence. Another related work, \cite{an2015can}, also assumes the data is from two arbitrary nonlinearly-separated sets. They prove there exists a \emph{deterministic} two-layer ReLU network that makes these sets linearly separable.

\smallskip

\noindent \textbf{XOR data.} Previous works on neural network analyses have considered XOR input data \citep{glasgow2024sgd,meng2024benign}, which is a special case of our UoS data model. To visualize this, consider an arbitrary data point $\bm x \in \left\{-1, +1\right\}^2 = \begin{bmatrix}
    x_1 & x_2
\end{bmatrix}$, where $x_i \in \{-1, +1\}$. Then, the corresponding label is $y = \operatorname{XOR}(x_1, x_2) = \begin{cases}
    +1 & x_1 = x_2 \\
    -1 & x_1 \neq x_2
\end{cases}$. \Cref{fig:xor-data} shows these data points lie on a union of two one-dimensional linear subspaces, where each linear subspace represents a different class. 
\begin{wrapfigure}{r}{0.4\textwidth}
    \begin{tikzpicture}[scale=0.8]

  \draw[<->, thick] (-2.2,0) -- (2.4,0);
  \draw[<->, thick] (0,-2.2) -- (0,2.4);

  \draw[dashed, blue!70, line width=1.2pt]
    (-1.8,-1.8) -- (1.8,1.8)
    node[above right, blue!80] {$\mathcal{S}_1$};

  \draw[dashed, orange!70, line width=1.2pt]
    (-1.8,1.8) node[above left, orange!70] {$\mathcal{S}_2$} -- (1.8,-1.8);

  \fill[blue!80]  ( 1.3, 1.3) circle (4pt)
    node[right, blue!90, font=\small] {$\quad y=+1$};
  \fill[blue!80]  (-1.3,-1.3) circle (4pt)
    node[left,  blue!90, font=\small] {$y=+1 \quad$};

  \fill[orange!80]  ( 1.3, -1.3) circle (4pt)
    node[right, orange!80, font=\small] {$\quad y=-1$};
  \fill[orange!80]  (-1.3,1.3) circle (4pt)
    node[left,  orange!80, font=\small] {$y=-1 \quad$};
    
    \end{tikzpicture}
    \caption{Two-dimensional XOR data lies on a union of one-dimensional subspaces.}
    \label{fig:xor-data}
\end{wrapfigure}

\smallskip 

\noindent \textbf{Neural collapse in shallow nonlinear networks.} Recently, \cite{hong2024beyond} studied the Neural Collapse (NC) phenomenon in shallow ReLU networks. Specifically, they identified sufficient data-dependent conditions on when shallow ReLU networks exhibit NC. Although our work and \cite{hong2024beyond} study the properties of the features in nonlinear networks, the settings have fundamental differences. Notably, NC characterizes the structure of the features from the \emph{penultimate} layer. Additionally, \cite{hong2024beyond} consider shallow ReLU networks to analyze NC in more realistic settings compared to previous works. In contrast, we study the linear separability of the features in the \emph{early} layers in DNNs, and study a shallow nonlinear network to facilitate such analysis. 

\smallskip

\noindent \textbf{Learning with random features.} Our theoretical result uses random weights in the nonlinear layer, yielding a random feature map. Learning with random features was introduced in \citep{rahimi2007random} as an alternative to kernel methods, and its generalization properties have been widely studied \citep{rahimi2008weighted, rudi2017generalization, bach2017equivalence, li2021towards, chen2024conditioning}. Although our result does not directly imply broader conclusions about learning with random features, we show that a random feature map can transform subspaces into linearly separable sets with high probability. This implies if train and test samples lie on the same subspaces, a linear classifier can perfectly classify the test samples. 

\smallskip

\noindent \textbf{Neural tangent kernel.} As  discussed in \Cref{ssec:theorem}, previous works have shown the global convergence of gradient descent on overparameterized two-layer networks using NTK techniques \citep{du2019gradient,huang2020dynamics}. These works showed that during training, a highly overparameterized two-layer network remains close to its random initialization, which is closely related to our random feature model. These results showed gradient descent converges to zero training loss for wide two-layer networks. In contrast, our work does not consider any training. Rather, we directly assume a random feature model, showing a nonlinear layer with random weights makes two linear subspaces linearly separable with high probability. 

\smallskip 

\noindent \textbf{Analysis of quadratic-activation networks.} 
Our theoretical result assumes the nonlinear activation is the entry-wise quadratic function. While previous works on quadratic activation have focused on the optimization landscape and generalization abilities of overparameterized networks \citep{li2018algorithmic, soltanolkotabi2018theoretical, du2018power,  sarao2020optimization, gamarnik2024stationary}, our contribution lies in demonstrating that data on a union of subspaces can be made linearly separable with high probability under this quadratic activation. This perspective provides new insights into the feature separation properties of quadratic activation networks, complementing the optimization-centric findings of prior studies. 


\smallskip

\noindent \textbf{Rare Eclipse problem.} Finally, our problem shares conceptual similarities with the Rare Eclipse problem studied in \cite{bandeira2017compressive, cambareri2017rare}, which focuses on mapping two linearly separable sets into a lower-dimensional space where they become disjoint with high probability. Using Gordon's Escape through a Mesh \citep{gordon1988milman}, \cite{bandeira2017compressive} demonstrated that a random Gaussian matrix achieves this and provides a lower bound on the required dimension. Similarly, we show that a nonlinear random mapping can \emph{transform} two sets (linear subspaces) into linearly separable sets with high probability. However, beyond this shared goal of increasing separability, the two problems differ fundamentally in approach and context.

\section{UNION OF SUBSPACES IN REAL IMAGE DATASETS}
\label{app:uos-image-data}
In this section, we provide empirical evidence showing common image datasets approximately lie on a union of low-dimensional subspaces, where each subspace represents a different class in the dataset. 

\paragraph{Setup.} We considered the MNIST, Fashion MNIST \citep{xiao2017fashion}, and CIFAR-10 \citep{krizhevsky2009learning} datasets. In each dataset, we first flattened the images so that they are $d$-dimensional vectors, where $d$ is the number channels multiplied by the pixels. For MNIST and Fashion MNIST, $d = 1 \times 28 \times 28  = 784$, while for CIFAR-10, $d = 3 \times 32 \times 32 = 3072$. Then, for each class, we create a $d \times N$ data matrix, where $N = 5000$ is the number of data points in each class. Finally, for each class data matrix, we compute the number of singular values that account for $95\%$ and $99\%$ of the Frobenius norm in each class's data matrix. 

\paragraph{Results.} \Cref{fig:image_data_class_svals} shows the proportion of singular values needed to capture $95\%$ (darker bars) and $99\%$ (lighter bars) of each class data matrix's Frobenius norms. For MNIST, Fashion MNIST, and CIFAR-10 respectively, about the first $15 - 20\%$ (about $115 - 155$ out of $784$), $5 - 10\%$ (about $40 - 80$ out of $784$), and $2 - 4\%$ (about $60 - 120$ out of $3072$) of the singular values account for $99\%$ of most class data matrix's Frobenius norms --- the lone outlier is class $5$ in Fashion MNIST. This implies each class \emph{approximately} lies on its own low-dimensional subspace. In other words, \textbf{these datasets approximately satisfy the UoS data model.}

\begin{figure}
    \centering
    \includegraphics[width=\linewidth]{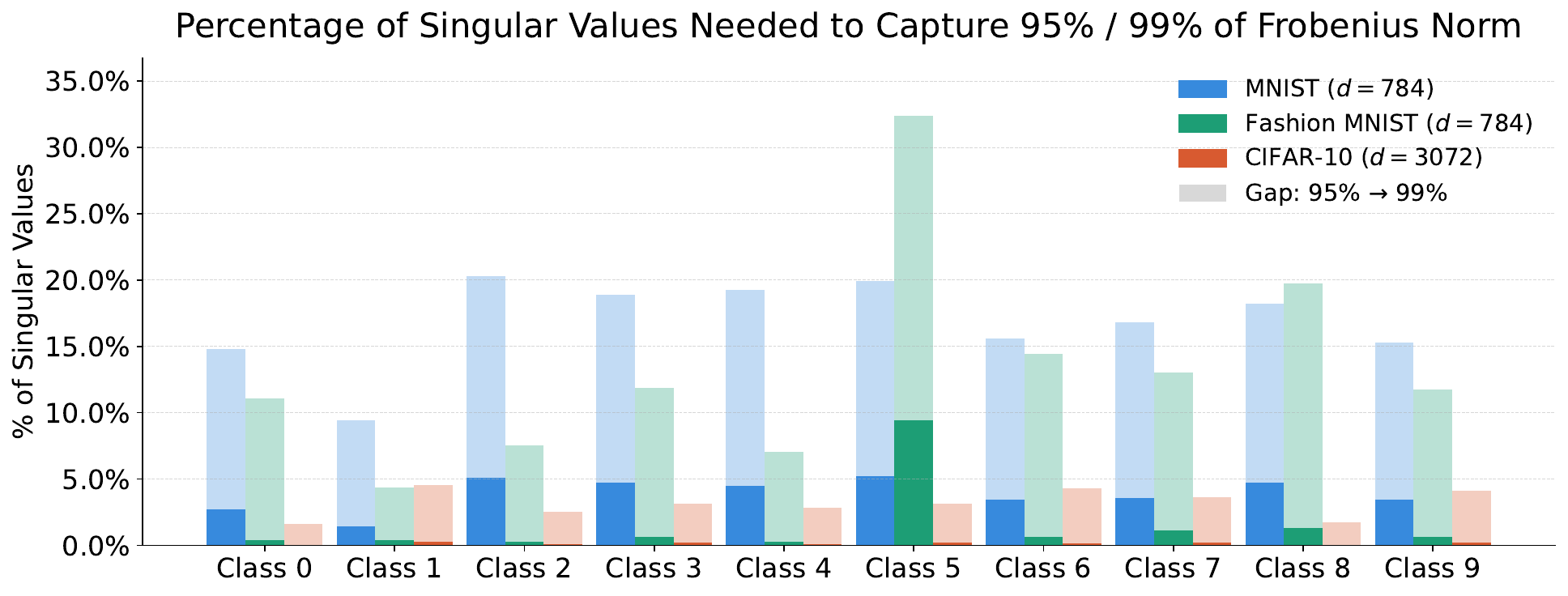}
    \caption{\textbf{In common image datasets, a small proportion of singular values in each class's data matrix capture $95\%$ / $99\%$ of the Frobenius norm.} The darker bars indicate the proportion of singular values needed to reach $95\%$ of that class data matrix's Frobenius norm, while the lighter bars indicate the proportion needed to reach $99\%$.}
    \label{fig:image_data_class_svals}
\end{figure}
\section{LINEAR SEPARABILITY OF A UNION OF SUBSPACES}\label{app:problem} In this section, we discuss why a nonlinear  layer $f_{\bm W}(\bm x) = \sigma(\bm W \bm x)$ is necessary to transform a UoS into linearly separable sets. We first show two subspaces themselves are not linearly separable. Next, we show applying either a linear transformation or a nonlinear activation \emph{individually} are insufficient in making a UoS linearly separable. 

\noindent \textbf{Subspaces are not linearly separable in general.} Suppose we have two one-dimensional subspaces $\calS_1, \calS_2 \subset \reals^2$ with bases $\u_1 = \begin{bmatrix}
        1 & 1
\end{bmatrix}^\top$ and $\u_2 = \begin{bmatrix}
    -1 & 1
\end{bmatrix}^\top$, respectively. As shown in \Cref{fig:activations-only}, there does not exist any hyperplane (line) that can linearly separate $\calS_1$ and $\calS_2$ because they both pass through the origin.

\smallskip

\noindent \textbf{Linear mapping alone is insufficient for linear separability.} Now suppose $\calS_1$ and $\calS_2$ are arbitrary $r_1$ and $r_2$-dimensional subspaces of $\reals^d$, and define $g_{\W}(\x) := \W \x$, where $\x \in \reals^d$ and $\W \in \reals^{D \times d}$. We show $g_{\W}(\calS_1)$ and $g_{\W}(\calS_2)$ are not linearly separable sets. For any $k \in \{1, 2\}$ and arbitrary $\bm \alpha^{(k)} \in \reals^{r_k}$, we have
\begin{equation*}
    \W \U_k \bm \alpha^{(k)} = \tilde{\U}_k \bm \alpha^{(k)},
\end{equation*}
where $\tilde{\U}_k \in \reals^{D \times r_k}$. Therefore, the point $\W\U_k \bm \alpha^{(k)}$ lies in an $r_k$-dimensional subspace in $\reals^D$. Since this holds for all $\bm \alpha^{(k)} \in \reals^{r_k}$, the sets $g_{\W}(\calS_1)$ and $g_{\W}(\calS_2)$ remain as linear subspaces of $\reals^D$ that pass through the origin, which are not linearly separable in general. 

\smallskip

\noindent \textbf{Nonlinear activations alone are insufficient for linear separability.} Second, for various activation functions, we show $\sigma(\calS_1)$ and $\sigma(\calS_2)$ are not linearly separable sets through counterexamples.  Again suppose the bases of $\calS_1, \calS_2 \subset \reals^2$ are $\u_1 = \begin{bmatrix}
    1 & 1
\end{bmatrix}^\top$ and $\u_2 = \begin{bmatrix}
    -1 & 1
\end{bmatrix}^\top$, respectively. Let us first consider the entry-wise quadratic activation, which we considered in our theoretical analysis. For any $\alpha \in \reals$, we have
    \begin{align*}
        &\sigma(\u_1 \alpha) = 
        \begin{bmatrix}
            1^2 \\
            1^2
        \end{bmatrix} \alpha^2 = 
        \begin{bmatrix}
            \alpha^2 \\
            \alpha^2
        \end{bmatrix} \; \text{and} \;
        \sigma(\u_2 \alpha) = 
        \begin{bmatrix}
            (-1)^2 \\
            1^2
        \end{bmatrix} \alpha^2 = 
        \begin{bmatrix}
            \alpha^2 \\
            \alpha^2
        \end{bmatrix},
    \end{align*}
    so $\sigma(\u_1 \alpha) = \sigma(\u_2 \alpha)$. 
    Therefore, the sets $\sigma(\calS_1)$ and $\sigma(\calS_2)$ are \emph{identical} (see \Cref{fig:activations-only}, left), clearly implying they are not distinguishable.
    
    \smallskip
    
    Next, consider $\sigma(\cdot) = \text{ReLU}(\cdot)$, which is more commonly used in practice. For any nonzero $\alpha \in \reals$, $\sigma(\u_1 \alpha) = \u_1 \alpha$ if $\alpha > 0$, and $\sigma(\u_1 \alpha) = \0_2$ if $\alpha < 0$. Additionally, $\sigma(\u_2 \alpha) = \begin{bmatrix}
        0 & \alpha
    \end{bmatrix}^\top$ if $\alpha > 0$, and $\sigma(\u_2 \alpha) = \begin{bmatrix}
        -\alpha & 0
    \end{bmatrix}^\top$ if $\alpha < 0$. 
    Therefore, $\sigma(\calS_1) = \big\{\x \in \reals^2: x_1 > 0, x_2 > 0\} \cup \{\0_2\}$, while $\sigma(\calS_2) = \big\{ \x \in \reals^2: x_1 > 0, x_2 = 0 \big\} \cup \big\{ \x \in \reals^2: x_1 = 0, x_2 > 0 \big\}$, where $x_1$ and $x_2$ respectively denote the first and second elements of $\x \in \reals^2$. These sets are \emph{not} linearly separable (see \Cref{fig:activations-only}, right), since some points in $\sigma(\mathcal{S}_2)$ are above $\sigma(\mathcal{S}_1)$, and some points are below.

\begin{figure}[t]
    \centering
    \begin{minipage}{0.5\textwidth}
        \centering
        \begin{subfigure}{\textwidth}
        \centering
        \begin{tikzpicture}[scale=0.5]
            \draw[<->, thick] (-2, 0) -- (2, 0); 
            \draw[<->, thick] (0, -2) -- (0, 2); 
            
            \draw[<->, thick, blue] (-1.5, -1.5) -- (1.5, 1.5) node[above right] {$\calS_1$}; 
            \draw[<->, thick, orange] (1.5, -1.5) -- (-1.5, 1.5) node[above left] {$\calS_2$}; 
            
            \draw[->, thick, black] (3.0, 0) -- (4.5, 0) node[midway, above] {Quadratic};
            
            \draw[<->, thick] (5.5, 0) -- (9.5, 0); 
            \draw[<->, thick] (7.5, -2) -- (7.5, 2); 
            
            \draw[->, thick, blue] (7.5, 0) -- (9, 1.5) node[above right] {$\sigma(\calS_1)$}; 
            \draw[->, thick, orange, dashed] (7.5, 0) -- (9, 1.5) node[above left] {$\sigma(\calS_2)$}; 
        \end{tikzpicture}
        \caption{Quadratic applied to $\mathrm{span}(\u_1) \cup \mathrm{span}(\u_2)$.}
        \end{subfigure}
    \end{minipage}%
    \hfill
    \begin{minipage}{0.5\textwidth}
        \centering
        \begin{subfigure}{\textwidth}
            \centering
            \begin{tikzpicture}[scale=0.5]
            \draw[<->, thick] (-2, 0) -- (2, 0); 
            \draw[<->, thick] (0, -2) -- (0, 2); 
            
            \draw[<->, thick, blue] (-1.5, -1.5) -- (1.5, 1.5) node[above right] {$\calS_1$}; 
            \draw[<->, thick, orange] (1.5, -1.5) -- (-1.5, 1.5) node[above left] {$\calS_2$}; 
            
            \draw[->, thick, black] (3.0, 0) -- (4.5, 0) node[midway, above] {ReLU};
            
            \draw[<->, thick] (5.5, 0) -- (9.5, 0); 
            \draw[<->, thick] (7.5, -2) -- (7.5, 2); 
            
            \draw[->, thick, blue] (7.5, 0) -- (9, 1.5) node[above right] {$\sigma(\calS_1)$}; 
            \draw[->, thick, orange, dashed] (7.5, 0) -- (7.5, 1.5) node[above left] {$\sigma(\calS_2)$}; 
            \draw[->, thick, orange, dashed] (7.5, 0) -- (9.0, 0) node[below] {$\sigma(\calS_2)$}; 
        \end{tikzpicture}
        \caption{ReLU applied to $\mathrm{span}(\u_1) \cup \mathrm{span}(\u_2)$.}
        \end{subfigure}
    \end{minipage}
    \caption{\textbf{Activation alone is insufficient for linearly separating two subspaces.} When $\calS_1 = \mathrm{span}(\u_1)$ and $\calS_2 = \mathrm{span}(\u_2)$, the sets $\sigma(\calS_1)$ and $\sigma(\calS_2)$ are not linearly separable for $\sigma(\cdot) = $ quadratic (left) and $\sigma(\cdot) = \mathrm{ReLU}(\cdot)$ (right).} 
    \label{fig:activations-only}
\end{figure}
\section{AUXILIARY RESULTS} \label{app:supporting}
We provide auxiliary lemmas that are useful in proving Theorem~\ref{thm:binary-lin-sep}. Let $\w \sim \calN(\0_d, \I_d)$, $\U_1, \U_2 \in \reals^{d \times r}$ respectively be orthonormal bases for subspaces $\calS_1$ and $\calS_2$ that satisfy \Cref{assum:subspaces}. Also let $\x := \U_1^\top \w$, and $\y := \U_2^\top \w$. Note $\x \sim \calN(\0_r, \I_r)$ and $\y \sim \calN(\0_r, \I_r)$ are \emph{correlated}. Finally, let $\ab$ and $\blb$ be random vectors with the following distributions:
\begin{align*}
    \ab \sim \x \: \big| \: \|\x\|^2 > \|\y\|^2 \; \; \text{and} \; \; \blb \sim \y \: \big| \: \|\x\|^2 > \|\y\|^2.
\end{align*}

\subsection{EXPECTATION OF ORDER STATISTICS: \texorpdfstring{$\chi^2_m$}{ } RANDOM VARIABLES}
\Cref{lem:expec-max-min-chi-squares} provides exact expressions for the expectation of the maximum and minimum of two iid $\chi^2_m$ random variables.
\begin{lemma} \label{lem:expec-max-min-chi-squares}
Let $X, Y \overset{\text{iid}}{\sim} \chi^2_m$, $A = \maxrm\{X, Y\}$, and $B = \minrm\{X, Y\}$. Then,
    \begin{align*}
        &\eE[A] = m + \frac{2}{\sqrt{\pi}} \frac{\Gamma((m+1)/2)}{\Gamma(m/2)} \; \; \text{and} \; \; \eE[B] = m - \frac{2}{\sqrt{\pi}} \frac{\Gamma((m+1)/2)}{\Gamma(m/2)},
    \end{align*}
    where $\Gamma(\cdot)$ denotes the Gamma function.
\end{lemma}

\begin{proof}
Note $A + B = X + Y$, so $\eE[A + B] = \eE[X + Y] = 2m$. Therefore, it suffices to derive $\eE[A]$:
\begin{align*}
    \eE[A] &= \int\limits_0^\infty \int\limits_0^\infty \maxrm\{x, y\} f_X(x) f_Y(y) \: dx \: dy \\
    &= \int\limits_0^\infty \int\limits_y^\infty x f_X(x) f_Y(y) \: dx \: dy + \int\limits_0^\infty \int\limits_x^\infty y f_X(x) f_Y(y) \: dy \: dx = 2 \int\limits_0^\infty \int\limits_y^\infty x f_X(x) f_Y(y) \: dx \: dy \\
    &\overset{(a)}{=} \frac{2}{2^m \Gamma(m/2)^2} \int\limits_0^\infty \int\limits_y^\infty x^{m/2}e^{-x/2} y^{m/2 - 1} e^{-y/2} \: dx \: dy, 
\end{align*}
where we substituted the pdf of a $\chi^2_m$ distribution in $(a)$. Letting $t = \frac{x}{2}$ results in
\begin{align*}
    &\frac{2}{2^m \Gamma(m/2)^2} \int\limits_0^\infty \int\limits_y^\infty x^{m/2}e^{-x/2} y^{m/2 - 1} e^{-y/2} \: dx \: dy \\
    &= \frac{4}{2^{m/2} \Gamma(m/2)^2} \int\limits_0^\infty \int\limits_{y/2}^\infty t^{m/2} e^{-t} y^{m/2 - 1} e^{-y/2} \: dt \: dy \\
    &\overset{(b)}{=} \frac{4}{2^{m/2} \Gamma(m/2)^2} \int\limits_0^\infty \Gamma(m/2 + 1, y/2) y^{m/2 - 1}e^{-y/2} \: dy, 
\end{align*}
where in $(b)$, we substituted the definition of the upper incomplete Gamma function, denoted as $\Gamma(p, x)$. Using the recurrence relation $\Gamma(p + 1, x) = p\Gamma(p, x) + x^p e^{-x}$ yields
\begin{align*}
    &\frac{4}{2^{m/2} \Gamma(m/2)^2} \int\limits_0^\infty \Gamma(m/2 + 1, y/2) y^{m/2 - 1}e^{-y/2} \: dy \\
    &= \underbrace{\frac{m}{\Gamma(m/2)^2} \int\limits_0^\infty \Gamma(m/2, y/2) (y/2)^{m/2 - 1} e^{-y/2} \: dy}_{\text{$(c)$}} + \underbrace{\frac{1}{2^{m-2} \Gamma(m/2)^2} \int\limits_0^\infty y^{m-1} e^{-y} \: dy}_{\text{$(d)$}}. 
\end{align*}
We first simplify $(c)$. Letting $s = y / 2$, $(c)$ becomes
\begin{equation*}
    \frac{2m}{\Gamma(m/2)^2}\int\limits_0^\infty \Gamma(m/2, s) s^{m/2 - 1} e^{-s} \: ds.
\end{equation*}
From \cite[Pg. 137, Eq. (8)]{bateman1953higher}:
\begin{equation*}
    \int\limits_0^\infty \Gamma(m/2, s) s^{m/2 - 1} e^{-s} \: ds = \frac{\Gamma(m)}{(m/2) \cdot 2^m} {}_2F_1(1, m; m/2 + 1; 1/2)
\end{equation*}
where ${}_2F_1(a, b; c, d)$ denotes the ordinary hypergeometric function. By Gauss's Second Summation Theorem \citep{slater1966generalized}:
\begin{equation} \label{eq:gauss-second-sum}
    \frac{\Gamma(m)}{(m/2) \cdot 2^m} {}_2F_1(1, m; m/2 + 1; 1/2) = \frac{\Gamma(m) \Gamma(1/2) \Gamma(m/2 + 1)}{(m/2) \cdot 2^m \cdot \Gamma((m+1)/2)}.
\end{equation}
By Legendre's duplication formula, $\Gamma(m) = \frac{\Gamma(m/2) \Gamma((m + 1)/2)}{2^{1-m} \sqrt{\pi}}$. Additionally, the Gamma function satisfies the recurrence relation $\Gamma(z + 1) = z\Gamma(z)$ for all $z > 0$. Substituting these expressions into \eqref{eq:gauss-second-sum} leads to
\begin{equation*}
    \frac{\Gamma(m) \Gamma(1/2) \Gamma(m/2 + 1)}{(m/2) \cdot 2^m \cdot \Gamma((m+1)/2)} 
    = \frac{\Gamma(m/2)\Gamma(m/2 + 1)}{m} = \frac{\Gamma(m/2)^2}{2}.
\end{equation*}
Therefore, $(c)$ fully simplifies to the following:
\begin{equation*} 
    \frac{m}{\Gamma(m/2)^2} \int\limits_0^\infty \Gamma(m/2, y/2) (y/2)^{m/2 - 1} e^{-y/2} \: dy = \frac{2m}{\Gamma(m/2)^2}\frac{\Gamma(m/2)^2}{2} = m.
\end{equation*}
We now simplify $(d)$:
\begin{equation*}
    \frac{1}{2^{m-2} \Gamma(m/2)^2} \int\limits_0^\infty y^{m-1} e^{-y} \: dy \overset{(e)}{=} \frac{\Gamma(m)}{2^{m-2} \Gamma(m/2)^2} \overset{(f)}{=} \frac{2\Gamma((m+1)/2)}{\Gamma(m/2)\sqrt{\pi}},
\end{equation*}
where $(e)$ is by the definition of the Gamma function, and $(f)$ is by Legendre's duplication formula. Thus,
\begin{equation*}
    \eE[A] = m + \frac{2}{\sqrt{\pi}}\frac{\Gamma((m+1)/2)}{\Gamma(m/2)}.
\end{equation*}
We then use the property $\eE[A + B] = \eE[A] + \eE[B] = 2m$ to obtain $\eE[B]$:
\begin{equation*}
    \eE[B] = m - \frac{2}{\sqrt{\pi}}\frac{\Gamma((m+1)/2)}{\Gamma(m/2)}.
\end{equation*}
\end{proof}

\subsection{EIGENVALUES OF DIFFERENCE OF PROJECTION MATRICES}
Next, we provide a result about the eigenvalues of $\U_1\U_1^\top - \U_2\U_2^\top$.
\begin{lemma} \label{lem:proj-mat-diff-eigvals}
    Let $\U_1, \U_2 \in \reals^{d \times r}$ s.t. $\U_1^\top \U_1 = \U_2^\top \U_2 = \I_r$, and $\sigma_\ell(\U_1^\top \U_2) = \cos(\theta_\ell)$ for all $\ell \in [r]$, where $\theta_1 := \theta_{min} > 0$. Then, $\U_1\U_1^\top - \U_2\U_2^\top$ has $r$ eigenvalues equal to $\sin(\theta_1), \sin(\theta_2), \dots, \sin(\theta_r)$, $r$ eigenvalues equal to $-\sin(\theta_1), -\sin(\theta_2), \dots, -\sin(\theta_r)$, and $d - 2r$ eigenvalues equal to $0$.

    \begin{proof}
        Let $\Phi := \U_1\U_1^\top - \U_2\U_2^\top \in \reals^{d \times d}$. We derive an exact expression for the characteristic polynomial $\det\Big(\Phi - \lambda \I_d  \Big)$. First, note 
        \begin{equation*}
            \Phi = \begin{bmatrix}
                \U_1 & \U_2
            \end{bmatrix} \begin{bmatrix}
                \U_1^\top \\
                -\U_2^\top
            \end{bmatrix},
        \end{equation*}
        and let $\U \mSigma \V^\top$ be a singular value decomposition of $\U_1^\top \U_2  \in \reals^{r \times r}$. Then, assuming $\lambda \neq 0$,
        \begin{align*}
            &\det\Big(\Phi - \lambda \I_d \Big) = (-1)^d \lambda^d \det\Big(\I_d - \frac{1}{\lambda}\Phi\Big) =  (-1)^d \lambda^d \det\bigg(\I_d - \frac{1}{\lambda} \begin{bmatrix}
                \U_1 & \U_2
            \end{bmatrix} \begin{bmatrix}
                \U_1^\top \\
                -\U_2^\top
            \end{bmatrix} \bigg) \\
            &\overset{(a)}{=}  (-1)^d \lambda^d \det\bigg(\I_{2r} - \frac{1}{\lambda} \begin{bmatrix}
                \I_r & \U \mSigma \V^\top \\
                -\V \mSigma \U^\top & -\I_r
            \end{bmatrix} \bigg) \\
            &= (-1)^d \lambda^d \det\bigg( \begin{bmatrix}
                (1 - 1/\lambda) \I_r & -(1/\lambda) \U \mSigma \V^\top \\
                (1/\lambda)\V \mSigma \U^\top & (1 + 1/\lambda) \I_r
            \end{bmatrix} \bigg) \\
            &\overset{(b)}{=} (-1)^d \lambda^d (1 - 1/\lambda)^r \det\bigg((1 + 1/\lambda)\I_r + \frac{(1 / \lambda^2)}{1 - 1/\lambda} \V \mSigma^2 \V^\top  \bigg) \\
            &= (-1)^d \lambda^d (1 - 1/\lambda)^r \det(\V) \det\bigg( (1 + 1/\lambda) \I_r + \frac{(1/\lambda^2)}{1 - 1/\lambda} \mSigma^2 \bigg) \det(\V^\top) \\
             &= (-1)^d \lambda^d \det\bigg( (1 - 1/\lambda^2) \I_r + (1/\lambda^2) \mSigma^2 \bigg) = (-1)^d \lambda^{d-2r} \prod_{\ell}^r \Big[ \lambda^2 - 1 + \cos^2(\theta_\ell) \Big] \\
            &= (-1)^d \lambda^{d-2r} \prod_{\ell=1}^r \Big[ \big( \lambda + \sin(\theta_\ell) \big) \big(\lambda - \sin(\theta_\ell) \big)
            \Big], 
        \end{align*}
        where $(a)$ is from Sylvester's Determinant Identity, and $(b)$ is from the fact that $$\det\bigg( \begin{bmatrix}
            \A & \B \\ \mC & \D
        \end{bmatrix} \bigg) = \det(\A)\det(\D - \mC \A^{-1} \B)$$ for invertible $\A$.
        Solving for the roots of $\det\Big(\Phi - \lambda\I_d\Big) = 0$ yields $\lambda = \pm \sin(\theta_\ell)$ for all $\ell \in [r]$. Therefore, $\Phi$ has $2r$ eigenvalues equal to $\pm \sin(\theta_1), \pm \sin(\theta_2), \dots, \pm \sin(\theta_r)$. Although we also have $\lambda = 0$ with multiplicity $d - 2r$, we initially assumed $\lambda \neq 0$, so these roots are invalid.
        
        We now show the remaining $d - 2r$ eigenvalues must be $0$. We showed there are \emph{at least} $2r$ eigenvalues that are non-zero, so $2r \leq \rank(\Phi).$ Additionally, we have $$\rank(\Phi) = \rank(\U_1\U_1^\top - \U_2\U_2^\top) \leq \rank(\U_1 \U_1^\top) + \rank(-\U_2 \U_2^\top) = 2r.$$
        Thus, $2r \leq \rank(\Phi) \leq 2r,$ which implies $\rank(\Phi) = 2r$. Therefore, $\Phi$ must have \emph{exactly} $2r$ non-zero eigenvalues, implying the remaining $d - 2r$ eigenvalues must all be equal to $0$.
    \end{proof}
\end{lemma}

\subsection{EXPECTATION OF RANDOM SYMMETRIC RANK-\texorpdfstring{$1$}{ } MATRICES} \label{ssec:expec-symm-rank-1}
We provide upper and lower bounds for $\eE[\ab\ab^\top]$ and $\eE[\blb\blb^\top]$. We first show $\eE[\ab\ab^\top]$ and $\eE[\blb\blb^\top]$ are isotropic matrices. 
\begin{lemma} \label{lem:aa-bb-theta-isotropic}
    Let $\w \sim \mathcal{N}(\0_d, \I_d)$, $\x := \U_1^\top \w$, $\y := \U_2^\top \w$, $\ab \sim \x \: \big| \: \|\x\|^2 > \|\y\|^2$, and $\blb \sim \y \: \big| \: \|\x\|^2 > \|\y\|^2$. Then, $\eE\big[\ab \ab^\top\big]$ and $\eE\big[\blb \blb^\top\big]$ are both isotropic matrices.
\end{lemma}
\begin{proof}
    Since $\covrm(\x) = \I_r$ and $\covrm(\y) = \I_r$, which are isotropic matrices, $\covrm(\ab)$ and $\covrm(\blb)$ are also isotropic matrices. Thus, it suffices to show $\eE[\ab] = \0_r$ and $\eE[\blb] = \0_r$.
    \begin{align*}
        \eE&[\ab] = \eE_{\x, \y \sim \calN(\0_r, \I_r)}\big[ \x \: | \: \|\x\|^2 > \|\y\|^2 \big] = \int\limits_0^\infty \int\limits_y^\infty \eE_{\x \sim \calN(\0_r, \I_r)}\big[ \x \: | \: \|\x\|^2 = x \big] f_{X, Y}(x, y) \: dx \: dy \overset{(a)}{=} \0_r,
    \end{align*}
    where $X, Y \sim \chi^2_r$, and $(a)$ is because $\x \: \big| \: \|\x\|^2 = x$ is distributed uniformly on the sphere of radius $\sqrt{x}$, so $\eE\big[\x \: | \: \|\x\|^2 = x \big] = \0_r$. We can use the same argument to show $\eE[\blb] = \0_r$. Therefore, $\eE[\ab \ab^\top] = \covrm(\ab)$ and $\eE[\blb \blb^\top] = \covrm(\blb)$, which are both isotropic matrices.
\end{proof}

\smallskip

\noindent The next result provides upper and lower bounds on $\eE[\ab \ab^\top]$ and $\eE[\blb \blb^\top]$.
\begin{lemma} \label{lem:expec-aa-bb-theta}
    Let $\w \sim \calN(\0_d, \I_d)$, $\x := \U_1^\top \w$, $\y := \U_2^\top \w$, $\ab \sim \x \: \big| \: \|\x\|^2 > \|\y\|^2$, and $\blb \sim \y \: \big| \: \|\x\|^2 > \|\y\|^2$. Then, we have
    \begin{align*}
       &\Bigg(1 + \sqrt{\frac{2}{\pi}} \cdot \frac{\sin(\theta_1)}{\sqrt{r+1}}\Bigg) \I_r \preceq \eE\big[ \ab \ab^\top \big] \preceq \Bigg(1 + \frac{1}{r} \cdot \sqrt{\sum\limits_{\ell=1}^r \sin^2(\theta_\ell)} \Bigg) \I_r, \; \; \text{and} \\
        &\Bigg( 1 - \frac{1}{r} \cdot \sqrt{\sum\limits_{\ell=1}^r \sin^2(\theta_\ell)} \Bigg) \I_r \preceq \eE\big[ \blb \blb^\top \big] \preceq \Bigg(1 - \sqrt{\frac{2}{\pi}} \cdot \frac{\sin(\theta_1)}{\sqrt{r+1}}\Bigg) \I_r.
    \end{align*}
\end{lemma}
\begin{proof}
    By Lemma~\ref{lem:aa-bb-theta-isotropic}, $\eE[\ab\ab^\top]$ is an isotropic matrix, so it suffices to upper and lower bound  $\trrm\big(\eE[\ab\ab^\top]\big) = \eE\big[\trrm(\ab\ab^\top)\big] = \eE[\|\ab\|^2]$.  By definition of $\ab$, $\|\ab\|^2 \sim \maxrm\{X, Y\}$, where $X, Y \sim \chi^2_r$ are not necessarily independent. We first note
    \begin{equation*}
        \|\ab\|^2 = \frac{1}{2} \Big( \|\x\|^2 + \|\y\|^2 + \big| \|\x\|^2 - \|\y\|^2 \big| \Big).
    \end{equation*}
     Therefore,
     \begin{equation} \label{eq:norm_ab_theta}
         \eE[\|\ab\|^2] = \frac{1}{2} \Big(\eE[\|\x\|^2] + \eE[\|\y\|^2] + \eE\big[ | \|\x\|^2 - \|\y\|^2 | \big] \Big) = r + \frac{1}{2}\Big(\eE\big[ | \|\x\|^2 - \|\y\|^2 | \big] \Big),
     \end{equation}
     so it suffices to upper and lower bound $\eE\big[ | \|\x\|^2 - \|\y\|^2 | \big]$. First, we have
     \begin{equation} \label{eq:expect-abs-norm-diff}
        \eE\Big[ \big| \|\x\|^2 - \|\y\|^2 \big| \Big] = \eE\Big[ \big| \|\U_1^\top \w\|^2 - \|\U_2^\top \w\|^2 \big| \Big] 
        = \eE\Big[ \big| \w^\top (\U_1 \U_1^\top - \U_2 \U_2^\top) \w \big| \Big]. 
     \end{equation}
     Let $\Phi := \U_1\U_1^\top - \U_2\U_2^\top$. We establish an upper bound as such:
     \begin{align*}
         \eE&\Big[ \big| \w^\top (\U_1 \U_1^\top - \U_2 \U_2^\top) \w \big| \Big] = \eE\Big[ \sqrt{(\w^\top \Phi \w)^2 } \Big] 
         \overset{(a)}{\leq} \sqrt{\eE\Big[ (\w^\top \Phi \w)^2 \Big]} \\ 
         &= \sqrt{\varrm(\w^\top \Phi \w)} \overset{(b)}{=} \sqrt{2\trrm\Big( \Phi^2 \Big)} = 2 \sqrt{\sum\limits_{\ell=1}^r \sin^2(\theta_\ell)},
     \end{align*}
     where $(a)$ is from Jensen's inequality, $(b)$ is from \cite[Eq. 381]{petersen2008matrix}, and the last equality is due to \Cref{lem:proj-mat-diff-eigvals}. Therefore, 
     \begin{equation*}
         \eE[\|\ab\|^2] \leq r + \sqrt{\sum\limits_{\ell=1}^r \sin^2(\theta_\ell)} \implies \eE[\ab \ab^\top] \preceq \Bigg(1 + \frac{1}{r} \cdot \sqrt{\sum\limits_{\ell=1}^r \sin^2(\theta_\ell)} \Bigg) \I_r.
     \end{equation*}
     We now establish a lower bound for $\eE\Big[ \big| \|\x\|^2 - \|\y\|^2 \big| \Big]$. Note $\Phi$ is symmetric, so there exists an eigendecomposition $\Phi = \Q \Lambda \Q^\top$ where $\Q \in \reals^{d \times d}$ is an orthogonal matrix, and $\Lambda$ is a diagonal matrix consisting of the eigenvalues of $\Phi$. We assume the eigenvalues are listed in descending order in $\Lambda$. By \Cref{lem:proj-mat-diff-eigvals}, $\Phi$ has $2r$ non-zero eigenvalues equal to $\pm \sin(\theta_1), \pm\sin(\theta_2), \dots, \pm\sin(\theta_r)$. Therefore:
     \begin{equation} 
         \w^\top \Phi \w = \w^\top \Q \Lambda \Q^\top \w := \z^\top \Lambda \z = \sum\limits_{\ell=1}^r \sin(\theta_\ell) \big( z_\ell^2 - z_{d - \ell + 1}^2 \big),
     \end{equation}
    where $\z := \Q^\top \w \sim \calN(\0_d, \I_d)$. Then, we have
    \begin{equation*}
        \eE\Big[ \big| \|\x\|^2 - \|\y\|^2 \big| \Big] = \eE\Big[ | \z^\top \Lambda \z | \Big] = \eE \bigg[ \Big| \sum\limits_{\ell=1}^r \sin(\theta_\ell) \big( z_\ell^2 - z_{d - \ell + 1}^2 \big) \Big| \bigg] \geq \sin(\theta_1) \eE\left[ \Big| \sum\limits_{\ell=1}^r \big( z_\ell^2 - z_{d - \ell + 1}^2 \big) \Big| \right].
    \end{equation*}
    Let $Z_1 := \sum\limits_{i=1}^r z_i^2$ and $Z_2 := \sum\limits_{\ell=1}^r z_{d - \ell + 1}^2$. Note $Z_1, Z_2 \overset{\text{iid}}{\sim} \chi^2_r$. 
    Substituting this lower bound into \eqref{eq:expect-abs-norm-diff} yields
    \begin{align*}
        \eE\Big[ \big| \|\x\|^2 - \|\y\|^2 \big| \Big] &\geq \sin(\theta_1) \eE\Big[ \big| Z_1 - Z_2 \big| \Big] 
        \overset{(c)}{=} \frac{4\sin(\theta_1)}{\sqrt{\pi}} \frac{\Gamma((r+1)/2)}{\Gamma(r/2)},
    \end{align*}
    where $(c)$ is from the fact that $|Z_1 - Z_2| = \max\{Z_1, Z_2\} - \min\{Z_1, Z_2\}$ and \Cref{lem:expec-max-min-chi-squares}. We can then lower bound $\frac{\Gamma((r+1)/2)}{\Gamma(r/2)}$ as such. First, let $x := r/2$. Then, by Wendel's Inequality,
    \begin{align*}
        &\frac{\Gamma(x + 1/2)}{x^{1/2}\Gamma(x)} \geq \bigg( \frac{x}{x + 1/2} \bigg)^{1/2} \iff \frac{\Gamma((r + 1)/2)}{\Gamma(r/2)} \geq \frac{r}{\sqrt{2(r+1)}},
    \end{align*}
    so
    \begin{equation*}
        \eE\Big[ \big| \|\x\|^2 - \|\y\|^2 \big| \Big] \geq \frac{4r\sin(\theta_1)}{\sqrt{2\pi(r+1)}}.
    \end{equation*}
    Substituting this lower bound into \eqref{eq:norm_ab_theta} yields
    \begin{equation*}
        \eE\big[ \|\ab\|^2 \big] \geq r + \sqrt{\frac{2}{\pi}} \cdot \frac{r\sin(\theta_1)}{\sqrt{r+1}} \implies \eE\big[ \ab \ab^\top \big] \succeq \Bigg(1 + \sqrt{\frac{2}{\pi}} \cdot \frac{\sin(\theta_1)}{\sqrt{r+1}}\Bigg) \I_r.
    \end{equation*}
    We can then use the fact $\eE\big[ \|\ab\|^2 + \|\blb\|^2 \big] = 2r$ to show
    \begin{equation*}
         \Bigg( 1 - \frac{1}{r} \cdot \sqrt{\sum\limits_{\ell=1}^r \sin^2(\theta_\ell)} \Bigg) \I_r \preceq \eE\big[ \blb \blb^\top \big] \preceq \Bigg(1 - \sqrt{\frac{2}{\pi}} \cdot \frac{\sin(\theta_1)}{\sqrt{r+1}}\Bigg) \I_r.
    \end{equation*}
\end{proof}

\subsection{MATRIX BERNSTEIN'S INEQUALITY}
We use Bernstein's matrix inequality to bound the largest and smallest eigenvalues of sums of independent, random symmetric matrices.

\begin{lemma}[ Bernstein's Inequality, adapted from Theorem 6.2 in \cite{tropp2012user} ]
\label{lem:bernstein-inequality}
    Let $\X_1, \dots, \X_n$ be independent random symmetric matrices of dimension $m$. Assume that there exist a positive number $R$ and matrices $\A_i$ such that
    \begin{equation*}
        \eE[\X_i^p] \preceq \frac{p!}{2} \cdot R^{p-2} \cdot \A_i^2 
    \end{equation*}
    for all $i \in [n]$ and integers $p \geq 2$. Then, for all $t \geq 0$:
    \begin{equation*}
        P\Bigg( \lambda_1\bigg( \sum\limits_{i=1}^n \X_i - \eE[\X_i] \bigg) \geq t \Bigg) \leq m \cdot \exprm\bigg( -\frac{t^2}{2(\sigma^2 + Rt)} \bigg),
    \end{equation*}
    where $\sigma^2 = \sigma_1\bigg( \sum\limits_{i=1}^n \A_i^2 \bigg)$.
\end{lemma}
\noindent We refer to the condition $\eE[\X_i^p] \preceq \frac{p!}{2} \cdot R^{p-2} \cdot \A_i^2$ as Bernstein's condition. Our next result states $\ab\ab^\top$ and $\blb\blb^\top$ satisfy Bernstein's condition.
\begin{lemma} \label{lem:aa-bb-theta-bernstein-condition}
    Let $\w \sim \calN(\0_d, \I_d)$ $\x := \U_1^\top \w$, $\y := \U_2^\top \w$, $\ab \sim \x \: \big| \: \|\x\|^2 > \|\y\|^2$, and $\blb \sim \y \: \big| \: \|\x\|^2 > \|\y\|^2$. Then, we have
    \vspace{-0.25cm}
    \begin{align*}
        &\eE\big[ (\ab \ab^\top)^{^p} \big] \preceq \frac{p!}{2} \cdot (2r)^{p-2} \cdot 8r^2 \I_r, \; \; \text{and} \; \; \eE\big[ (\blb \blb^\top)^{^p} \big] \preceq \frac{p!}{2} \cdot (2r)^{p-2} \cdot 8r^2 \I_r
    \end{align*}
    for all integers $p \geq 1$.
\end{lemma}
\begin{proof}
    We first focus on $\eE\big[(\ab \ab^\top)^{^p}\big]$. It suffices to upper bound $\lambda_1\Big(\eE\big[ (\ab \ab^\top)^{^p} \big] \Big)$:
    \begin{equation*}
        \lambda_1\Big(\eE\big[ (\ab \ab^\top) \big] \Big) \overset{(a)}{\leq} \eE\Big[ \lambda_1\big( (\ab \ab^\top)^{^p} \big) \Big] \overset{(b)}{=} \eE\big[ (\|\ab\|^2)^{^p} \big],
    \end{equation*}
    where $(a)$ is due to Jensen's inequality, and $(b)$ is because $(\ab \ab^\top)^{^p}$ is a rank-$1$ matrix for all integers $p \geq 1$. Recall $\|\ab\|^2 \sim \maxrm\{X, Y\}$, where $X := \|\x\|^2$ and $Y := \|\y\|^2$. Therefore,
    \begin{align*}
        &\eE\big[ (\|\ab\|^2)^{^p} \big] =  \int\limits_0^\infty \int\limits_0^\infty \maxrm\{x, y\}^p f_{X, Y}(x, y) \: dx \: dy = \int\limits_0^\infty \int\limits_0^\infty \maxrm\{x^p, y^p\} f_{X, Y}(x, y) \: dx \: dy \\
        &= 2 \int\limits_0^\infty \int\limits_{y}^\infty x^p f_{X, Y}(x, y) \: dx \: dy \overset{(a)}{\leq} 2 \int\limits_0^\infty \int\limits_0^\infty x^p f_{X | Y}(x | y) f_{Y}(y) \: dx \: dy = 2 \int\limits_0^\infty \eE_{X \sim \chi^2_r}\big[ X^p \: | \: Y \big] f_{Y}(y) \: dy \\ &= 2 \eE_{Y \sim \chi^2_r} \Big[ \eE_{X \sim \chi^2_r} \big[ X^p \: | \: Y \big] \Big]
        = 2 \eE\big[ X^p \big] \overset{(b)}{\leq} p! (2r)^p = \frac{p!}{2} \cdot (2r)^{p-2} \cdot 8r^2,
    \end{align*}
    where $(a)$ is because $X$ and $Y$ have non-negative support, and $(b)$ is from Lemma A.6 in \citep{qu2014finding}. Therefore:
    \begin{equation*}
        \eE\big[ (\ab \ab^\top)^{^p} \big] \preceq \frac{p!}{2} \cdot (2r)^{p-2} \cdot 8r^2 \I_r = \frac{p!}{2} \cdot R_a^{p-2} \cdot \A^2, 
    \end{equation*}
    where $R_a = 2r$ and $\A^2 = 8r^2 \I_r$. We can bound $\eE\big[(\blb \blb^\top)^{^p}\big]$ in a similar manner to obtain:
    \begin{equation*}
        \eE\big[ (\blb \blb^\top)^{^p} \big] \preceq \frac{p!}{2} \cdot (2r)^{p-2} \cdot 8r^2 \I_r = \frac{p!}{2} \cdot R_b^{p-2} \cdot \B^2, 
    \end{equation*}
    where $R_b = 2r$ and $\B^2 = 8r^2 \I_r$.
\end{proof}

\section{PROOF OF THEOREM~\ref{thm:binary-lin-sep}} \label{app:thm-1-proof}
We now provide the full proof of \Cref{thm:binary-lin-sep}. Let $\X := \W \U_1$ and $\Y := \W \U_2$, and $\x_n$ and $\y_n$ denote the $n^{th}$ row in $\X$ and $\Y$, respectively, written as column vectors. Note $\x_n = \U_1^\top \w_n$ and $\y_n = \U_2^\top \w_n$, where $\w_n \overset{iid}{\sim} \calN(\0_d, \I_d)$. 

\subsection{CONDITIONS FOR LINEAR SEPARABILITY} \label{ssec:lin-sep-conditions}
We first identify necessary and sufficient conditions to achieve linear separability between $f(\calS_1)$ and $f(\calS_2)$. By definition of linear separability, we aim to show there exists a $\v \in \reals^D$ such that \eqref{eq:lin-sep-problem} holds for all $\bm \alpha \in \reals^r \setminus \{\0_r\}$. Focusing only on $\U_1$, we can re-write \eqref{eq:lin-sep-problem} under \Cref{assum:network} as such:
\begin{align*}
    \v^\top &f\big( \U_1 \bm \alpha \big) = \sum\limits_{n=1}^D v_n (\w_n^\top \U_1 \bm \alpha)^2 = \sum\limits_{n=1}^D v_n (\w_n^\top \U_1 \bm \alpha) (\w_n^\top \U_1 \bm \alpha) = \sum\limits_{n=1}^D v_n (\bm \alpha^\top \U_1^\top \w_n)(\w_n^\top \U_1 \bm \alpha) \\ 
    &= \bm \alpha^\top \bigg( \sum\limits_{n=1}^D v_n \U_1^\top \w_n \w_n^\top \U_1 \bigg) \bm \alpha = \bm \alpha^\top \bigg( \sum\limits_{n=1}^D v_n \x_n \x_n^\top \bigg) \bm \alpha > 0 \iff \sum\limits_{n=1}^D v_n \x_n \x_n^\top \succ 0. \label{eq:X-outer-pd}
\end{align*}
We can re-write the $\U_2$ part of \eqref{eq:lin-sep-problem} similarly to obtain the following necessary and sufficient conditions for linear separability:
\begin{equation} \label{eq:suff-nec-conditions}
    \sum\limits_{n=1}^D v_n \x_n \x_n^\top \succ 0 \; \; \text{and} \; \; \sum\limits_{n=1}^D v_n \y_n \y_n^\top \prec 0.
\end{equation}

\noindent We then construct the linear classifier $\v$ with the following entries:

\smallskip

\begin{center}
    \textit{For all $n \in [D]$, $v_n = \mathrm{sign}\big( \|\x_n\|^2 - \|\y_n\|^2 \big)$.}
\end{center}

\smallskip

\noindent With this choice of $\v$, \eqref{eq:suff-nec-conditions} becomes
\begin{equation*} 
    \mS_1 := \sum\limits_{i \in \calI} \x_i \x_i^\top - \sum\limits_{j \in \calI^c} \x_j \x_j^\top \succ 0 \; \; \text{and} \; \; \mS_2 := \sum\limits_{i \in \calI} \y_i \y_i^\top - \sum\limits_{j \in \calI^c} \y_j \y_j^\top \prec 0,
\end{equation*}
where $\calI := \{n \in [D]: v_n = +1\}$ and $\calI^c := \{n \in [D]: v_n = -1\}$. We now upper bound the failure probability $P\Big( \mS_1 \not \succ 0 \cup \mS_2 \not \prec 0 \Big)$.

\smallskip

\subsection{BOUNDING THE FAILURE PROBABILITY} We aim to upper bound $P\Big(\mS_1 \not \succ 0 \cup \mS_2 \not \prec 0\Big)$ by some (arbitrarily) small $\delta \in (0, 1)$.
It suffices to upper bound $P\Big(\mS_1 \not \succ 0\Big)$ and $P\Big(\mS_2 \not \prec 0\Big)$ individually due to the union bound. Let $\gamma_1 := \sqrt{\frac{2}{\pi}} \cdot \frac{\sin(\theta_1)}{\sqrt{r+1}}$ and $\gamma_2 := \frac{1}{r} \cdot \sqrt{\sum\limits_{\ell=1}^r \sin^2(\theta_\ell)}$. Also let $\alpha_1 := 1 + \gamma_1$, $\alpha_2 := 1 + \gamma_2$, $\beta_1 := 1 - \gamma_1$, and $\beta_2 := 1 - \gamma_2$. 

We first upper bound $P\Big(\mS_1 \not \succ 0\Big).$ Note $\mS_1 \not \succ 0$ if and only if $\lambda_r(\mS_1) \leq 0$. By Lemma~\ref{lem:aa-bb-theta-bernstein-condition}, $\mS_1$ and $\mS_2$ are sums of random matrices that satisfy Bernstein's condition. Therefore, we can bound $P\Big(\mS_1 \not \succ 0\Big) = P\Big(\lambda_r(\mS_1) \leq 0\Big)$ using Bernstein's inequality:
\begin{align}
    P&\Big(\Sb_1 \not \succ 0\Big) = P\Big(\lambda_r(\Sb_1) \leq 0\Big) \nonumber = P\Big(\lambda_r(\Sb_1) - \lambda_r\big( \eE[\Sb_1] \big) \leq -\lambda_r\big( \eE[\Sb_1]\big) \Big) \nonumber \\
    &\overset{(a)}{\leq} P\Big( \lambda_r\big(\Sb_1 - \eE[\Sb_1]\big) \leq -\lambda_r\big(\eE[\Sb_1]\big) \Big) \nonumber = P\Big( \lambda_1\big(-\Sb_1 - \eE[-\Sb_1]\big) \geq \lambda_r\big(\eE[\Sb_1]\big) \Big) \nonumber \\
    &\overset{(b)}{\leq} r \cdot \exprm\Bigg( -\frac{\lambda_r\big(\eE[\Sb_1]\big)^2}{16r^2D + 4r\lambda_r\big(\eE[\Sb_1]\big)} \Bigg), \label{eq:S1-bernstein}
\end{align}
where $(a)$ is due to Weyl's inequality, and $(b)$ is from Lemma~\ref{lem:bernstein-inequality}. We now upper and lower bound $\eE[\mS_1]$ as follows. Let $Q := \frac{1}{D} \sum\limits_{n=1}^D \mathbbm{1}[v_n = +1]$. First, using \Cref{lem:expec-aa-bb-theta},
\begin{equation*}
   (2Q - \beta_1)D \I_r \preceq \eE\Big[ \mS_1 \: | \: Q \Big] \preceq (2Q - \beta_2) D \I_r. 
\end{equation*}
Then, taking the expectation over $Q$ yields
\begin{equation} \label{eq:E-S1}
    \gamma_1 D \I_r \preceq \eE[\mS_1] \preceq \gamma_2 D \I_r.
\end{equation}
Therefore, $\gamma_1 D \leq \lambda_r\big(\eE[\mS_1]\big) \leq \gamma_2 D $. Substituting \eqref{eq:E-S1} into \eqref{eq:S1-bernstein} leads to
\begin{equation*}
    P\Big(\Sb_1 \not \succ 0\Big) \leq r \cdot \exprm\Bigg( -\frac{\gamma_1^2 D}{16r^2 + 4\gamma_2r} \Bigg).
\end{equation*}
By similar argument, we can show by $-\gamma_2D \I_r \preceq \eE[\mS_2] \preceq -\gamma_1D \I_r$ that
\begin{equation*}
    P\Big( \mS_2 \not \prec 0 \Big) \leq r \cdot \exprm\bigg( -\frac{\gamma_1^2 D}{16r^2 + 4\gamma_2 r} \bigg).
\end{equation*}

\smallskip


\noindent We then apply the union bound on the failure probability:
\begin{equation} \label{eq:failure-prob-bound}
    P\Big(\mS_1 \not \succ 0 \cup \mS_2 \not \prec 0\Big) \leq 2r \cdot \exprm\bigg( -\frac{\gamma_1^2 D}{16r^2 + 4\gamma_2 r} \bigg).
\end{equation}

\subsection{FINAL RESULT}
Upper bounding \eqref{eq:failure-prob-bound} by some (arbitrarily small) $\delta \in (0, 1)$, and then re-arranging the terms to lower bound $D$, results in
\begin{equation} \label{eq:D-gammas}
    D \geq \frac{16r^2 + 4\gamma_2 r}{\gamma_1^2} \cdot \log\bigg(\frac{2r}{\delta}\bigg).
\end{equation}
Substituting the definitions of $\gamma_1$ and $\gamma_2$, as well as $\theta_{min} := \theta_1$, into \eqref{eq:D-gammas} leads to our final result. Let $\delta \in (0, 1)$. Then, $\mS_1 \succ 0$ and $\mS_2 \prec 0$, and thus $f(\calS_1)$ and $f(\calS_2)$ are linearly separable, if the network width $D$ satisfies
\begin{equation*} 
    D \geq \frac{2\pi  \Bigg(4 r^2 + \sqrt{\sum\limits_{\ell=1}^r\sin^2(\theta_\ell)} \ \Bigg)  (r+1)}{\sin^2(\theta_{min})} \cdot \log\bigg(\frac{2r}{\delta}\bigg) = \mathcal{O}\left( \frac{r^3}{\sin^2(\theta_{min})} \cdot \log\left(\frac{r}{\delta}\right) \right).
\end{equation*}

\section{ADDITIONAL EXPERIMENTAL DETAILS}
In this section, we discuss the experimental setup and results for \Cref{fig:separability-d-r,fig:rank-K-sweep,fig:linear_probe_depth_3}. 

\subsection{PHASE TRANSITION IN TERMS OF INTRINSIC DIMENSION} \label{ssec:phase-transition} 
In this subsection, we describe the setup and results in \Cref{fig:separability-d-r}, which verifies the required network width to achieve linear separability of the initial-layer features grows polynomially w.r.t. the intrinsic dimension. This experiment was run on a MacBook Air with an Apple M3 chip.

\paragraph{Setup.} Over 25 trials, we randomly sampled two matrices $\U_1, \U_2$ from the $d \times r$ Stiefel manifold, and a weight matrix $\W \in \reals^{D \times d}$ with iid standard Gaussian entries.
We varied the ambient dimension $d$ while keeping the intrinsic dimension $r$ fixed, and also varied $r$ while keeping $d$ fixed. In both settings, we tested different layer widths $D$. For each combination of $(D,d)$ and $(D,r)$, we checked for linear separability using the necessary and sufficient conditions \eqref{eq:lin-sep-outer-sum}, and  recorded the proportion of successful trials.

\paragraph{Results.} As seen in \Cref{fig:separability-d-r}, when $d$ increases for a fixed $r$, the values of $D$ at which the proportion of successful trials transitions from $0$ to $1$, or the \emph{phase transition}, remains constant. In contrast, as $r$ increases for a fixed $d$, this phase transition region clearly increases. Thus, \Cref{fig:separability-d-r} verifies the required width to achieve linear separability of the random features only depends on the intrinsic dimension of the subspaces.

\subsection{DEPENDENCE ON DIMENSION AND NUMBER OF CLASSES: QUADRATIC VS. RELU} \label{sapp:quad-relu-comp}
Here, we describe the setup and results in \Cref{fig:rank-K-sweep}, which shows the required widths of quadratic and ReLU random feature models have similar dependence on the subspace dimension $r$, and number of classes $K$, to achieve linear separability of a UoS. All experiments were run on a single NVIDIA A40 GPU.

\paragraph{Setup.} For all experiments, we set $d = 128$. In \Cref{subfig:rank-sweep}, we set $K = 2$ and swept through $r$ from $2^2$ to $2^6$ by powers of $2$. In \Cref{subfig:K-sweep}, we fixed $r = 4$ and swept through the number of subspaces $K$ from $2$ to $2^5$, again by powers of $2$. In both sweeps, we varied the network width $D$ from $2^5$ to $2^{10}$ by powers of $2$.

\subsection{LINEAR SEPARABILITY OF FEATURES: RANDOM VS. TRAINED WEIGHTS} \label{sapp:random_trained_lin_sep} 
We first describe the setup and results in \Cref{fig:linear_probe_depth_3}, which investigates how training the network weights away from their random initialization impacts the linear separability of the initial-layer features. Here, all experiments were run on a single NVIDIA V100 GPU.

\paragraph{Setup.} We first created a training set using the above data generation process with $K = 2$, $d = 16$, and $r = 4$. We then trained two 3-layer MLPs of width $D=128$ for $100$ epochs. One MLP had ReLU activations, and the other had quadratic activations. After each training epoch, we performed a linear probing on the features extracted by the two hidden layers. At initialization (marked by a star), all weights were sampled i.i.d. from a zero-mean Gaussian distribution. We averaged the results over $5$ trials. 

\paragraph{Results.} Across all $5$ trials in both MLPs, the features from the hidden layers were linearly separable at random initialization, as evidenced by the perfect linear probing accuracy. After each epoch, the linear probe accuracy remained perfect, implying the features from the hidden layers remained linearly separable during training. Thus, training the weights away from the random initialization does not impact the linear separability of the features.
















\fi

\end{document}